\documentclass{article}

\usepackage[preprint,nonatbib]{neurips_2022}

\usepackage[utf8]{inputenc} % allow utf-8 input
\usepackage[T1]{fontenc}    % use 8-bit T1 fonts
\usepackage{hyperref}       % hyperlinks
\usepackage{url}            % simple URL typesetting
\usepackage{booktabs}       % professional-quality tables
\usepackage{amsfonts}       % blackboard math symbols
\usepackage{nicefrac}       % compact symbols for 1/2, etc.
\usepackage{microtype}      % microtypography
\usepackage{xcolor}         % colors
\usepackage{adjustbox}
\usepackage{multirow}
\usepackage{caption}
\usepackage{amssymb}
\usepackage{rotating}

\definecolor{mycolor}{HTML}{F7F8E0}
\definecolor{darkcyan}{RGB}{0,113,194}
\definecolor{forestgreen}{RGB}{34,139,34}
\definecolor{customyellow}{RGB}{255,217,102}
\definecolor{mypink3}{cmyk}{0, 0.7808, 0.4429, 0.1412}

\title{DISCO: Adversarial Defense with \\ Local Implicit Functions}

\author{%
  Chih-Hui Ho \quad Nuno Vasconcelos \\
  Department of Electrical and Computer Engineering\\
  University of California, San Diego\\
  \texttt{\{chh279, nvasconcelos\}@ucsd.edu} \\
}

\begin{document}

\maketitle

\begin{abstract}
The problem of adversarial defenses for image classification, where the goal is to robustify a classifier against adversarial examples, is considered. Inspired by the hypothesis that these examples lie beyond the natural image manifold, a novel {\it aDversarIal defenSe with local impliCit functiOns \/} (DISCO) is proposed to remove adversarial perturbations by localized manifold projections. DISCO consumes an adversarial image and a query pixel location and outputs a clean RGB value at the location. It is implemented with an encoder and a local implicit module, where the former produces per-pixel deep features and the latter uses the features in the neighborhood of query pixel for predicting the clean RGB value. Extensive experiments demonstrate that both DISCO and its cascade version outperform prior defenses, regardless of whether the defense is known to the attacker. DISCO is also shown to be data and parameter efficient and to mount defenses that transfers across datasets, classifiers and attacks. Code released. \footnote{Code availabe at \url{https://github.com/chihhuiho/disco.git}}

\end{abstract}

%\vspace{-10pt}
\section{Introduction}
%\vspace{-5pt}
\begin{figure}[htb]
\begin{tabular}{ccc}
    \includegraphics[width=0.3\linewidth]{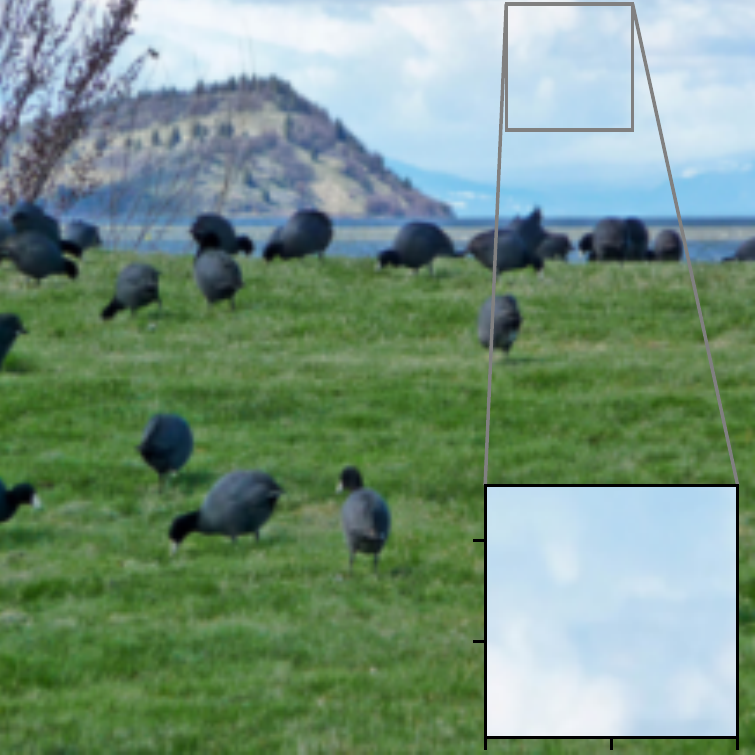} & \includegraphics[width=0.3\linewidth]{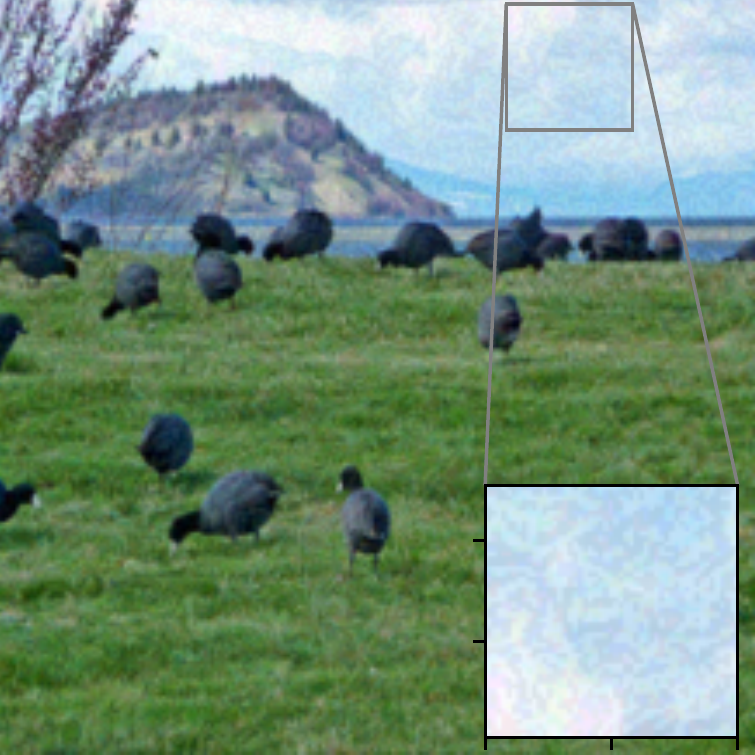} & \includegraphics[width=0.3\linewidth]{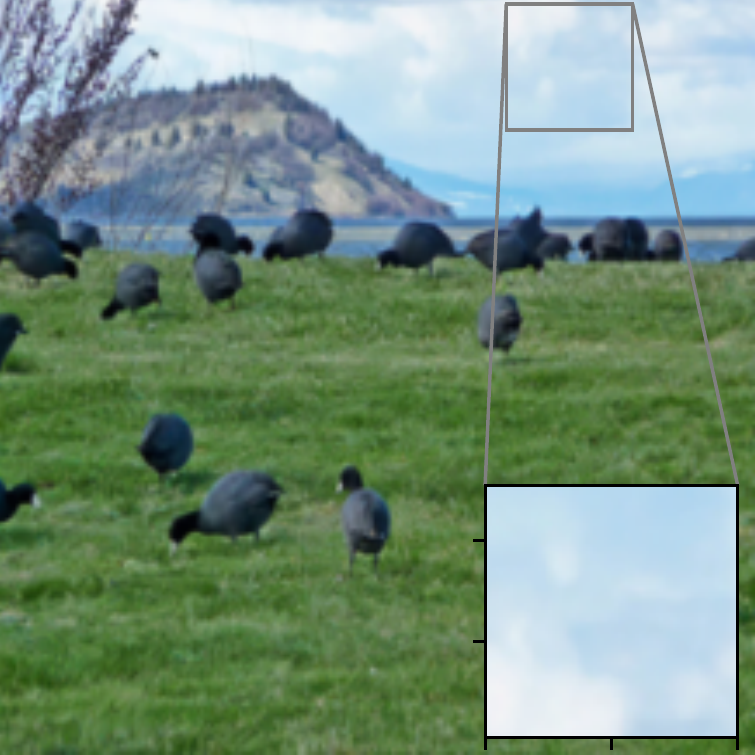} \\
    \scriptsize{(a) Clean Image} & \scriptsize{(b) Adversarial Image} & \scriptsize{(c) DISCO Output}
\end{tabular}
\vspace{-5pt}
\caption{Qualitative performance of DISCO output of a randomly selected ImageNet~\cite{deng2009imagenet} image.}
\label{fig:vis}
\vspace{-5pt}
\end{figure}

\begin{figure}[htb]
\begin{tabular}{cccc}
    \includegraphics[width=0.22\linewidth]{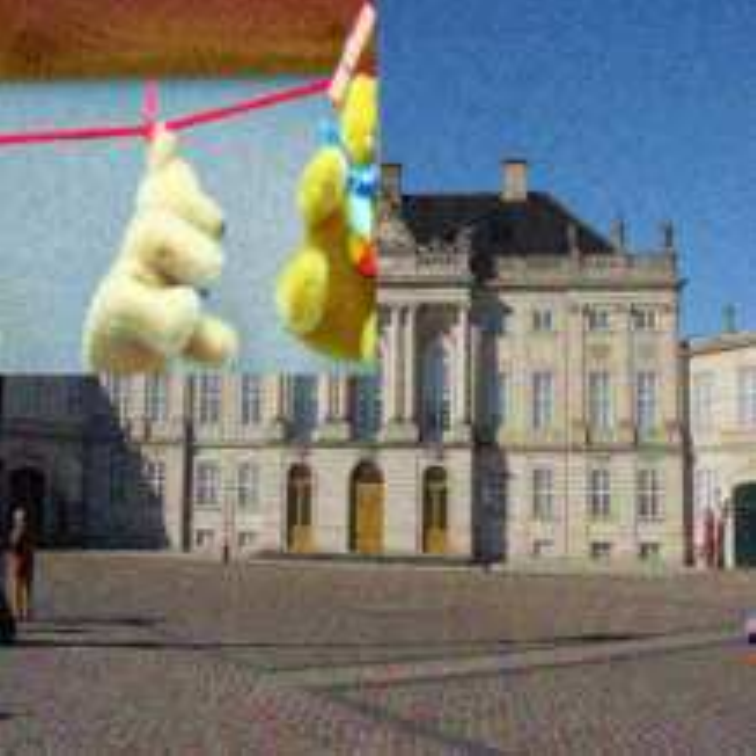} & \includegraphics[width=0.22\linewidth]{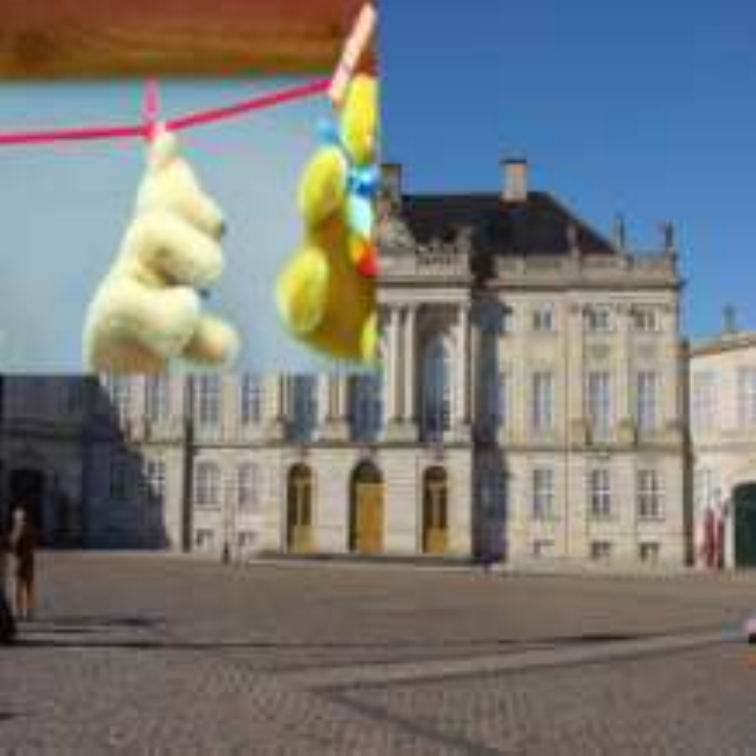} & \includegraphics[width=0.22\linewidth]{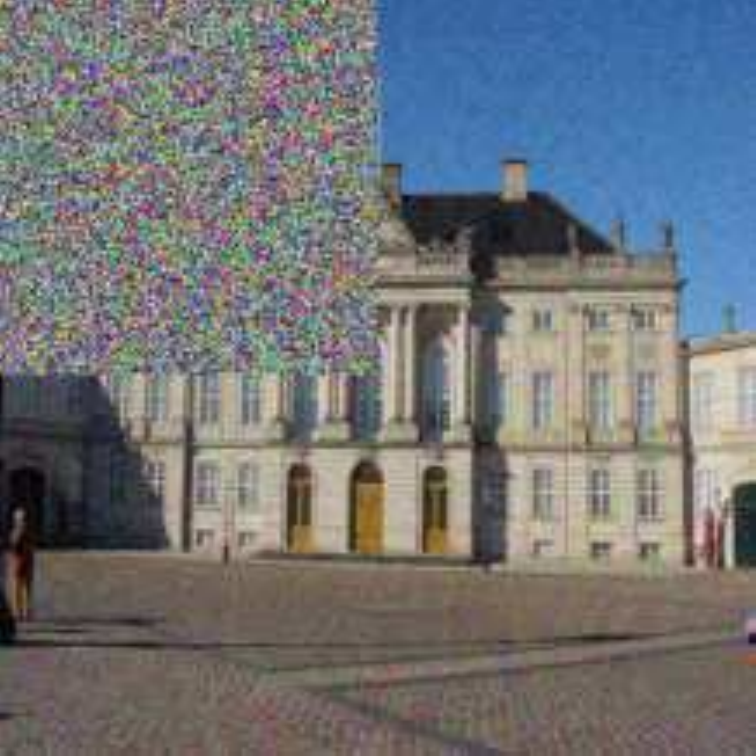}
    & \includegraphics[width=0.22\linewidth]{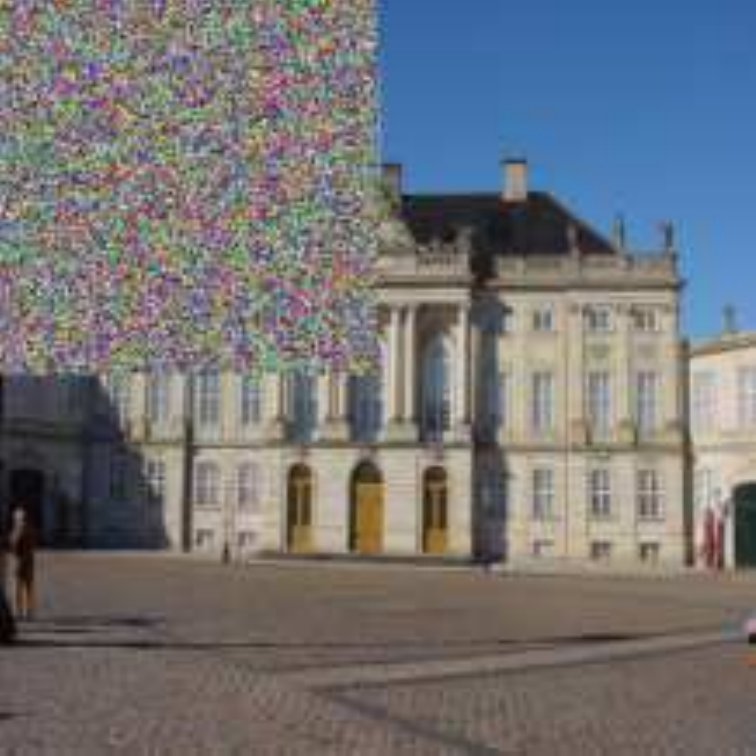}\\
    \scriptsize{(a)} & \scriptsize{(b)} & \scriptsize{(c)} & \scriptsize{(d)}
\end{tabular}
%\vspace{-5pt}
\caption{DISCO is a conditional model of local 
%image
patch statistics. It performs a local manifold projection per pixel neighborhood, conditional on feature vectors of the adversarial image. This is critical to enable learning from limited data while achieving the model expressiveness needed to precisely control the manifold projection. (a) A mixed image of 2 adversarial images. (b) DISCO output for (a). (c) A mixed image of an adversarial image and noise. (d) DISCO output for (c).
}
\label{fig:mixes}
\vspace{-15pt}
\end{figure}

It has long been hypothesized that vision is only possible because the natural world contains substantial statistical regularities, which are exploited by the vision system to overcome the difficulty of scene understanding~\cite{barlow,field,Ruderman1994TheSO,Field1987RelationsBT,vasconcelos08,Field1994WhatIT,786990,Simoncelli2001NaturalIS,Srivastava2004OnAI,Olshausen1996EmergenceOS,BELL19973327,Torralba2003StatisticsON}. Under this hypothesis, natural images form a low-dimension manifold in image space, denoted as the {\it image manifold\/}, to which human vision is highly tuned. While deep neural networks (DNNs)~\cite{inceptionnet,He2016DeepRL,BMVC2016_87,efficientnet,vgg} aim to classify natural images with human-like accuracy, they have been shown prone to adversarial attacks that, although imperceptible to humans, significantly decrease their performance.
As shown in Fig.~\ref{fig:vis} (a) and (b), these attacks typically consist of adding an imperceptible perturbation, which can be generated in various manners~\cite{fgsm, pgd, cw, bim}, to the image. Over the past few years, the introduction of more sophisticated attacks has  exposed a significant vulnerability of DNNs to this problem~\cite{onepixelattack,tifgsm,mifgsm,autoattack,apgd,jitter}. In fact, it has been shown that adversarial examples crafted with different classifiers and optimization procedures even transfer across networks~\cite{Demontis2019WhyDA,Luke21,difgsm,Huang2020BlackBox,Papernot17}.

% Such phenomenon has been attributed to the reason that natural images lie on the low-dimensional manifolds and the adversarial examples are beyond the manifold of natural images~\cite{pixeldefend,defensegan,Lee2017GenerativeAT,gilmer2018adversarial,Khoury2018OnTG,Bai2019HilbertBasedGD}.

A potential justification for the success of adversarial attacks and their transferability is that they create images that lie just barely outside the image manifold~\cite{pixeldefend,defensegan,Lee2017GenerativeAT,gilmer2018adversarial,Khoury2018OnTG,Bai2019HilbertBasedGD}. We refer to these images as {\it barely outliers\/}. While humans have the ability to project these images into the manifold, probably due to an history of training under adverse conditions, such as environments with low-light or ridden with occlusions, this is not the case for current classifiers. A key factor to the success of the human projection is likely the accurate modeling of the image manifold.
%, but this has been shown challenging~\cite{Oord2016PixelRN,Benigno14,Benigno13,Oord2016ConditionalIG,vasconcelos08}. Nevertheless, s
Hence, several defenses against adversarial attacks are based on models of natural image statistics. These are usually global image representations~\cite{magnet,stl,defensegan,Yuan2020EnsembleGC} or conditional 
models of image pixel statistics~\cite{pixeldefend,Bai2019HilbertBasedGD, Oord2016PixelRN,Salimans2017PixelCNNIT,Kolesnikov17}.
For example, PixelDefend~\cite{pixeldefend} and HPD~\cite{Bai2019HilbertBasedGD} project malicious images into the natural image manifold using the PixelCNN model~\cite{Kolesnikov17}, which predicts a pixel value conditioned on the remainder of the image. However, these  strategies~\cite{pixeldefend,Bai2019HilbertBasedGD,magnet,stl,defensegan} can be easily defeated by existing attacks. We hypothesize that this is due to the difficulty of learning generative image models, which require {\it global\/} image modelling, a highly complex task. It is well known that the synthesis of realistic natural images requires very large model sizes and training datasets~\cite{Karras2019ASG,biggan,Karras2020AnalyzingAI}. Even with these, it is not clear that the manifold is modeled in enough detail to defend adversarial attacks.
%, e.g. whether these even lie outside the set of images considered "realistic" by the model. 

In this work, we argue that, unlike image synthesis, the manifold projection required for adversarial defense is a {\it conditional\/} operation: the synthesis of a natural image {\it given\/} the perturbed one. Assuming that the attack does not alter the {\it global\/} structure of the image (which would likely not be {\it imperceptible\/} to humans) it should suffice for this function to be a conditional model of {\it local\/} image (i.e. patch) statistics. We argue that this conditional modeling can be implemented with an implicit function~\cite{chen2021learning,Sitzmann2020ImplicitNR,Mildenhall2020NeRFRS, Saito2019PIFuPI,Niemeyer2020DifferentiableVR,semantic_srn,Jiang2020LocalIG,Chen2019LearningIF}, where the network learns a conditional representation of the image appearance in the neighborhood of each pixel, given a feature extracted at that pixel. This strategy is denoted {\it aDversarIal defenSe with local impliCit functiOns\/} (DISCO). Local implicit models have recently been shown to provide impressive results for 3D  modeling~\cite{Sitzmann2020ImplicitNR,Mildenhall2020NeRFRS, Saito2019PIFuPI,Niemeyer2020DifferentiableVR,NIPS2019_8340,semantic_srn,Jiang2020LocalIG,Chen2019LearningIF,Mescheder2019OccupancyNL,Park2019DeepSDFLC,Genova2020LocalDI,Wu2020IFDefense3A} and image interpolation~\cite{chen2021learning}. We show that such models can be trained to project barely outliers into the 
%image manifold
patch manifold, with much smaller parameter and dataset sizes than generative models, while enabling much more precise control of the manifold projection operation. This is illustrated in Fig.~\ref{fig:vis}, which presents an image, its adversarial attack, and the output of the DISCO defense. 
%In particular, to perform the manifold projection, it should suffice to learn the conditional statistics of small pixel neighborhoods ($3 \times 3$ pixels) given a feature vector that encodes the appearance of that neighborhood.  This
%quite learn the global feature and are computationally inefficient when tracking the data likelihood of all the pixels.
% DISCO predicts the pixel value using the deep feature of a local patch and multiple pixels can be predicted simultaneously. It adopts the technique of implicit function~\cite{Sitzmann2020ImplicitNR}, which represents a signal in continuous fashion using DNN. Despite implicit function has been adopted in tasks like super-resolution~\cite{chen2021learning}, shape modeling~\cite{Niemeyer2020DifferentiableVR,Saito2019PIFuPI,Park2019DeepSDFLC}, 3D scene reconstruction~\cite{Mildenhall2020NeRFRS} or even point cloud adversarial defense~\cite{Mildenhall2020NeRFRS}, it remains an unexplored area for adversarial defense on images. 
To train DISCO, a dataset of adversarial-clean pairs is first curated. During training, DISCO inputs an adversarial image and a query pixel location, for which it predicts a new RGB value. This is implemented with a feature encoder and a local implicit function. The former is composed by a set of residual blocks with stacked convolution layers and produces a deep feature per pixel. The latter consumes the query location and the features in a small neighborhood of the query location. The implicit function is learned to minimize the $L_1$ loss between the predicted RGB value and that of the clean image. 

The restriction of the manifold modeling to small image neighborhoods is a critical difference with respect to previous defenses based on the modeling of the natural image manifold. Note that, as shown in Fig.~\ref{fig:mixes}, DISCO does not project the entire image into the manifold, only each pixel neighborhood. This considerably simplifies the modeling and allows a much more robust defense in a parameter and data efficient manner. This is demonstrated by evaluating the performance of DISCO under both the oblivious and adaptive settings~\cite{Yin2020GAT,REN2020346,Xu2018FeatureSD}. Under the oblivious setting, the popular RobustBench~\cite{croce2021robustbench} benchmark is considered, for both $L_{\infty}$ and $L_{2}$ attacks with Autoattack~\cite{autoattack}. DISCO achieves SOTA robust accuracy (RA) performance, e.g. outperforming the prior art on Cifar10, Cifar100 and ImageNet by 17\%, 19\% and 19\% on $L_{\infty}$ Autoattack. %{\color{red} To compare with recent test-time defenses~\cite{Yoon2021AdversarialPW, Alfarra_Perez_Thabet_Bibi_Torr_Ghanem_2022,rusu22a,croce22a,9710949,nie2022DiffPure}, DISCO is evaluated under the same setting as several prior works and experiments show that DISCO outperforms these test-time defenses across various datasets and attacks.} 
A comparison to recent test-time defenses~\cite{Yoon2021AdversarialPW, Alfarra_Perez_Thabet_Bibi_Torr_Ghanem_2022,rusu22a,croce22a,9710949,nie2022DiffPure} also shows that DISCO is a more effective defensive strategy across various datasets and attacks. Furthermore, a study of the defense transferability across datasets, classifiers and attacks shows that the DISCO defense maintains much of its robustness even when deployed in a setting that differs from that used for training by any combination of these three factors. Finally, the importance of the local manifold modeling is illustrated by experiments on ImageNet~\cite{deng2009imagenet}, where it is shown that even when trained with only 0.5\% of the dataset DISCO outperforms all prior defenses. 
Under the adaptive setting, DISCO is evaluated using the BPDA~\cite{bpda} attack, known to circumvent most defenses based on image transformation~\cite{pixeldefend,defensegan,stl,Xu2018FeatureSD,Dziugaite2016ASO}. While DISCO is more vulnerable under this setting, where the defense is known to the attacker, it still outperforms existing approaches by 46.76\%. More importantly, we show that the defense can be substantially strengthened by cascading DISCO stages, which magnifies the gains of DISCO to 57.77\%. This again leverages the parameter efficiency of the local modeling of image statistics, which allows the implementation of DISCO cascades with low complexity. The ability to cascade DISCO stages also allows a new type of defense, where the number of DISCO stages is randomized on a per image basis. This introduces some degree of uncertainty about the defense even under the adaptive setting and further improves robustness. 
% Since a cascade of DISCO can be used for more effective defense, a more general \textit{easy defend and hard attack} (EDHA) setting is proposed, where the attacker is aware of DISCO but not the cascade stages. Under the EDHA setting, the cascade DISCO demonstrates better robustness and becomes more computationally challenging to attack. 

Overall, this work makes four contributions. First, it proposes the use of defenses based purely on the conditional modeling of local image statistics. Second, it introduces a novel defense of this type, DISCO, based on local implicit functions. Third, it leverages the parameter efficiency of the local modeling to propose a cascaded version of DISCO that is shown robust even to adaptive attacks. 
Finally, DISCO is shown to outperform prior defenses on RobustBench~\cite{croce2021robustbench} and other 11 attacks, as well as test-time defenses under various experimental settings. Extensive ablations demonstrate that DISCO has unmatched defense transferrability in the literature, across datasets, attacks and classifiers.

%\vspace{-10pt}
\section{Related Work}
%\vspace{-5pt}
\noindent\textbf{Adversarial Attack and Defense.}
We briefly review adversarial attacks and defenses for classification and prior art related to our work. 
Please refer to~\cite{Chakraborty2018AdversarialAA, OZDAG2018152, 9614158} for more complete reviews. 

\textit{Adversarial Attacks} aim to fool the classifier by generating an imperceptible perturbation (under $L_p$ norm constraint) that is applied to the clean image. Attack methods have evolved from simple addition of sign gradient, as in FGSM~\cite{fgsm}, to more sophisticated approaches~\cite{bim,cw,onepixelattack,rfgsm,pgd,autoattack,tifgsm,mifgsm,apgd}. While most white-box attacks assume access to the classifier gradient, BPDA~\cite{bpda} proposed a gradient approximation attack that can circumvent defenses built on obfuscated gradients. In general, these attacks fall into two settings, oblivious or adaptive, depending on whether the attacker is aware of the defense strategy~\cite{Yin2020GAT,REN2020346,Xu2018FeatureSD}. DISCO is evaluated under both settings.

\textit{Adversarial Defenses} can be categorized into adversarially trained and transformation based. The former are trained against adversarial examples generated on-the-fly during training~\cite{Rebuffi2021FixingDA,NEURIPS2021_Gowal,gowal2020uncovering,rade2021helperbased,sehwag2022robust,Pang22,sehwag2022robust}, allowing the resulting robust classifier to defend against the adversarial examples. While adversarially trained defenses dominate the literature, they are bound together with the classifier. Hence, re-training is required if the classifier changes and the cost of adversarial training increases for larger classifiers. Transformation based defenses~\cite{guo2018countering, Xu2018FeatureSD,Jia2019ComDefendAE,pixeldefend,stl} instead introduce an additional defense module, which can be applied to many classifiers. This module preprocesses the input image before passing it to the classifier. The proposed preprocessing steps include JPEG compression~\cite{Das2017KeepingTB,Dziugaite2016ASO,Liu2019FeatureDD}, bit reduction~\cite{guo2018countering, Xu2018FeatureSD,Jia2019ComDefendAE}, pixel deflection~\cite{Prakash2018DeflectingAA} or applications of random transformations~\cite{xie2017mitigating,guo2018countering} and median filters~\cite{Osadchy17}. Beyond pixel space defenses, malicious images can also be reconstructed to better match natural image statistics using autoencoders~\cite{magnet,stl}, GANs~\cite{defensegan,Ali20,Yuan2020EnsembleGC}, or other generative models, such as the PixelCNN~\cite{Salimans2017PixelCNNIT}. The latter is used to project the malicious image into the natural image manifold by methods like PixelDefend~\cite{pixeldefend} or HPD~\cite{Bai2019HilbertBasedGD}. These methods can only produce images of fixed size~\cite{magnet,defensegan,stl,Yuan2020EnsembleGC} and model pixel likelihoods over the entire image~\cite{pixeldefend,Bai2019HilbertBasedGD}. 
This is unlike DISCO, which models conditional local statistics and can produce outputs of various size. 

The idea of performing adversarial purification before feeding the image into the classifier is central to a recent wave of test-time defenses \cite{Yoon2021AdversarialPW, Alfarra_Perez_Thabet_Bibi_Torr_Ghanem_2022,9710949,nie2022DiffPure}. 
\cite{Yoon2021AdversarialPW} addresses the impracticality of previous Monte-Carlo purification models by introducing a Denoising Score-Matching and a random noise injection mechanism.
\cite{Alfarra_Perez_Thabet_Bibi_Torr_Ghanem_2022} prepends an anti-adversary layer to the classifier, with the goal of maximizing the classifier confidence of the predicted label. \cite{9710949} reverses the adversarial examples using self-supervised contrastive loss. 
\cite{nie2022DiffPure} proposed a diffusion model for adversarial removal. Unlike these prior works, DISCO purifies the adversarial image by modeling the local patch statistics. Such characteristics results in data and parameter efficiency, which have not been demonstrated for \cite{Yoon2021AdversarialPW, Alfarra_Perez_Thabet_Bibi_Torr_Ghanem_2022,9710949,nie2022DiffPure}. Furthermore, DISCO outperforms all prior works in terms of robust accuracy, under the various settings they proposed.

\noindent\textbf{Implicit Function.}
refers to the use of a neural network to model a continuous function~\cite{Sitzmann2020ImplicitNR}. This has been widely used in applications involving audio~\cite{zuiderveld2021towards, guo2021adnerf,Sitzmann2020ImplicitNR}, 2D images~\cite{chen2021learning,dupont2021coin} and 3D shapes~\cite{Sitzmann2020ImplicitNR,Mildenhall2020NeRFRS, Saito2019PIFuPI,Niemeyer2020DifferentiableVR,NIPS2019_8340,semantic_srn,Jiang2020LocalIG,Chen2019LearningIF,Mescheder2019OccupancyNL,leung2022black,Park2019DeepSDFLC,Genova2020LocalDI,Wu2020IFDefense3A}. In the 3D literature, local implicit functions have become popular models of object shape~\cite{Niemeyer2020DifferentiableVR,Saito2019PIFuPI,Park2019DeepSDFLC} or complex 3D scenes~\cite{Mildenhall2020NeRFRS}. This also inspired 2D applications to super-resolution~\cite{chen2021learning}, image~\cite{Karras2021AliasFreeGA} and video generation~\cite{yu2022generating}. In the adversarial attack literature, implicit functions have recently been proposed to restore adversarial point clouds of 3D shape, through the IF-Defense~\cite{Wu2020IFDefense3A}.  To the best of our knowledge, ours is the first paper to propose local implicit functions for 2D adversarial defense.

% \begin{figure}
%     \centering
%     \includegraphics[width=\linewidth,height=0.4\linewidth]{fig/teaser.pdf}
%     \caption{(a) Data preparation, (b) training and (c) testing phase of DISCO. DISCO supports different configurations of attack and classifier for training and testing. For cascade DISCO, $K>1$.}
%     \label{fig:teaser}
%     \vspace{-10pt}
% \end{figure}

\begin{table}
\begin{minipage}{.64\linewidth}
    \begin{tabular}{c}
        \includegraphics[width=0.95\linewidth, height=0.39\linewidth]{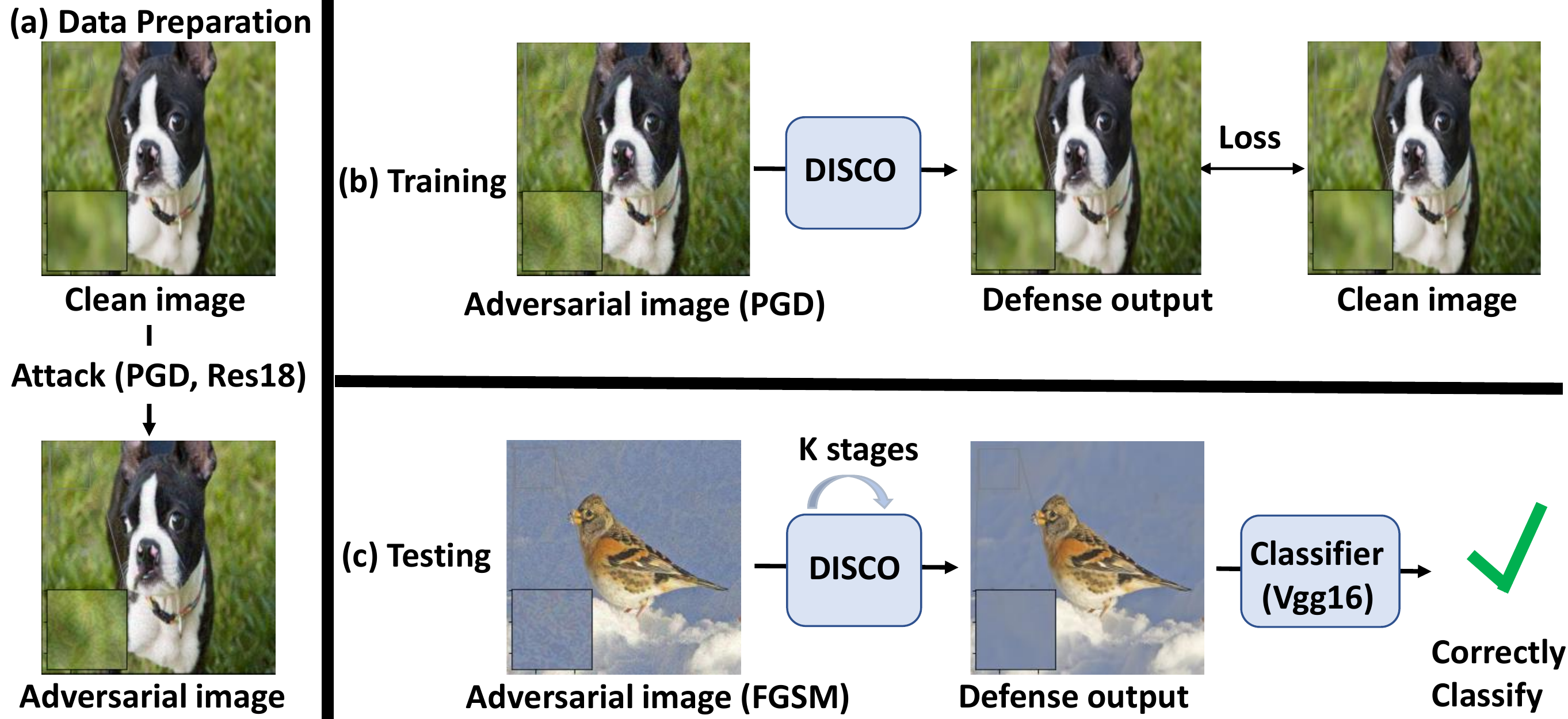}
    \end{tabular}
    \captionof{figure}{(a) Data preparation, (b) training and (c) testing phase of DISCO. DISCO supports different configurations of attack and classifier for training and testing. For cascade DISCO, $K>1$.}
    \label{fig:teaser}
\end{minipage}
\hspace{2pt}
\begin{minipage}{.28\linewidth}
    \begin{tabular}{c}
\includegraphics[width=1\linewidth,, height=0.75\linewidth ]{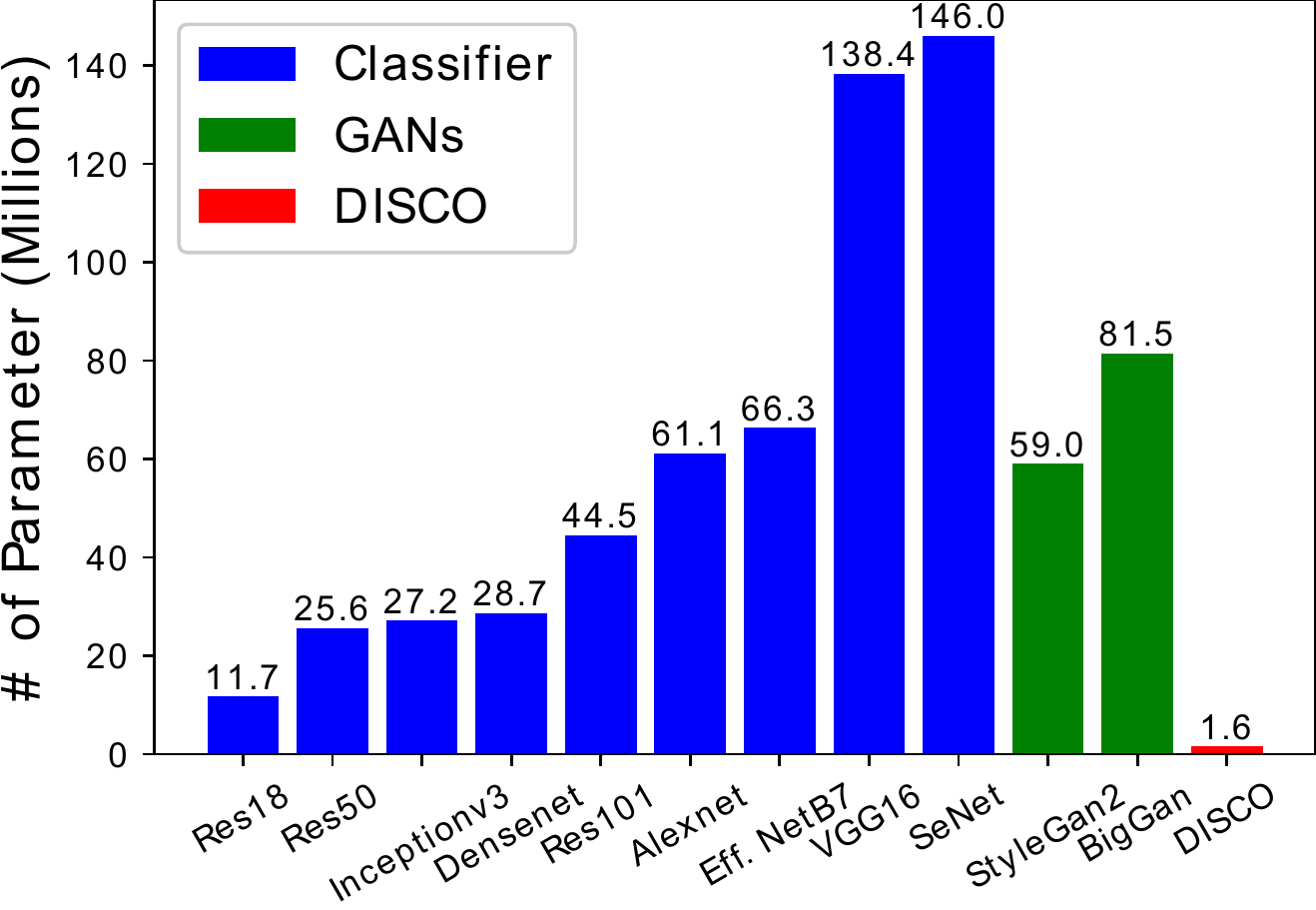}
    \end{tabular}
    \captionof{figure}{Number of parameters (Millions) of  recent classifiers, GANs and DISCO.
    }
    \label{fig:parm}
\end{minipage}
    \vspace{-20pt}
\end{table}

%\vspace{-10pt}
\section{Method}
%\vspace{-5pt}

In this section, we introduce the architecture of DISCO and its training and testing procedure.

%\vspace{-5pt}
\subsection{Motivation}
%\vspace{-5pt}
% It has long been hypothesized that vision is only possible because the natural world contains substantial statistical regularities, which are exploited by the vision system to overcome the difficulty of scene understanding~\cite{vasconcelos08,barlow,field}. Under this hypothesis, natural images form a low-dimension manifold in image space, denoted as the {\it image manifold\/}, and human vision is highly tuned to that manifold. A potential justification for the success of adversarial attacks in the machine learning literature is that these attacks create images that are just barely outside the image manifold. We refer to these images as {\it barely outliers\/}. While humans have the ability to project these images into the manifold, probably due to an history of training under adverse conditions, such as low-light environments and environments riddled with occlusions, this is not the case for current classification networks. 

Under the hypothesis that natural images lie on a low-dimension image manifold, classification networks can be robustified by learning to project barely outliers (i.e. adversarial images) into the manifold, a process that can be seen as {\it manifold thickening\/}. Images in a shell around the manifold are projected into it, leaving a larger margin to images that should be recognized as outliers. While this idea has been studied~\cite{gilmer2018adversarial,Khoury2018OnTG,NEURIPS2020_23937b42,Jha18}, its success hinges on the ability of classification models to capture the complexities of the image manifold. This is a very hard problem, as evidenced by the difficulty of model inversion algorithms that aim to synthesize images with a classifier~\cite{Wang2021IMAGINEIS,Mahendran2015UnderstandingDI,Nguyen2017PlugP,Yin2020DreamingTD}. These algorithms fail to produce images comparable to the state of the art in generative modeling, such as GANs~\cite{Karras2020AnalyzingAI,Karras2019ASG,Brock2019LargeSG}. Recently, however, it has been shown that it is possible to synthesize  realistic images and 3D shapes with implicit functions, through the use of deep networks~\cite{Genova2020LocalDI,chen2021learning,Sitzmann2020ImplicitNR,Mildenhall2020NeRFRS, Saito2019PIFuPI,Niemeyer2020DifferentiableVR,Jiang2020LocalIG,Chen2019LearningIF} that basically memorize images or objects as continuous functions. 
The assumption behind DISCO is that these implicit functions can capture the local statistics of  images or 3D shapes, and can be trained for manifold thickening, that is to learn how to 
projecting barely outliers into the image manifold.

% \subsection{Training Data Preparation}
% To train DISCO, a dataset $D=\{(x^i_{cln}, x^i_{adv})\}_{i=1}^N$ containing a set of paired clean $x^i_{cln}$ and adversarial $x^i_{adv}$ images is curated. For this, a classifier $P_{trn}$ and an attack procedure $A_{trn}$ are first selected. For each image $x^i_{cln}$, the adversarial image $x^i_{adv}$ is the generated by attacking the predictions $P_{trn}(x^i_{cln})$ using $A_{trn}$.

% , which can be used for various applications
% with classification-like architectures
% Most relevant to this work are implicit models of the local statistics of natural images, where the network learns a conditional representation of the image appearance in the neighborhood of each pixel, given a feature extracted at that pixel. This has, for example, produced impressive results for image interpolation~\cite{chen2021learning}.
% The assumption behind DISCO is that these models can be trained for manifold thickening, that is to learn how to project barely outliers into the image manifold.  

%\vspace{-5pt}
\subsection{Model Architecture and Training}
%\vspace{-5pt}
\noindent\textbf{Data Preparation}
To train DISCO, a dataset $D=\{(x^i_{cln}, x^i_{adv})\}_{i=1}^N$ containing a set of paired clean $x^i_{cln}$ and adversarial $x^i_{adv}$ images is curated. For this, a classifier $P_{trn}$ and an attack procedure $A_{trn}$ are first selected. For each image $x^i_{cln}$, the adversarial image $x^i_{adv}$ is generated by attacking the predictions $P_{trn}(x^i_{cln})$ using $A_{trn}$, as shown in Fig.~\ref{fig:teaser}(a).

\noindent\textbf{Training}
As shown in Fig.~\ref{fig:teaser}(b), the DISCO defense is trained with pairs of random patches cropped at the same location of the images $x_{cln}$ and $x_{adv}$. For example, random patches of size $48\times 48$ are sampled from training pairs of the ImageNet dataset~\cite{deng2009imagenet}. The network is then trained to minimize the L1 loss between the RGB values of the clean $x_{cln}$ and defense output $x_{def}$.

\noindent\textbf{Architecture}
DISCO performs manifold thickening by leveraging the LIIF~\cite{Chen2019LearningIF} architecture to purify adversarial patches. It takes an adversarial image $x_{adv} \in \mathbb{R}^{H\times W \times 3}$ and a query pixel location $p=[i,j] \in \mathbb{R}^2$ as input and predicts a clean RGB value $v \in \mathbb{R}^3$ at the query pixel location, ideally identical to the pixel value of the clean image $x_{cln} \in \mathbb{R}^{H\times W \times 3}$ at that location. The defense output $x_{def} \in \mathbb{R}^{H'\times W' \times 3}$  can then be synthesized by predicting a RGB value for each pixel location in a grid of size ${H'\times W' \times 3}$. Note that it is not a requirement that the size of $x_{def}$ be the same as that of $x_{cln}$. In fact, the size of $x_{def}$ could be changed during inference.

\begin{figure}
    \centering
    \includegraphics[width=\linewidth]{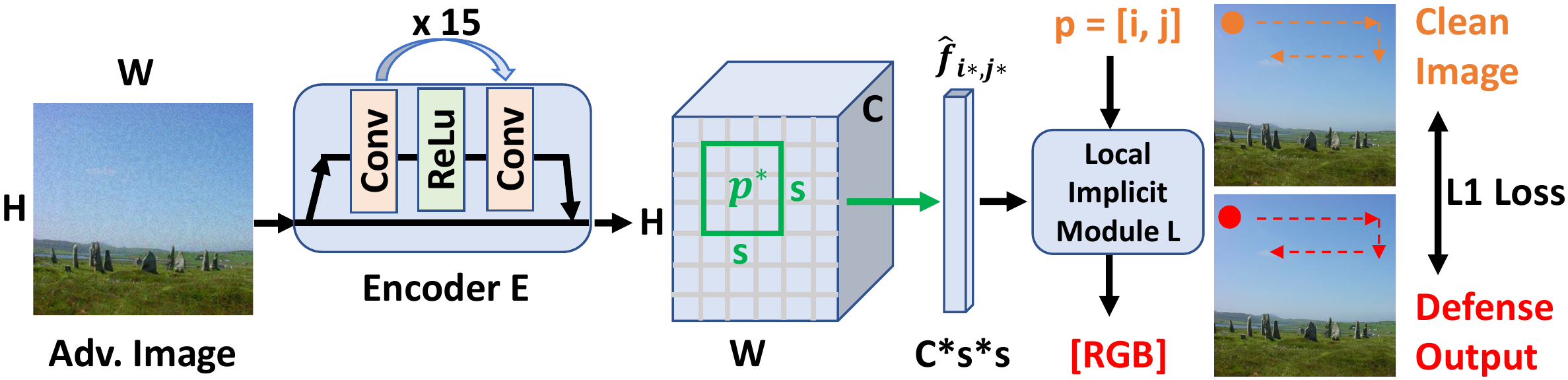}
    \caption{The DISCO architecture includes an encoder and a local implicit module. The network is trained to map Adversarial into Defense images, using an $L_1$ loss to Clean images.}
    \label{fig:main}
    \vspace{-15pt}
\end{figure}

To implement this, DISCO is composed of two modules, illustrated in Fig.~\ref{fig:main}. The first is an encoder $E$ that extracts the per-pixel feature of an input image $x$. The encoder architecture resembles the design of EDSR~\cite{edsr}, originally proposed for super-resolution. It contains a sequence of 15 residual blocks, each composed of a convolution layer, a ReLu layer and a second convolution layer. The encoder output is a feature map $f=E(x) \in \mathbb{R}^{H\times W \times C}$, with $C=64$ channels. The feature at location $p=[i,j]$ is denoted as $f_{ij}$. The second module is the the local implicit module $L$, which is implemented by a MLP. 
Given query pixel location $p$, $L$ first finds the nearest pixel value $p^*=[i^*,j^*]$ in the input image $x$ and corresponding feature $f_{i^*j^*}$. $L$ then takes the features in the neighborhood of $p^*$ into consideration to predict the clean RGB value $v$. More specifically, let $\hat{f}_{i^*j^*}$ denote a concatenation of the features whose location is within the kernel size $s$ of $p^*$. The local implicit module $L$ takes the concatenated feature, the relative position $r=p-p^*$ between $p$ and $p^*$, and the pixel shape as input, to predict a RGB value $v$. By default, the kernel size $s$ is set to be 3. Since the network implements a continuous function, based only on neighboring pixels, the original grid size $H\times W$ is not important. 
The image coordinates can be normalized so that $(i,j) \in [-1,1]^2$ and the pixel shape is the height and width of the output pixel in the normalized coordinates.
This makes DISCO independent of the original image size or resolution.
%To represent the output image $x_{def}$ from the proposed DISCO defense in a continuous space and allow generating various $x_{def}$ size during inference,  is considered. 
% I will include more details, but from high level yes
% leaving to grab some food. Me too L-) ^_^
% {\color{red} mentioned how the query details for any output size} --> fixed coord normalized betwee [-1, 1]
% The mapping $L:\mathbb{R}^{C^*s^*s+2} \rightarrow \mathbb{R}^{3}$, is implemented by an MLP. 

\subsection{Inference}
%\vspace{-5pt}
For inference, DISCO takes either a clean or an adversarial image as input. Given a specified output size for $x_{def}$, DISCO loops over all the output pixel locations, predicting an RGB value per location. Note that this is not computationally intensive because the encoder feature map $f=E(x)$ is computed once and used to the predict the RGB values of all query pixel locations. Furthermore, while the training pairs are generated with classifier $P_{trn}$ and attack $A_{trn}$, the inference time classifier $P_{tst}$ and attack $A_{tst}$ could be different. In the experimental section we show that DISCO is quite flexible, performing well when  (1) $P_{trn}$ and $P_{tst}$ consume images of different input size and (2) the attack, classifier and dataset used for inference are different than those used for training. In fact, DISCO is shown to be more robust to these configuration changes than previous methods.

\subsection{DISCO Cascades}\label{sec:edha}
%\vspace{-5pt}
DISCO is computationally very appealing because it disentangles the training of the defense from that of the classifier. This can be a big practical advantage, since classifier retraining is needed whenever training settings, such as architecture, hyper-parameters, or number of classes, change. Adversarial defenses require retraining on the entire dataset when this is the case, which is particularly expensive for large models (like SENet~\cite{senet} or EfficientNet~\cite{efficientnet}) trained on large datasets (like ImageNet~\cite{deng2009imagenet} and OpenImages~\cite{openimages}). Unsurprisingly,  RobustBench~\cite{croce2021robustbench}, one of the largest adversarial learning benchmarks, reports more than 70 baselines for Cifar10, but less than 5 on ImageNet. 

DISCO does not have this defense complexity, since it is trained independently of the classifier. Furthermore, because DISCO is a model of local statistics, it is particularly parameter efficient. As shown in Fig.~\ref{fig:parm}, DISCO has a lightweight design with only 1.6M parameters, which is significantly less than most recent classifier~\cite{efficientnet,senet,vgg,He2016DeepRL} and GAN~\cite{Karras2020AnalyzingAI,biggan} models with good performance for ImageNet-like images.
This also leads to a computationally efficient defense. Our experiments show that DISCO can be trained with only 50,000 training pairs. In fact, we show that it can beat the prior SOTA using less than 0.5\% of ImageNet as training data (Table~\ref{tab:dataset_size}). One major benefit of this efficiency is that it creates a {\it large unbalance between the costs of defense and attack.\/}  Consider memory usage, which is dominated by the computation of gradients needed for either the attack or the backpropagation of training. Let $N_d$ and $N_c$ be the number of parameters of the DISCO network and classifier, respectively. The per image memory cost of training the DISCO defense is $O(N_d)$. On the other hand, the attack cost depends on the information available to the attacker. We consider two settings, commonly considered in the literature. In the {\it oblivious\/} setting, only the classifier is known to that attacker and the attack has cost $O(N_c)$. In the {\it adaptive\/} setting, both the classifier and the DISCO are exposed and backpropagation has memory cost $O(N_c + N_d)$. In experiments, we show that DISCO is quite effective against oblivious attacks. Adaptive attacks are more challenging. However, as shown in Fig~\ref{fig:parm}, it is usually the case that $N_c > N_d$, making the complexity of the attack larger than that of the defense. This is unlike adversarial training, where attack and defense require backpropagation on the same model and thus have the same per-image cost. 

This asymmetry between the memory cost of the attack and defense under DISCO can be {\it magnified\/} by cascading DISCO networks. If $K$ identical stages of DISCO are cascaded, the defense complexity remains $O(N_d)$ but that of the attack raises to  $O(N_c + K N_d)$. Hence the ratio of attack-to-defense cost raises to $O(K + N_c/N_d)$. Interestingly, our experiments (see Section~\ref{sec:adaptive}) show that when $K$ is increased the defense performance of the DISCO cascade increases as well. Hence, DISCO cascades combine high robust accuracy with a large ratio of attack-to-defense cost.

%This means that, by increasing $K$, it should be possible to create an {\it Easy to Defend but Hard to Attack\/} (EDHA) setting. Eventually, by adding enough DISCO stages, it should be possible to make adaptive attacks practically unfeasible while maintaining the cost of the defense. In the experime    nt, we show that the time to compute and adversarial image is much greater than that of defense under the same $K$ and the gap significantly increases with larger $K$.
%In Section~\ref{sec:...} we show that, under our computational set-up (GPU BLAHBLAH BLAH, XXX MEMORY) this holds for small values of $K$. 
%We observe that for cascades of size XXX, DISCO achieves RA equivalent to that of the oblivious adversary. To the best of our knowledge, this is the first time that such an observation has been made. 

%Ideally, a defense should be unbreakable with a practical amount of computation, memory, or time. This is unfeasible for adversarially trained defenses, since it would also require an untrainable model. It is, however, made possible by the  significant 

%In both cases, the bulk of the memory consumption is in the computation of gradients needed for either the attack or backpropagation training. For DISCO, however, there is a . This is because the defense network can be trained independently of the classifier.  

%\vspace{-10pt}
\section{Experiments}
%\vspace{-10pt}
In this section, we discuss experiments performed to evaluate the robustness of DISCO.
%, including the procedure to curate the training pairs for DISCO, the dataset and benchmark. Ablation studies on defense transferability across datasets, classifiers, and attacks are also performed. 
Results are discussed for both the oblivious and adaptive settings~\cite{Yin2020GAT,REN2020346,Xu2018FeatureSD} and each result is averaged over 3 trials. $\epsilon_{p}$ denotes the perturbation magnitude under the $L_p$ norm. All experiments are conducted on a single Nvidia Titan Xp GPU with Intel Xeon CPU E5-2630 using Pytorch~\cite{pytorch}. Please see appendix for more training details, quantitative results and visualizations. We adopt the code from LIIF~\cite{chen2021learning} for implementation.

\noindent\textbf{Training Dataset:} The following training configuration is used unless noted. Three datasets are considered: Cifar10~\cite{cifar10}, Cifar100~\cite{cifar100} and Imagenet~\cite{deng2009imagenet}. For each, 50,000 adversarial-clean training pairs are curated. For Cifar10 and Cifar100, these are the images in the training set, while for ImageNet, 50 images are randomly sampled per class. Following RobustBench~\cite{croce2021robustbench}, the evaluation is then conducted on the test set of each dataset. To create training pairs, PGD~\cite{pgd} ($\epsilon_{\infty}=8/255$ with step size is 2/255 and the number of steps 100) is used to attack a ResNet18 and a WideResNet28 on Cifar10/ImageNet and Cifar100, respectively.

\noindent\textbf{Attack and Benchmark:} DISCO is evaluated on RobustBench~\cite{croce2021robustbench}, which contains more than 100 baselines evaluated using Autoattack~\cite{autoattack}. This is an ensemble of four sequential attacks, including the PGD~\cite{pgd} attack with two different optimization losses, the FAB attack~\cite{fab} and the black-box Square Attack~\cite{squareattack}. DISCO is compared to defense baselines under both $L_{\infty}$ and $L_{2}$ norms. To study defense generalization, 11 additional attacks are considered, including FGSM~\cite{fgsm}, BIM~\cite{bim}, BPDA~\cite{bpda} and  EotPgd~\cite{eotpgd}. Note that DISCO is not trained specifically for these attacks.

\noindent\textbf{Metric:}
Standard (\textbf{SA}) and robust  (\textbf{RA}) accuracy are considered. The former measures classification accuracy on clean examples, the latter on adversarial. The average of SA and RA is also used.

\begin{table}
\begin{minipage}{.5\linewidth}
\captionof{table}{Compare DISCO to the selected baselines on Cifar10 ($\epsilon_{\infty}=8/255$).}
\label{tab:cifar10_robustbench_selected}
\adjustbox{max width=\linewidth}{%
    \begin{tabular}{|c|cccc|}
      \hline 
      Method &  SA & RA & Avg. &  Classifier\\
      \hline
      \hline
      No Defense  &  \textbf{94.78} & 0 & 47.39 & WRN28-10 \\
      \hline
      \hline
      Rebuffi et al.~\cite{Rebuffi2021FixingDA} & 92.23 & 66.58 & 79.41 & WRN70-16 \\ 
      Gowal et al.~\cite{NEURIPS2021_Gowal} &  88.74 & 66.11 & 77.43 & WRN70-16 \\
      Gowal et al.~\cite{NEURIPS2021_Gowal} & 87.5 & 63.44 & 75.47 & WRN28-10 \\
      \hline
      \hline
      Bit Reduction~\cite{Xu2018FeatureSD} & 92.66 & 1.04 & 46.85 & WRN28-10 \\ 
      Jpeg~\cite{Dziugaite2016ASO}  &  83.9 & 50.73 & 67.32 & WRN28-10 \\
      %\hline
      Input Rand.~\cite{xie2017mitigating} & 94.3 & 8.59 & 51.45 & WRN28-10 \\ 
      %\hline
      AutoEncoder &  76.54 & 67.41 & 71.98 & WRN28-10 \\
      %\hline 
      STL~\cite{stl} &  82.22 & 67.92 & 75.07 & WRN28-10 \\
      \hline
      \hline
      DISCO & 89.26 & \textbf{85.56} & \textbf{87.41} & WRN28-10 \\
      \hline
\end{tabular}
}
\end{minipage}%
\hspace{.5pt}
\begin{minipage}{.5\linewidth}
\captionof{table}{Compare DISCO to the selected baselines on Cifar10 ($\epsilon_{2}=0.5$).}
\label{tab:cifar10_robustbench_selected_l2}
\adjustbox{max width=\linewidth}{%
    \begin{tabular}{|c|cccc|}
      \hline 
      Method &  SA & RA & Avg. &  Classifier\\
      \hline
      \hline
      No Defense  &  \textbf{94.78} & 0 & 47.39 & WRN28-10 \\
      \hline
      \hline
      Rebuffi et al.~\cite{Rebuffi2021FixingDA} &  95.74 & 82.32 & \textbf{89.03} & WRN70-16 \\
      Gowal et al.~\cite{gowal2020uncovering} &  94.74 & 80.53 & 87.64 & WRN70-16\\
      Rebuffi et al.~\cite{Rebuffi2021FixingDA} &  91.79 & 78.8 & 85.30 & WRN28-10 \\
      \hline
      \hline
      Bit Reduction~\cite{Xu2018FeatureSD} & 92.66 & 3.8 & 48.23 & WRN28-10 \\ 
      Jpeg~\cite{Dziugaite2016ASO}  &  83.9  & 69.85 & 76.88 & WRN28-10 \\
      %\hline
      Input Rand.~\cite{xie2017mitigating} & 94.3 &  25.71 & 60.01 & WRN28-10 \\ 
      %\hline
      AutoEncoder &  76.54 & 71.71 & 74.13 & WRN28-10 \\
      %\hline 
      STL~\cite{stl} &  82.22  & 74.33 & 78.28 & WRN28-10 \\
      \hline
      \hline
      DISCO & 89.26 & \textbf{88.47} & 88.87 & WRN28-10 \\
      \hline
\end{tabular}
}
\end{minipage}
%\vspace{-10pt}
\end{table}

\begin{figure}
 \begin{tabular}{ccc}
    \includegraphics[width=0.31\linewidth]{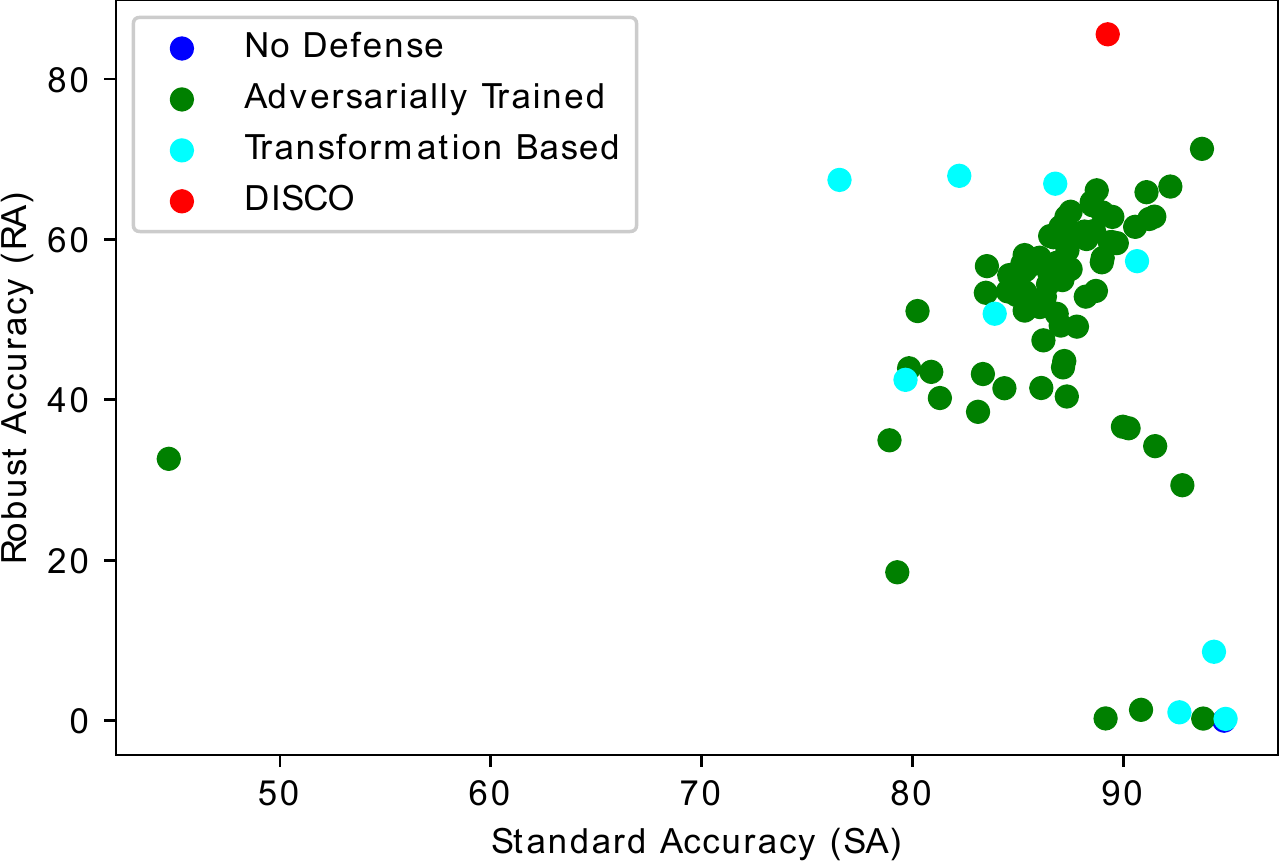}  & 
    \includegraphics[width=0.31\linewidth]{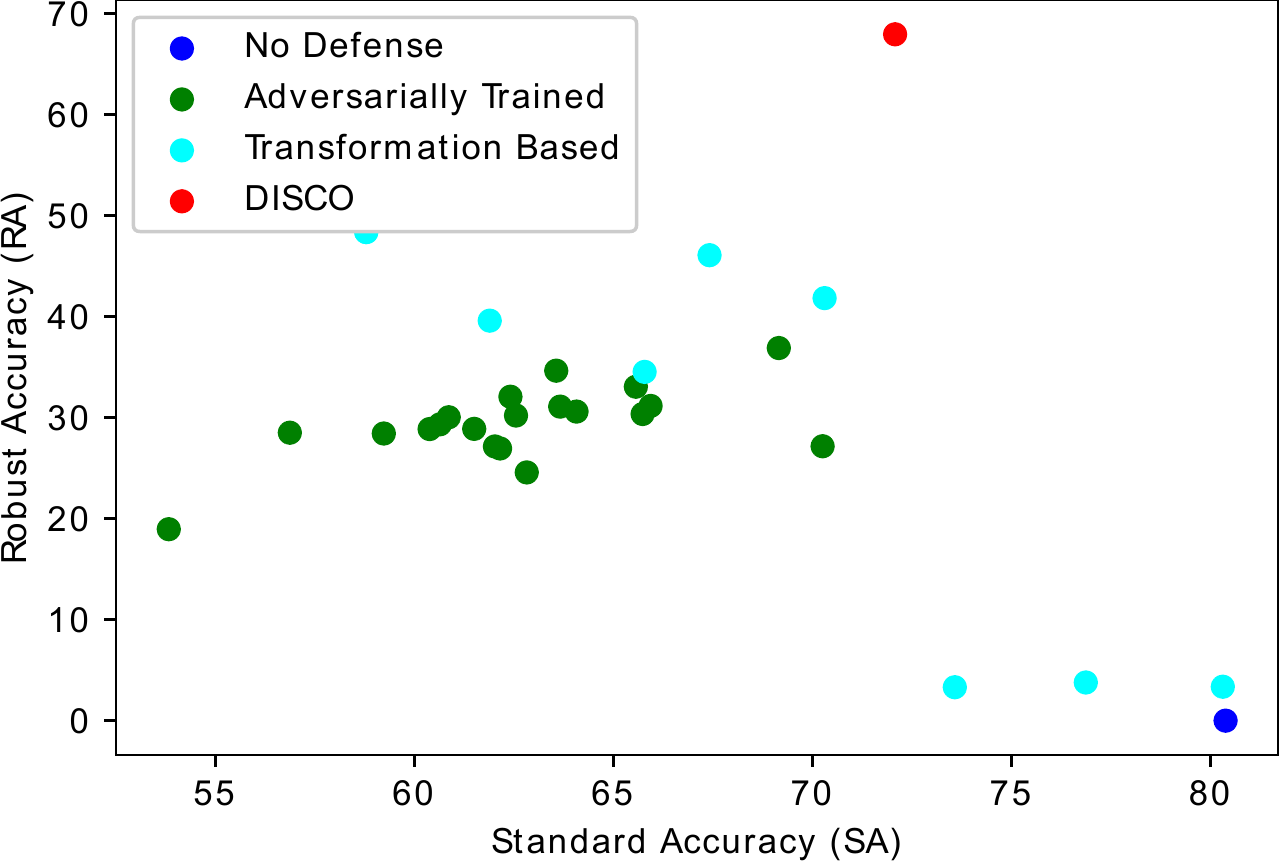} & 
    \includegraphics[width=0.31\linewidth]{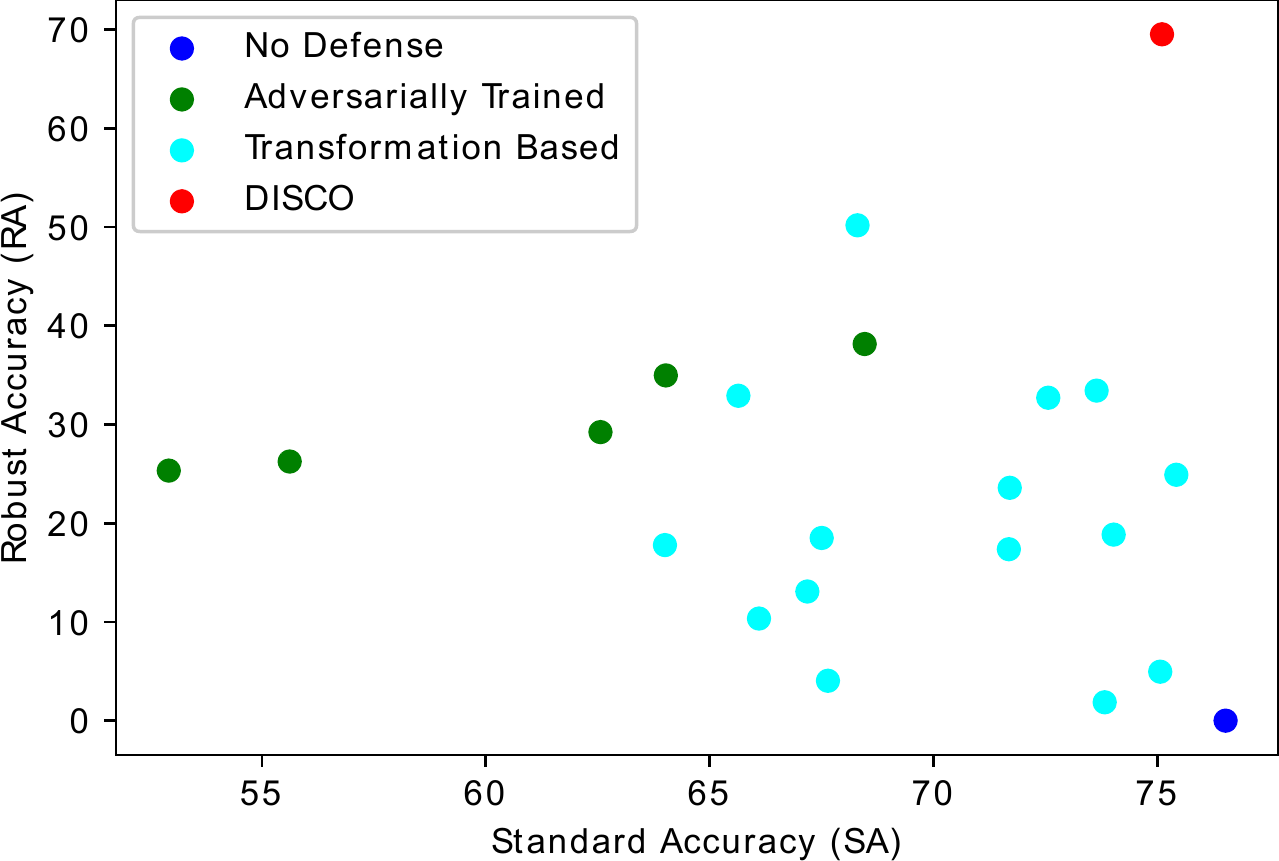} \\
    \includegraphics[width=0.31\linewidth]{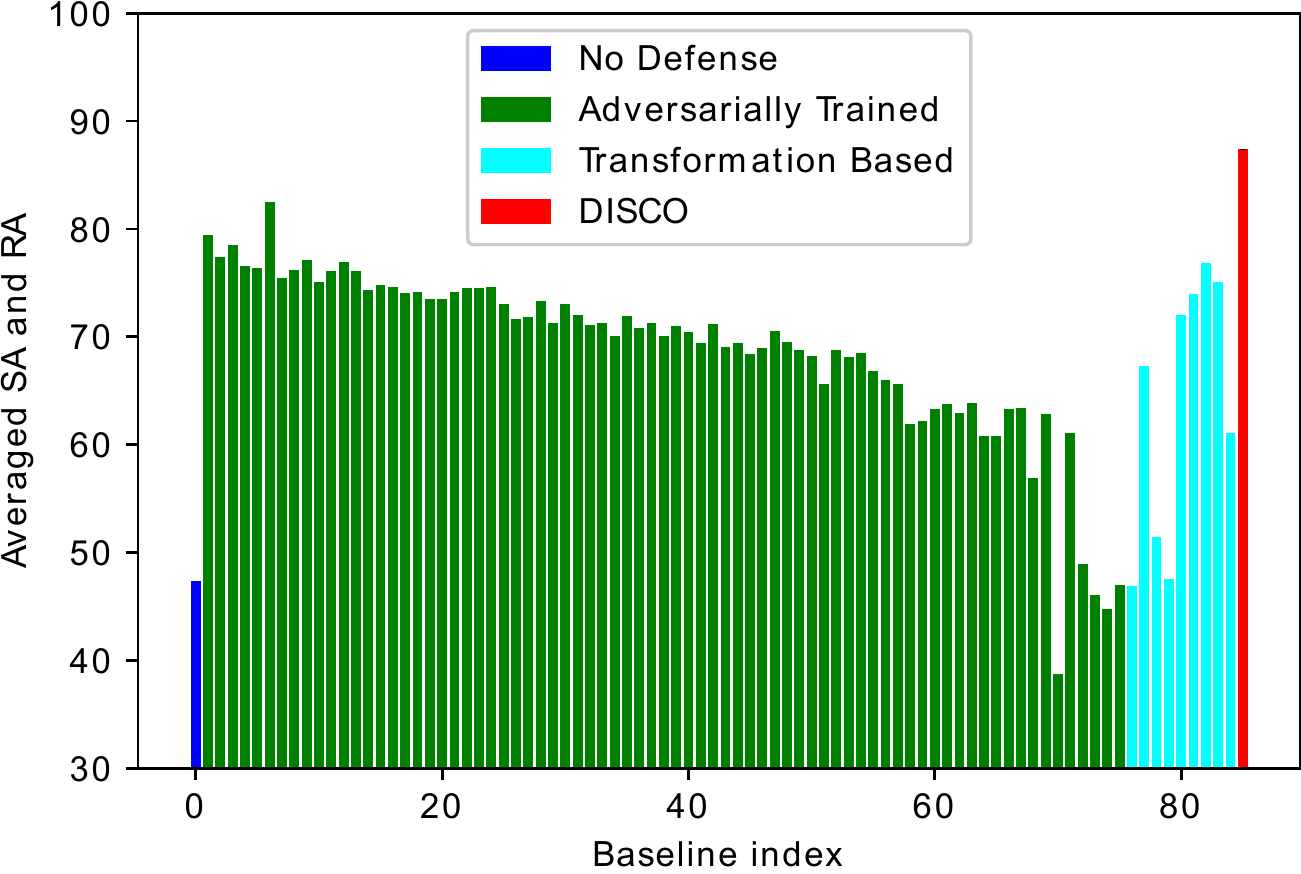}  & 
    \includegraphics[width=0.31\linewidth]{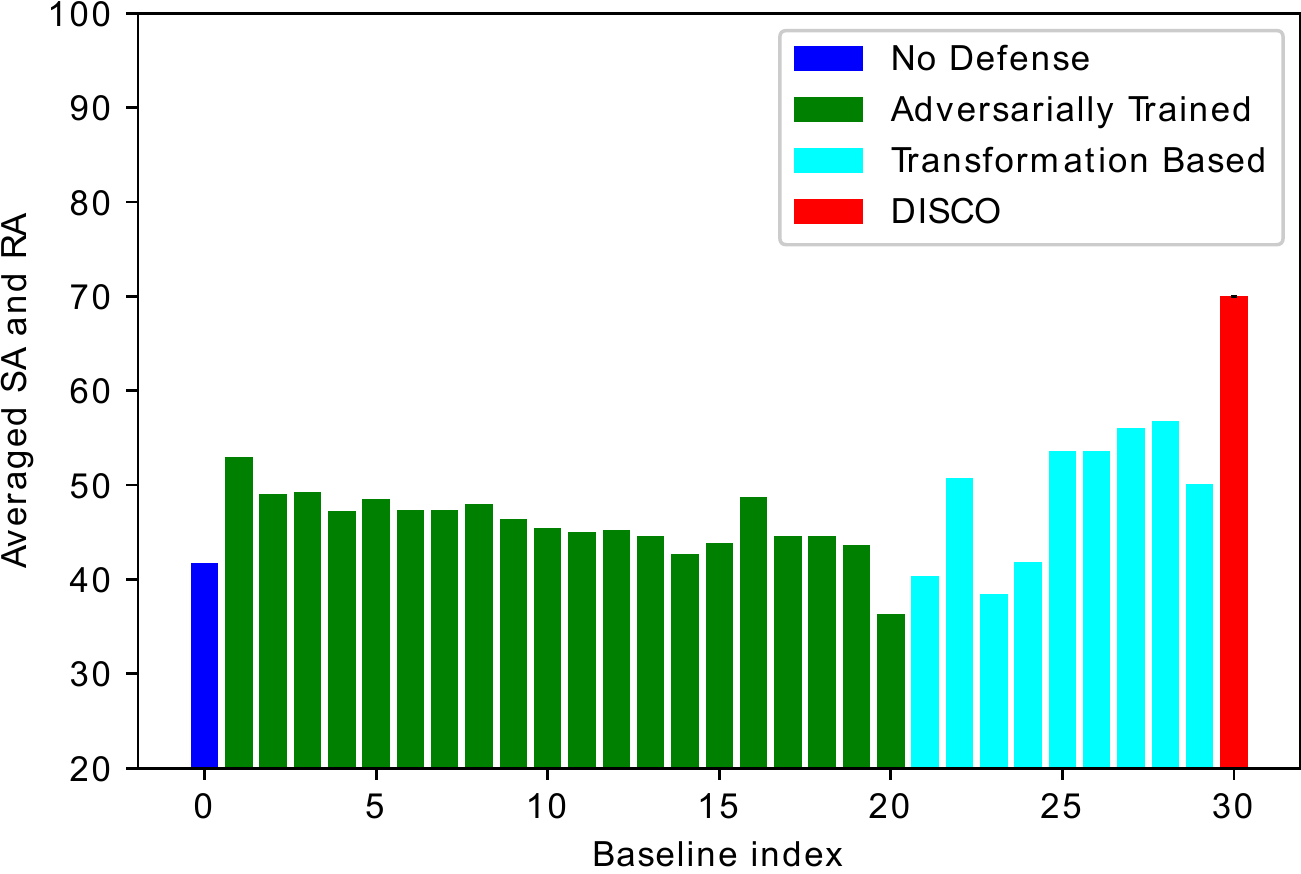} & 
    \includegraphics[width=0.31\linewidth]{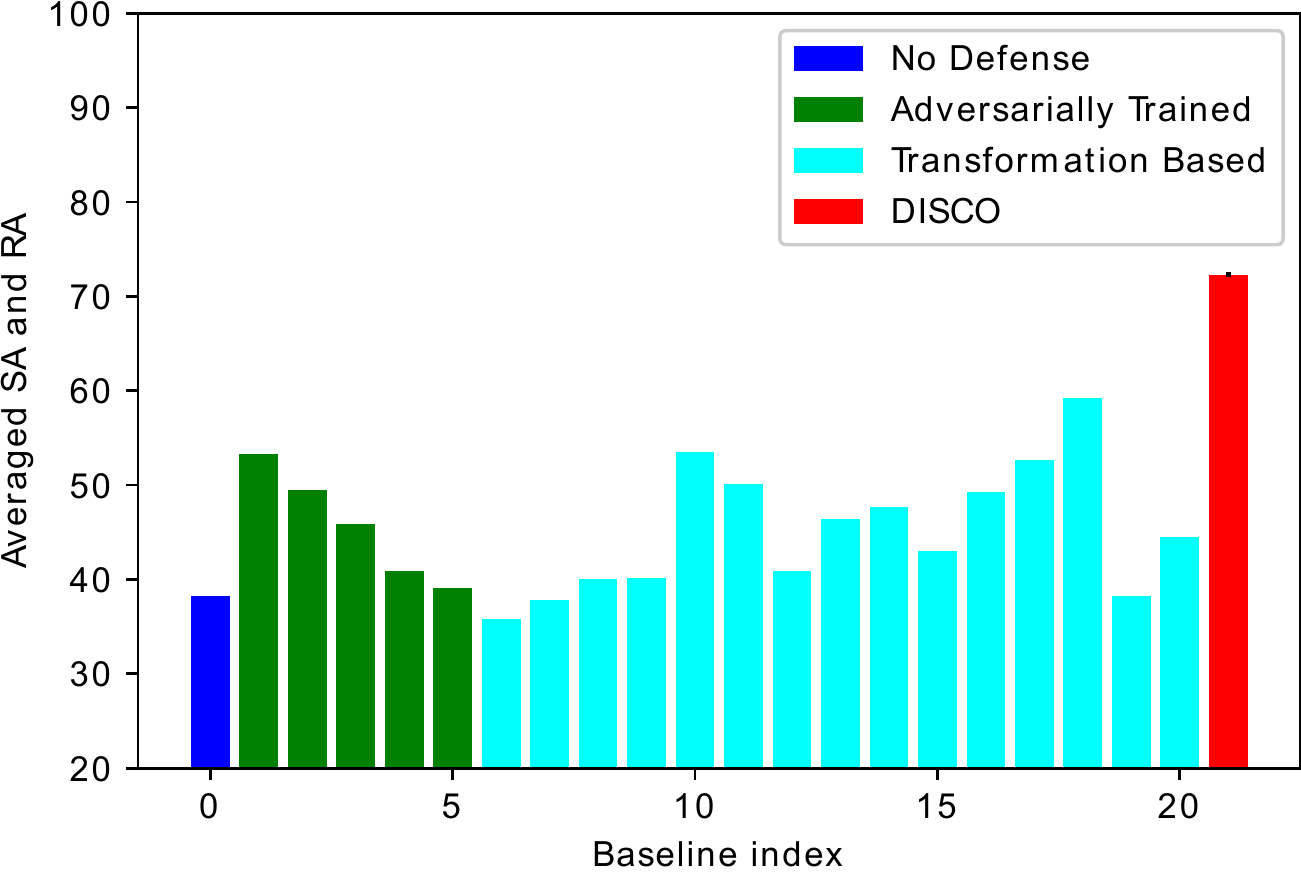} \\
    \scriptsize{(a)} & \scriptsize{(b)} & \scriptsize{(c)} \\
 \end{tabular}
 %\vspace{-7pt}
\caption{Comparison of DISCO to No Defense, Adversarially Trained, and Transformation based baselines.  (a) Cifar10, (b) Cifar100, and (c) ImageNet. Top-row: trade-off between SA and RA. Bottom row: average accuracy of each of the
% {\color{red} WHAT? RobustBench?}
RobustBench baselines and DISCO. %{\color{red} METHOD IS CALLED PURIFIER}
}
\label{fig:performance_plot}
\vspace{-15pt}
\end{figure}

\subsection{Oblivious Adversary}
%\vspace{-5pt}
% The oblivious adversary assumes the attacker has full knowledge of the classifier but not the defense.

\noindent\textbf{SOTA on RobustBench:}
DISCO achieves SOTA performance on RobustBench.  Table~\ref{tab:cifar10_robustbench_selected} and \ref{tab:cifar10_robustbench_selected_l2} compare DISCO to the RobustBench baselines on Cifar10 under $L_{\infty}$ and $L_2$ Autoattack, respectively. Baselines are categorized into (1) no defense (first block), (2) adversarially trained (second block) and (3) transformation based (third block). The methods presented in each table are those of highest RA performance in each category. The full table is given in the supplemental, together with those of Cifar100 and ImageNet. Note that the RobustBench comparison slightly favours the adversarially trained methods, which use a larger classifier. A detailed comparison to all RobustBench baselines is given in Fig.~\ref{fig:performance_plot}, for the three datasets. The upper row  visualizes the  trade-off between SA and RA. The bottom row plots the averaged SA/RA across baselines. Blue, green, cyan and red indicate no defense, adversarially trained baselines, transformation based baselines and DISCO, respectively.
% RobustBench shows adversarially trained methods with different classifier. However, classifier with larger network tends to have better performance. For example, for the REBUFFI baseline, it also have another version on smaller network, but the performance decreases, which is understandable. 

% {\color{red} On TABLE 2, REBUFFI HAS HIGHER AVG}.
% fixed. note that REBUFFI use WRN70 while we use WRN28. RobustBench just mix baselines with different classifier together so I list all of them
%BUt you cant say it is better if the reasult is weaker! And why bold then?
% this is a mistake. I just fixed it OK that is all I was saying.
% the full table is quite large if you see the appendix, I will double check everything before submission

These results show that, without a defense, the attack fools the classifier on nearly all examples. Adversarially trained baselines improve RA by training against the adversarial examples. Some of these~\cite{Rebuffi2021FixingDA, NEURIPS2021_Gowal, gowal2020uncovering, kang2021stable, rade2021helperbased, huang2021exploring, wu2021do, Sridhar2021RobustLV, wu2020adversarial} also leverage additional training data. Transformation based defenses require no modification of the pre-trained classifier and can generalize across attack strategies~\cite{defensegan, stl}. While early methods (like Jpeg Compression~\cite{Dziugaite2016ASO} and Bit  Reduction~\cite{Xu2018FeatureSD}) are not competitive, recent defenses~\cite{stl} outperform adversarially trained baselines on Cifar100 and ImageNet. DISCO is an even more powerful transformation-based defense, which clearly outperforms the prior SOTA RA by a large margin (17 \% on Cifar10, 19 \% on Cifar100 and 19 \% on ImageNet). In the upper row of Fig.~\ref{fig:performance_plot}, it clearly beats the prior Pareto front for the SA vs. RA trade-off.
% see appendix I move some table to appendix and keep plots for cifar100, imagenet
 Table~\ref{tab:cifar10_robustbench_selected} and Table~\ref{tab:cifar10_robustbench_selected_l2} also show that previous transformation based methods tend to perform better for $L_2$ than $L_{\infty}$ Autoattack. DISCO is more robust, achieving similar RA for $L_2$ and $L_{\infty}$ Autoattacks.

% {\color{red} WHY ARE THERE THREE or TWO RED METHODS for CIFAR100 and IMAGENET? redo plots keep the best model}
% {\color{red} redo plots keep the best model}
% {\color{red} I evalauted my method on multiple classifiers for CIFAR100 and ImageNet. Should I keep the best only? See table 3 and table 4 in appendix YES SHOW THE BEST ONLY. THERE IS NOT NEED TO MAKE IT MORE COMPLEX WITH DATA THAT DOES NOT ADD MUCH. AS IT IS YOU WOULD HAVE  TO EXPLAIN. OK will redo the plot}

\begin{table}
\centering
\begin{minipage}{.30\linewidth}
\centering
 \captionof{table}{Improving ResNet50 baselines on ImageNet.}
\label{tab:plug_and_play}
\adjustbox{max width=1\linewidth}{%
\begin{tabular}{|c|ccc|}
  \hline
  Method & SA & RA & Avg. \\
  \hline
  Hadi et al.~\cite{NEURIPS2020_24357dd0}   & \textbf{64.02} & 34.96 & 49.49 \\
  w/ DISCO & 63.66 & \textbf{50.6} & \textbf{57.13}\\
  \hline
  Engstrom et al.~\cite{robustness}   &  \textbf{62.56} & 29.22 & 45.89 \\
  w/ DISCO  & 62.48 & \textbf{49.44} & \textbf{55.96}\\
  \hline
  Wong et al.~\cite{Wong2020Fast} & \textbf{55.62} & 26.24 & 40.93 \\
  w/ DISCO & 54.52 & \textbf{40.68} & \textbf{47.6} \\
  \hline
\end{tabular}
  }
    \captionof{table}{Ablation of sampled classes.}
    \vspace{-8pt}
    \label{tab:dataset_size}
    \adjustbox{max width=1\linewidth}{%
    \begin{tabular}{|cc|ccc|}
    \hline
     Cls. \# & Dataset size & SA & RA & Avg.\\
     \hline
     100  & 5000 & \textbf{72.78} & 59.84 & 66.31\\
     500  & 25000 & 71.88 & 67.84 & 69.86\\
     1000  & 50000 & 72.64 & \textbf{68.2} & \textbf{70.42}\\
     \hline
    \end{tabular}
    }
\end{minipage}
\hspace{5pt}
\begin{minipage}{.67\linewidth}
    \centering
    \caption{Defense Transfer of $L_{\infty}$ trained defenses to $L_{2}$ attacks on Cifar10. Top block: adversarially trained, Bottom block: transformation based. }
    \label{tab:defense_transfer_l2}
    \adjustbox{max width=1\linewidth}{%
    \begin{tabular}{|cc|ccccc|}
    \hline
      Method & Classifier & Clean & FGSM & BIM & CW  & DeepFool\\
      \hline
      Adv. FGSM & ResNet & 91 & \textbf{91} & \textbf{91} & 7 & 0\\
      Adv. BIM & ResNet & 87 & 52 & 32 & \textbf{42} & \textbf{48}\\
      \hline
      \hline
      PixelDefend~\cite{pixeldefend} & VGG & 85 & 46 & 46 & 78 & 80\\
      PixelDefend~\cite{pixeldefend} & ResNet & 82 & 62 & 61 & 79 & 76\\
      PixelDefend (Adv.)~\cite{pixeldefend} & ResNet & 88 & 68 & 69 & 84 & 85\\
      Feature Squeezing~\cite{Xu2018FeatureSD} & ResNet & 84 & 20 & 0 & 78 & N/A\\
      EGC-FL~\cite{Yuan2020EnsembleGC} & ResNet & 91.65 & 88.51 & 88.75 & \textbf{90.03} & N/A\\
      STL~\cite{stl} & VGG16 & 90.11 & 87.15 & 88.03 & 89.04 & 88.9\\
      \hline
     DISCO & WRN28 & 89.26 & \textbf{89.53} & \textbf{89.58} & 89.3 & \textbf{89.58} \\
     \hline
    \end{tabular}
    }
\end{minipage}
    \vspace{-20pt}
\end{table}

\textbf{Improving SOTA Methods:} While DISCO outperforms the SOTA methods on RobustBench, it can also be combined with  the latter. Table~\ref{tab:plug_and_play} shows that adding DISCO improves the performance of top three ResNet50 robust baselines for ImageNet~\cite{NEURIPS2020_24357dd0, robustness, Wong2020Fast} by 16.77 (for RA) and 8.12 (for averaged SA/RA) on average. This demonstrates the plug-and-play ability of DISCO.

\textbf{Comparison to Test-Time Defenses} 
% Similar to DISCO, \cite{nie2022DiffPure} also compares with SOTA defenses in RobustBench. When Cifar10 and WRN28-10 classifier is considered, \cite{nie2022DiffPure} achieves 70.64/78.58 RA under  $\epsilon_\infty=8/255$ and $\epsilon_2=0.5$ respectively, while DISCO achieves 85.56/88.47  (Table 1 \& Table 2). When ImageNet is considered, \cite{nie2022DiffPure} achieves 40.93/44.39 RA with ResNet50/WRN50, while DISCO achieves 68.2/69.5 (Appendix Table D)
% When APgd~\cite{apgd} attack is considered, \cite{Alfarra_Perez_Thabet_Bibi_Torr_Ghanem_2022} achieves 80.65/47.63 RA on Cifar10/Cifar100 dataset, while DISCO achieves 85.79/77.33 (Appendix Table E \& Table 3). 
% When PGD40 attack ($\epsilon=8/255$) and the setup in \cite{Yoon2021AdversarialPW} are considered , DISCO is evaluated on Cifar10 using WRN28-10. While \cite{Yoon2021AdversarialPW} reported RA of 80.24 using the default setting, DISCO achieves RA of 80.80. Note that DISCO is not optimized for this experiment. With similar RA, DISCO has much less parameter than \cite{Yoon2021AdversarialPW} (1.6M vs 29.7M).
% When AutoAtack is considered, \cite{Alfarra_Perez_Thabet_Bibi_Torr_Ghanem_2022} achieves 79.21/40.68 RA and \cite{9710949} achieves 67.79/33.16 RA on Cifar10/Cifar100 dataset, while DISCO achieves 85.56/67.93 (Table 1 \& Appendix Table C). 
First, we compare DISCO to four recent test-time defenses. Following the setup of~\cite{Yoon2021AdversarialPW}, DISCO is evaluated on Cifar10 using a WRN28-10 network under the PGD40 attack ($\epsilon=8/255$). While \cite{Yoon2021AdversarialPW} reported an RA of 80.24 for the default setting, DISCO achieves 80.80, even though it is not optimized for this experiment and has much fewer parameters (1.6M vs 29.7M). Second, a comparison to \cite{Alfarra_Perez_Thabet_Bibi_Torr_Ghanem_2022,9710949} under Autoattack, shows that  \cite{Alfarra_Perez_Thabet_Bibi_Torr_Ghanem_2022} achieves RAs of 79.21/40.68 and \cite{9710949} of 67.79/33.16 on the Cifar10/Cifar100 datasets. These numbers are much lower than those reported for DISCO (85.56/67.93) on Table 1 \& Appendix Table C. 
Third, under the APgd~\cite{apgd} attack, \cite{Alfarra_Perez_Thabet_Bibi_Torr_Ghanem_2022} achieves 80.65/47.63 RA on Cifar10/Cifar100 dataset, while DISCO achieves 85.79/77.33 (Appendix Table E \& Table 3). This shows that DISCO clearly outperforms \cite{Alfarra_Perez_Thabet_Bibi_Torr_Ghanem_2022} on two different attacks and datasets. Finally, like DISCO, \cite{nie2022DiffPure} compares to defenses in RobustBench. For Cifar10 and a WRN28-10 classifier, \cite{nie2022DiffPure} achieves 70.64/78.58 RA under  $\epsilon_\infty=8/255$ and $\epsilon_2=0.5$ respectively, while DISCO achieves 85.56/88.47  (Table 1 \& Table 2). On ImageNet, \cite{nie2022DiffPure} achieves 40.93/44.39 RA with ResNet50/WRN50, while DISCO achieves 68.2/69.5 (Appendix Table D). In summary, DISCO outperforms all these approaches in the various settings they considered, frequently achieving large gains in RA.

\noindent\textbf{Dataset Size:}
%While most defense literature focuses on the improvement of robustness, less attention has been placed on the simplicity of using the defense strategy and the data usage efficiency. In the real world application, a classifier is subject to be changed if any of training configuration is different, including the selection of different architecture, hyper-parameter and the increase/decrease of object classes. Unlike the transformation based baselines, adversarially trained baselines requires re-training if any of the training configuration is modified. While this problem could be minor on the relatively small classifier trained on small dataset like Cifar10, it could be a concern when training robust model on large dataset (like ImageNet~\cite{deng2009imagenet} and OpenImages~\cite{openimages}) using large classifier (like SENet~\cite{senet} and EfficientNet~\cite{efficientnet}). This also reflects the fact on RobustBench that there are more than 70 reported baselines on Cifar10, but less than 5 baselines on ImageNet. Take ImageNet for example. DISCO is trained on only 50000 training pairs, while the prior robust model~\cite{NEURIPS2020_24357dd0, robustness, Wong2020Fast} is trained on the entire ImageNet, which contains 1,281,167 images. To further demonstrates the efficiency of DISCO, 
Table~\ref{tab:dataset_size} shows the SA and RA performance of DISCO when training pairs are sampled from a random subset of the ImageNet classes (100 and 500). Compared to the ImageNet SOTA~\cite{NEURIPS2020_24357dd0} RA of 38.14\% (See Appendix), DISCO outperforms the prior art by 21.7\% (59.84 vs 38.14) when trained on about 0.5\% of the ImageNet training data.

\noindent\textbf{Defense Transferability} The transferability of the DISCO defense is investigated  across attacks, classifiers and datasets.

\noindent\textbf{\textit{Transfer across Attacks.}} RobustBench evaluates the model on Autoattack, which includes the PGD attack used to train DISCO. Table~\ref{tab:cifar100_defense_transfer_linf}  summarizes the  transfer performance of DISCO, trained with PGD attacks, when subject to ten different $L_{\infty}$ attacks at inference. This is compared to the transfer peformance of the two top Cifar100 baselines on RobustBench.
% For cifar100_defense_transfer_linf, I just pick the top 2 baseline on robustbench and test them on the other 10 attack. I am not sure how they are trained (using what kind of attack for adversarial training).
% what do u mean by what they are? like the name of the attack ? yes or the citations. No? Table 3 has the citation and the name of the attacks. Do you mean write them in the text? yes this is correct. This pargraph fine? yes
DISCO outperforms these baselines with an average RA gain greater than 24.49\%. When compared to the baseline that uses the same classifier (WRN28-10), this gain increases to 29.5\%. Fig.~\ref{fig:defense_transfer_linf} visualizes the gains of DISCO (red bar) on Cifar10 and ImageNet. Among 3 datasets and 10 attacks, DISCO outperforms the baselines on 24 results. The average gains are largest on Cifar100 and ImageNet, where the RA of the prior approaches is lower. Note that the defense is more challenging on ImageNet, due to the higher dimensionality of its images~\cite{shafahi2018are}.
The full table can be found in the supplemental. 

We next evaluate the transfer ability of DISCO trained with the $L_{\infty}$ PGD attacks to four $L_2$ norm inference attacks: FGSM~\cite{fgsm}, BIM~\cite{bim}, CW~\cite{cw} and DeepFool~\cite{deepfool}. Table~\ref{tab:defense_transfer_l2} compares the defense transferability of DISCO to both adversarially trained (top block) and transformation baselines (lower block). 
DISCO generalizes well to $L_2$ attacks. It can defend more attacks than adversarially trained baselines (top block) and is more robust than the prior SOTA transformation based defenses.

Beyond different test attacks $A_{tst}$ on a PGD-trained DISCO, we also evaluated the effect of changing the training attack $A_{trn}$ used to generate the adversarial-clean pairs on which DISCO is trained. In rows 3-5 of Table~\ref{tab:transfer_dataset_classifier}, PGD~\cite{pgd}, BIM~\cite{bim} and FGSM~\cite{fgsm} are used to generate training pairs, while Autoattack is used as testing attack. BIM and PGD have comparable results, which are stronger than those of FGSM. Nevertheless all methods outperform the SOTA RobustBench defense~ \cite{Rebuffi2021FixingDA} for Autoattack on Cifar10, shown in the first row. 
These results suggests that DISCO is robust to many combinations of training and inference attacks. 

\begin{table}[tb]
\centering
\begin{minipage}{.4\linewidth}
    \centering
    \adjustbox{max width=1\linewidth}{%
    \begin{tabular}{|c|cc|c|}
    \hline
      Method & Gowal~\cite{gowal2020uncovering}  &  Rebuffi~\cite{Rebuffi2021FixingDA}  & DISCO\\
      Classifier & WRN70-16 & WRN28-10 & WRN28-10\\
      \hline
      FGSM~\cite{fgsm} &  44.53 & 38.57 & \textbf{50.4}\\
      PGD~\cite{pgd} & 40.46 & 36.09  & \textbf{74.51}\\
      BIM~\cite{bim} & 40.38 & 36.03  & \textbf{72.25}\\
      RFGSM~\cite{rfgsm} & 40.42 & 35.99  & \textbf{72.1}\\
      EotPgd~\cite{eotpgd} & 41.07 & 36.45  & \textbf{74.8}\\
      TPgd~\cite{zhang2019theoretically} & 57.52 & 52.01 &  \textbf{74.06}\\
      FFgsm~\cite{ffgsm} & 47.61 & 41.47  & \textbf{64.29}\\
      MiFgsm~\cite{mifgsm} & 42.37 & 37.31 &  \textbf{44.14} \\
      APgd~\cite{apgd} & 39.99 & 35.64 &  \textbf{77.33} \\
      Jitter~\cite{jitter} & 38.38 & 33.04 &  \textbf{73.75} \\
      \hline
      Avg. & 43.27 & 38.26  &  \textbf{67.76}  \\
     \hline
    \end{tabular}
    }
    \captionof{table}{Defense transfer across ten $L_{\infty}$ attacks ($\epsilon_{\infty}=8/255$) on Cifar100.}
    \label{tab:cifar100_defense_transfer_linf}
\end{minipage}%
\hspace{.5pt}
\begin{minipage}{.59\linewidth}
\centering
    \begin{tabular}{cc}
        \includegraphics[width=0.45\linewidth, height=0.34\linewidth]{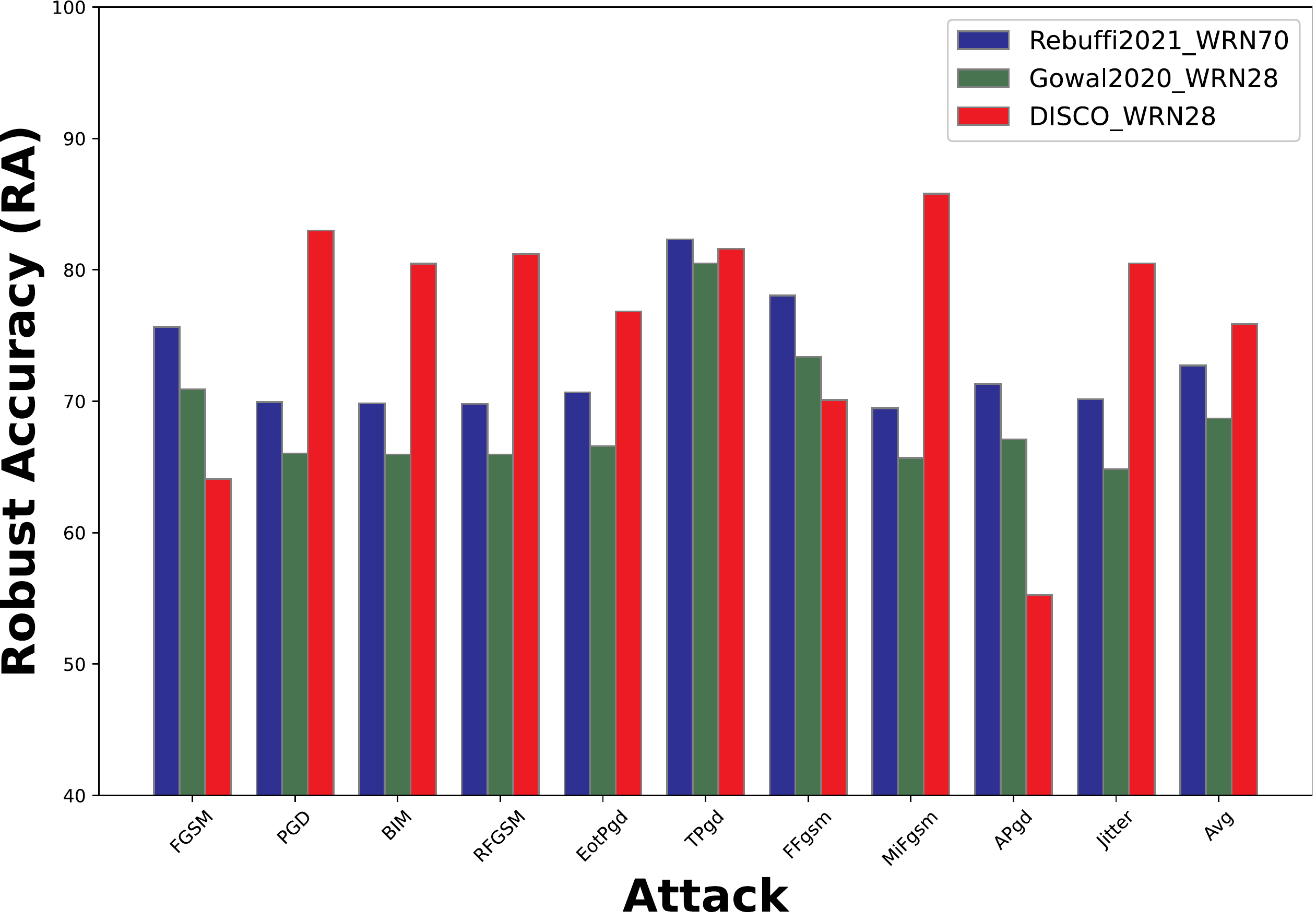} &
        \includegraphics[width=0.45\linewidth, height=0.34\linewidth]{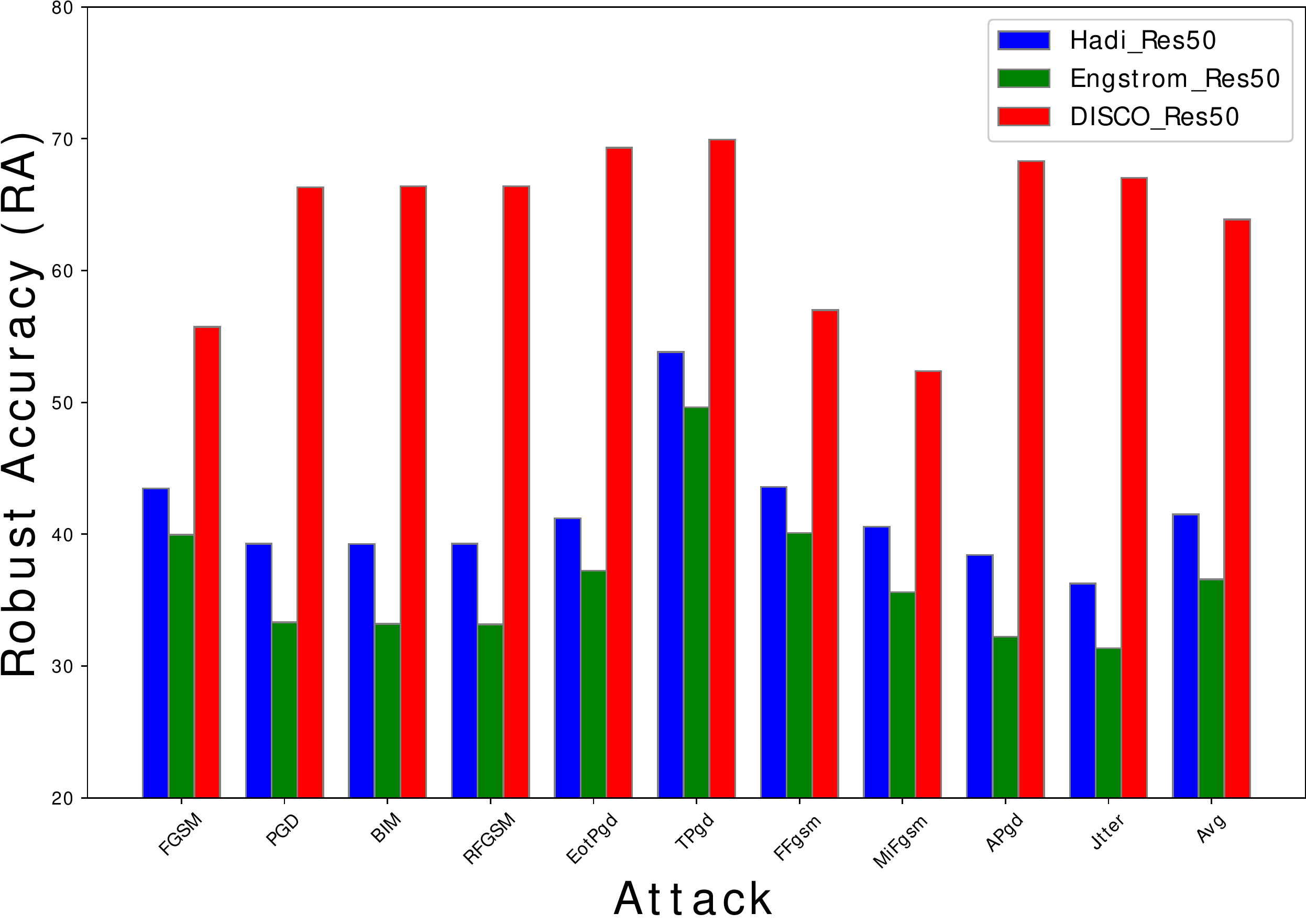}
        \\
        \scriptsize{(a)} & \scriptsize{(b)}
    \end{tabular}
    \captionof{figure}{Defense transfer across $L_{\infty}$ attacks on (a) Cifar10 and (b) ImageNet. (Blue, Green) Baselines, (Red) DISCO.}
    \label{fig:defense_transfer_linf}
\end{minipage}
\vspace{-20pt}
\end{table}

% \begin{table}[]

%     \centering
%     \adjustbox{max width=0.7\linewidth}{%
%     \begin{tabular}{|ccc|cccc|}
%       \hline
%       \multicolumn{3}{|c|}{Training} & \multicolumn{4}{|c|}{Testing}  \\
%       Attack  & Classifier & Dataset &  Classifier & SA & RA & Avg. \\
%       \hline 
%         PGD & Res18 & Cifar10 &  Vgg16 & 89.12 &	86.27 & 87.7   \\
%         PGD & Res18 & Cifar10 &  WRN28 & 89.26 & 85.56 &  87.41  \\
%         BIM & Res18 & Cifar10 &  WRN28 & 91.96 & 84.92 & 88.44  \\
%         FGSM & Res18 & Cifar10 &  WRN28 & 84.07 &	77.13 & 80.6  \\
%         FGSM & Res18 & Cifar100 &  WRN28 & 84.23 &	86.16 & 85.2\\
%         FGSM & Res18 & ImageNet &  WRN28 & 88.91 &	74.3 & 81.61\\
%     \hline 
%     \end{tabular}
%     }
%     \caption{Evaluating Cifar10 under Autoattack with different training and testing configurations.}
%     \label{tab:transfer_dataset_classifier}
% \end{table}

\begin{table}[t]
\begin{minipage}{1\linewidth}
    \centering
     \captionof{table}{Defense transfer of DISCO across training attacks, classifiers, and datasets. In all cases the inference setting is: Cifar10 dataset with Autoattack. For comparison, the RobustBench SOTA~\cite{Rebuffi2021FixingDA} for no transfer is also shown.
    % {\color{red} Could also add colums for method (DISCO VS [66]), and three "checkmark" columns for the three types of transfer.}
    % {\color{red} ADD COLUMN ON THE LEFT}
    }
    \label{tab:transfer_dataset_classifier}
    \adjustbox{max width=1\linewidth}{%
    \begin{tabular}{|c|ccc|ccc|cccc|}
       
      \hline
     
      & \multicolumn{3}{|c|}{Transfer} & \multicolumn{3}{|c|}{Training} & \multicolumn{4}{|c|}{Testing}  \\
       & Classifier & Attack & Dataset & Attack  & Classifier & Dataset &  Classifier & SA & RA & Avg. 
       \\
       
    \hline
    
    \cite{Rebuffi2021FixingDA} & & & &   Autoattack & WRN70-16 &  Cifar10& WRN70-16 & 92.23 & 66.58 & 79.41\\
     \hline
        \hline
        %\multicolumn{7}{|c|}{Classifier Transfer} \\
        \multirow{7}{*}{\begin{turn}{90}DISCO \end{turn}} & & & &  PGD & Res18 & Cifar10 &  Res18 & 89.57 & 76.03 & 82.8   \\   
       & \checkmark & & &  PGD & Res18 & Cifar10 &  VGG16 & 89.12 &	86.27 & 87.7  \\
       & \checkmark & & & PGD & Res18 & Cifar10 &  WRN28 & 89.26 & 85.56 &  87.41 \\
        \cline{2-11}
        %\multicolumn{7}{|c|}{Attack and Classifier Transfer} \\
       & \checkmark & \checkmark & & BIM & Res18 & Cifar10 &  WRN28 & 91.96 & 84.92 & 88.44  \\
      &  \checkmark & \checkmark & & FGSM & Res18 & Cifar10 &  WRN28 & 84.07 &	77.13 & 80.6 \\
        \cline{2-11}
        %\hline\multicolumn{7}{|c|}{Attack, Classifier, and Dataset Transfer} \\
     &   \checkmark & \checkmark & \checkmark & FGSM & Res18 & Cifar100 &  WRN28 & 84.23 &	86.16 & 85.2 \\
      &  \checkmark & \checkmark & \checkmark & FGSM & Res18 & ImageNet &  WRN28 & 88.91 &	74.3 & 81.61 \\
    \hline 

    \end{tabular}
    }
\end{minipage}%
% \hspace{1pt}
% \begin{minipage}{.35\linewidth}
% \centering
% \adjustbox{max width=0.95\linewidth}{%
% \begin{tabular}{|c|ccc|}
%   \hline
%   Hadi et al.~\cite{NEURIPS2020_24357dd0}   & \textbf{64.02} & 34.96 & 49.49 \\
%   w/ DISCO & 63.66 & \textbf{50.6} & \textbf{57.13}\\
%   \hline
%   Engstrom et al.~\cite{robustness}   &  \textbf{62.56} & 29.22 & 45.89 \\
%   w/ DISCO  & 62.48 & \textbf{49.44} & \textbf{55.96}\\
%   \hline
%   Wong et al.~\cite{Wong2020Fast} & \textbf{55.62} & 26.24 & 40.93 \\
%   w/ DISCO & 54.52 & \textbf{40.68} & \textbf{47.6} \\
%   \hline
% \end{tabular}
%   }
%  \captionof{table}{Improving the top ranked ResNet50 baselines on ImageNet.}
% \label{tab:plug_and_play}
    
% \end{minipage}
\vspace{-10pt}
\end{table}

\noindent\textbf{\textit{Transfer across Classifiers.}}
The first section of Table~\ref{tab:transfer_dataset_classifier} shows the results when the testing classifier is different from the training classifier. While the ResNet18 is always used to curate the training pairs of DISCO, the testing classifier varies between ResNet18, WideResNet28 and VGG16. The small impact of the classifier used for inference on the overall RA shows that DISCO is classifier agnostic and can be applied to multiple testing classifiers once it is trained.

\noindent\textbf{\textit{Transfer across Datasets.}}
The evidence that adversarial attacks push images away from the natural image manifold~\cite{Zhou_eccv20, Li2021ExploringAF,stl,Dziugaite2016ASO,Jha18,NEURIPS2020_23937b42} and that attacks can be transferred across classifiers~\cite{Demontis2019WhyDA, Wang2021AdmixET, Wu2020Skip, huang2022transferable, Wang_2021_CVPR}, suggest that it may be possible to transfer defenses across datasets. This, however, has not been studied in detail in the literature, partly because adversarially trained baselines entangle the defense and the classifier training. This is unlike DISCO and other transformation based baselines, which can be transferred across datasets. The bottom section of  Table~\ref{tab:transfer_dataset_classifier} shows the test performance on Cifar10 of DISCO trained on Cifar100 and ImageNet. Since Cifar100 images are more similar to those of Cifar10 than ImageNet ones, the Cifar100 trained DISCO transfers better than that trained on ImageNet. However, even the RA of the latter is 7.72\% higher than the best RA reported on RobustBench~\cite{Rebuffi2021FixingDA}. Note that the DISCO trained on Cifar100 and ImageNet never see images from Cifar10 and the  transfer is feasible because no limitation is imposed on the output size of DISCO. 

%\vspace{-5pt}
\subsection{Adaptive Adversary}\label{sec:adaptive}
%\vspace{-5pt}
The adaptive adversary assumes both the classifier and defense strategy are exposed to the attacker. As noted by~\cite{bpda,stl, adaptiveattack}, this setting is more challenging, especially for transformation based defenses. We adopt the BPDA~\cite{bpda} attack, which is known as an effective attack for transformation based defenses, such as DISCO. Fig.~\ref{fig:bpda_sota} compares the RA of DISCO trained with PGD attack to the results published for other methods in~\cite{stl}. For fair comparison, DISCO is combined with a VGG16 classifier. The figure confirms that both prior transformation defenses and a single stage DISCO ($K=1$) are vulnerable to an adaptive adversary. However, without training against BPDA, DISCO is 46.76\% better than prior methods. More importantly, this gain increases substantially with $K$, saturating at  RA of 57.77\% for $K=5$ stages, which significantly 
outperforms the SOTA by 57.35\%.
% {\color{red} SAY SOMETHING HERE, LIKE BETTER THAN THE SOTA FOR THIS DATA, OR ONLY SLIGHTLY WEAKER THAN THE OBLIVIOUS RESULTS OF TABLE XX OR WHATEVER IT IS.}
% {\color{red} SHOULDN'T THERE BE A PLOT JUST FOR THIS, WITH $K$ FROM 1 to 5? THIS IS IMPORTANT, BECAUSE IT IS THE ADAPTIVE SETTING. ALL WE ARE DOING IS ADDING MORE DISCOS.}
% {\color{blue}{Include the cascade result in Fig.~\ref{fig:bpda_sota}?}}
% {\color{red}  and suggests that it can be leveraged for defending adaptive attacks by randomly applying DISCO for defense,}

% I am discussing the adaptive setting in the above paragraph and figure 6 to avoid confusion. For the EDHA setting, the discussion is below and figure 7 contains the information. NO, DOSENT K=5 on both give much higer than K=3? Right noew, the adaptive performance of DISCO is 69%. THey could claim that we invented EDHA to make oursekves look good (which we did;-)). But this paragraph is completly legal under existing definitions, no? I don't understand your point of mixing the cascade disco into figure 6. The disco that I show in figure 6 now only has a single disco tps://ucsd.zoom.us/my/nunov

%\subsection{Easy to Defend but Hard to Attack}
%\vspace{-5pt}
\subsection{Cascade DISCO}
%\vspace{-5pt}
So far, we have considered the setting where the structure of the cascade DISCO is known to the attacker. DISCO supports a more sophisticated and practical setting, where the number $K$ of DISCO stages used by the defense is randomized on a per-image basis. In this case, even if the use of DISCO is exposed to the attacker, there is still uncertainty about how many stages to use in the attack. We investigated the consequences of this uncertainty by measuring the defense performance when different values of $K$ are use for attack and defense, denoted as $K_{adv}$ and $K_{def}$, respectively.
The oblivious setting has $K_{adv}=0$ and $K_{def}\geq1$ , while $K_{adv}=K_{def}$ in the adaptive setting. We now consider the case where $K_{adv} \neq K_{def}$. Fig.~\ref{fig:bpda_attack} investigates the effectiveness of cascade DISCO trained with PGD attack when faced with the BPDA~\cite{bpda} attack in this setting, where RA($K_{adv}$,$K_{def}$) is the RA when $K_{adv}$ and $K_{def}$ are used, and $K_{adv} \in \{i\}_{i=0}^5$, $K_{def} \in \{i\}_{i=1}^3$. 
%Under the oblivious setting, larger $K_{def}$ slightly improves the RA by 1.94\% (see RA(0,1) and RA(0,3)). For the standard adaptive setting, the performance of DISCO drops by 58\% when comparing RA(0,1) and RA(1,1). However, RA(1,2) boosts the gain by 54.6\% over RA(1,1) by using an addition DISCO for defense and outperforms the baselines in Fig.~\ref{fig:bpda_sota}. While the adaptive setting is harmful for DISCO, the adversary becomes weaker as $K_{adv}$ becomes larger, where RA(3,3) is significantly greater than RA(1,1) by 56.2\%. 
Under the setting of $K_{adv} \neq K_{def}$, the RA is higher than that of the adaptive setting. Take $K_{adv}=2$ for example. Both RA(2,1)=55.3 and RA(2,3)=59.8 outperform RA(2,2)=52.
% the RA further improves over the baselines in Fig.~\ref{fig:bpda_sota} and the RA of $K_{adv}=K_{def}$. For example, both RA(2,1)=55.3 and RA(2,3)=59.8 outperform RA(2,2)=52.
% The robust accuracy is quite high compared to prior arts with $K_{adv} \neq K_{def}$, with values ranging from RA(2,1)=55.3 to RA(1,3)=65.4. 
In addition, Fig.~\ref{fig:bpda_attack_def_time} compares the time to generate a single adversarial example on Cifar10 and defend against it using DISCO. Clearly, the computational resources needed to generate an attack are significantly higher than those of the defense and the ratio of attack-to-defense cost raises with $K$. Both this and the good defense performance for mismatched $K$s give the defender a strong advantage. It appears that the defense is more vulnerable when the attacker knows $K$ (adaptive setting) and even there, as seen in the previous section, the defense can obtain the upper hand by casacading several DISCO stages.

\begin{table}[t]
\begin{minipage}{.32\linewidth}
    \centering
    \begin{tabular}{c}
      \includegraphics[width=1\linewidth]{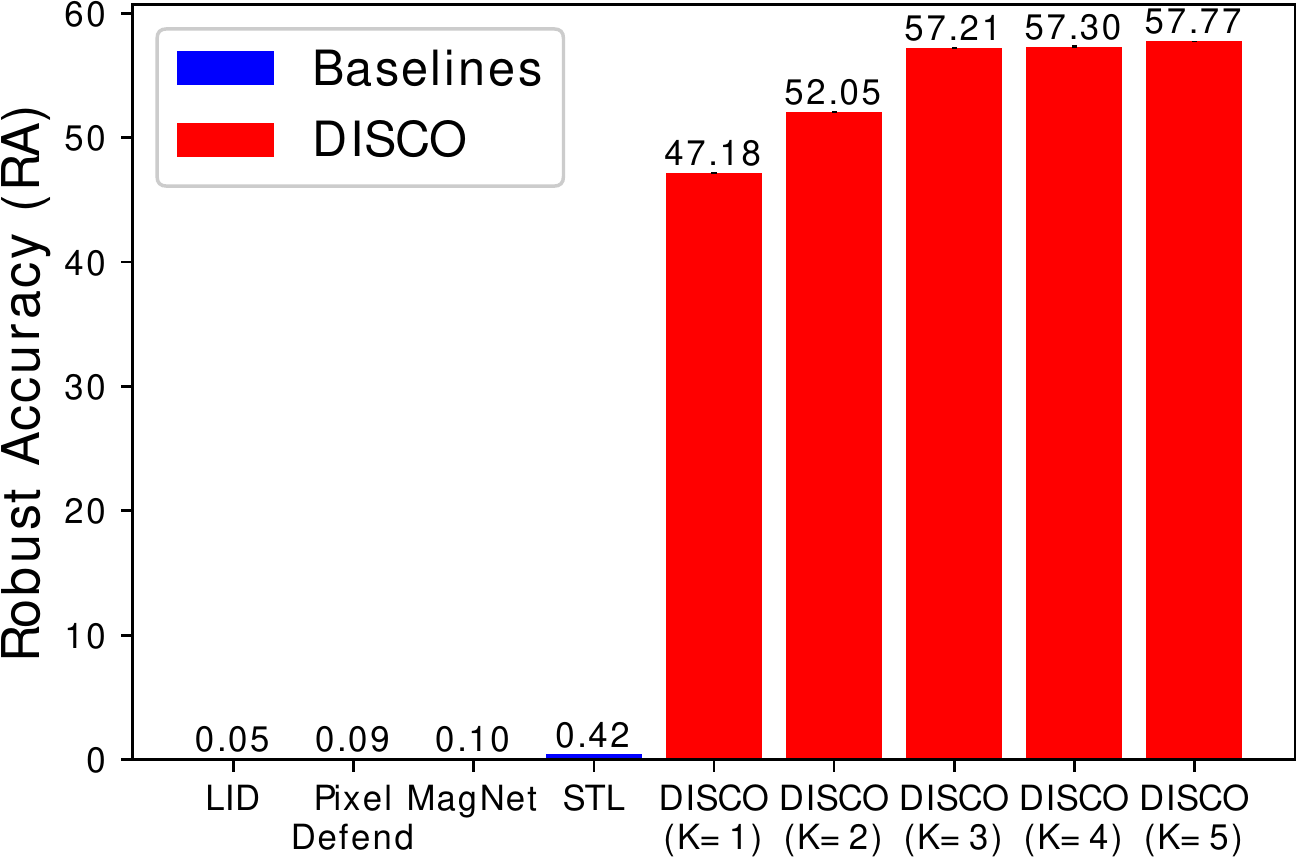}
    \end{tabular}  
 \captionof{figure}{BPDA attack on Cifar10 under the adaptive setting.}
\label{fig:bpda_sota}
\end{minipage}
\hspace{1pt}
\begin{minipage}{.32\linewidth}
    \centering
    \begin{tabular}{c}
      \includegraphics[width=\linewidth]{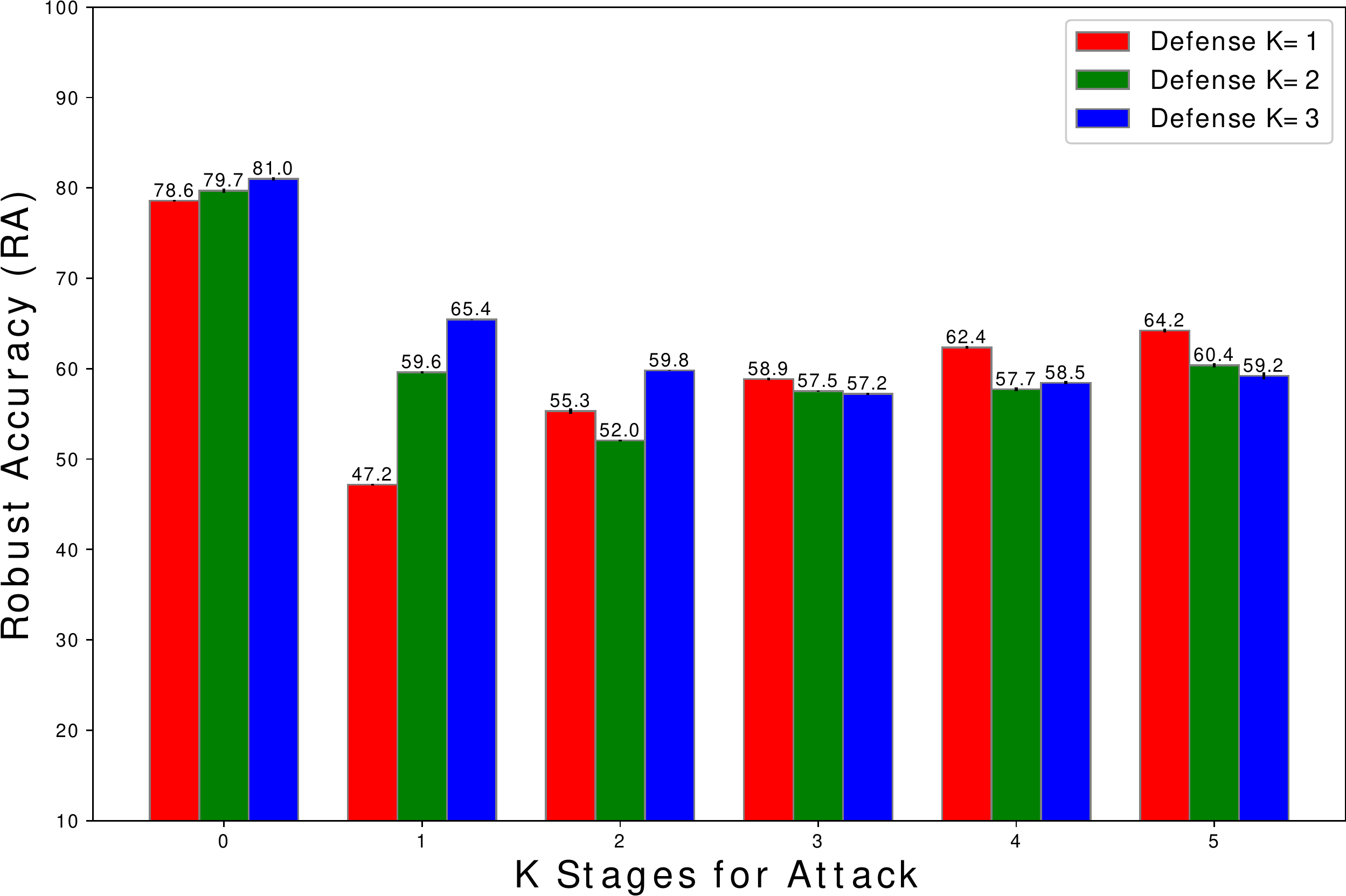}
    \end{tabular}    
    
    \captionof{figure}{BPDA attack with cascade DISCO on Cifar10. }
    \label{fig:bpda_attack}
\end{minipage} 
\hspace{1pt}
\begin{minipage}{.32\linewidth}
    \centering
    \begin{tabular}{c}
      \includegraphics[width=\linewidth]{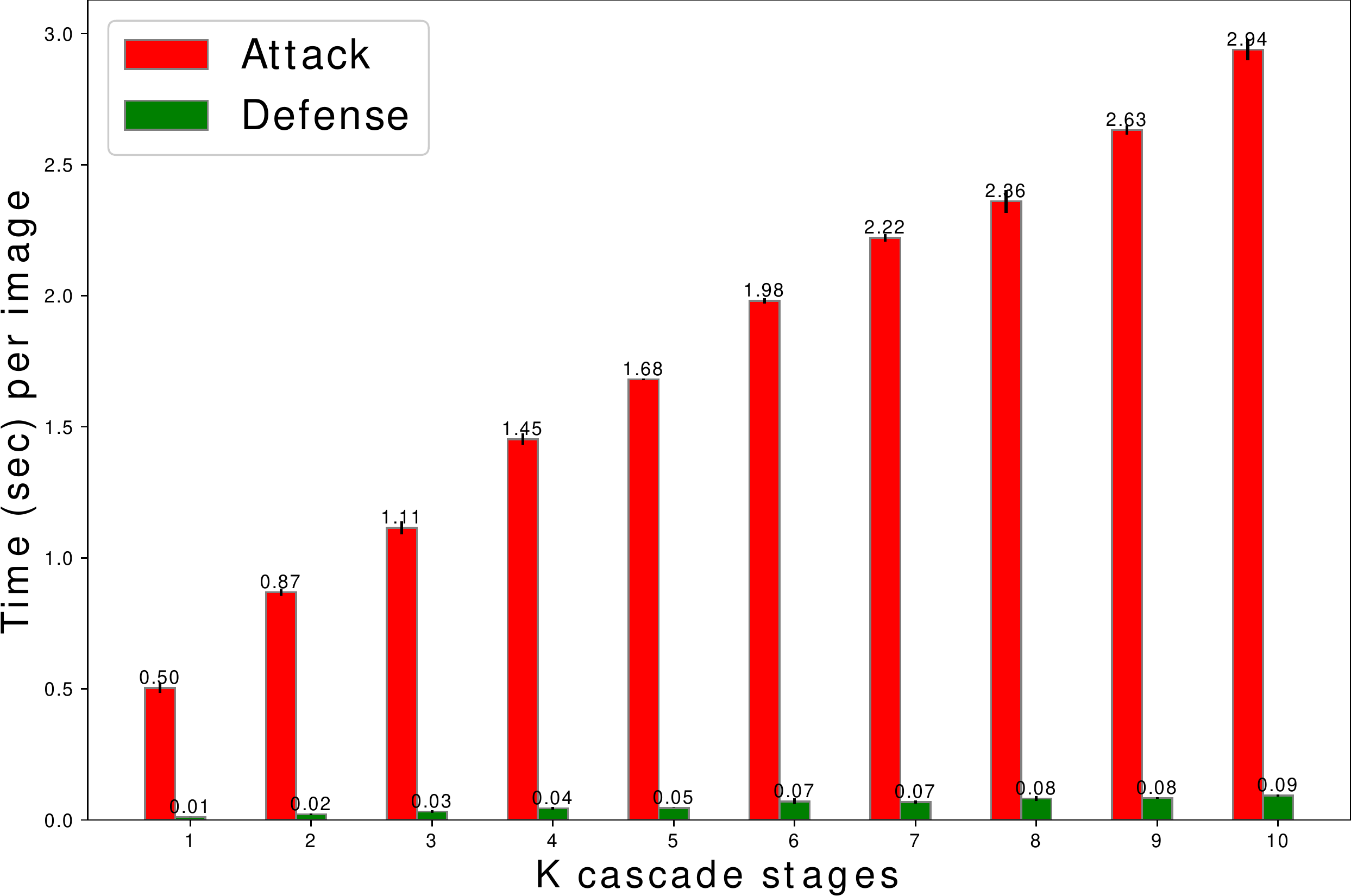}
    \end{tabular}    
    
    \captionof{figure}{Attack and defense times vs DISCO cascade length.}
    \label{fig:bpda_attack_def_time}
\end{minipage}
\vspace{-15pt}
\end{table}

% \begin{table}[]
%     \centering
%     \adjustbox{max width=0.7\linewidth}{%
%     \begin{tabular}{|c|ccc||ccc|}
%       \hline
%       \# of & \multicolumn{3}{c||}{Oblivious Attack} & \multicolumn{3}{c|}{Adaptive Attack} \\
%       DISCO & SA & RA & Avg. & SA & RA & Avg.\\
%       \hline
%       Single & 89.26 & 69.09 & 79.175
%       & 89.26 & 1.03 & 45.145\\
%       Cascade & 86.27 & 75.02 & 80.645
%       & 86.27 & 66.2 & 76.235 \\
%       \hline
%     \end{tabular}
%     }
%     \caption{BPDA attack with cascade DISCO.}
%     \label{tab:bpda}
% \end{table}

% \begin{table}[]
%     \centering
%     \adjustbox{max width=0.7\linewidth}{%
%     \begin{tabular}{|cccccc|cc|}
%     \hline
%     SAP & TE &  LID & PixelDefend & MagNet & STL &  Sinlge & Cascade \\
%     \cite{sap} & \cite{te} &  \cite{lid} & \cite{pixeldefend} & \cite{magnet} & \cite{stl} & DISCO & DISCO \\
%      \hline
%     0.00 & 0.00 &  0.05 & 0.09 & 0.10 & 0.42 & 1.03 & 66.2 \\
%      \hline
%     \end{tabular}
%      }
%     \caption{Caption}
%     \label{tab:bpda_sota}
% \end{table}

%\vspace{-5pt}
\section{Discussion, Societal  Impact and Limitations}
%\vspace{-10pt}
In this work, we have proposed the use of local implicit functions for adversarial defense. Given an input adversarial image and a query location, the DISCO model is proposed to project the RGB value of each image pixel into the image manifold, conditional on deep features centered in the pixel neighborhood. By training this projection with adversarial and clean images, DISCO learns to remove the adversarial perturbation. Experiments demonstrate DISCO's computational efficiency, its outstanding defense performance and transfer ability across attacks, datasets and classifiers. The cascaded version of DISCO further strengthens the defense with minor additional cost. 

{\bf Limitations:} While DISCO shows superior performance on the attacks studied in this work (mainly norm-bounded attacks), it remains to be tested whether it is robust to other type of attacks~\cite{onepixelattack,Brown2017AdversarialP,hu2021naturalistic, Ho2019CatastrophicCP, Laidlaw2019FunctionalAA}, such as one pixel attack~\cite{onepixelattack}, patch attacks~\cite{Brown2017AdversarialP,hu2021naturalistic} or functional adversarial attack~\cite{Laidlaw2019FunctionalAA}. In addition, more evaluation configurations across attacks, datasets and classifiers will be investigated in the future. 

{\bf Societal Impact:} We hope the idea of using local implicit functions can inspire better defenses and prevent the nefarious effects of deep learning attacks. Obviously, better defenses can also be leveraged by bad actors to improve resistance to the efforts of law enforcement, for example.

\section{Acknowledgments}
This work was partially funded by NSF awards IIS1924937 and IIS-2041009, a gift from Amazon, a gift from Qualcomm, and NVIDIA GPU donations. We also acknowledge and thank the use of the Nautilus platform for some of the experiments discussed above.

\clearpage
\appendix

\setcounter{table}{0}
\renewcommand\thetable{\Alph{table}}
\setcounter{figure}{0}
\renewcommand\thefigure{\Alph{figure}}

\begin{center} 
\huge \textbf{Appendix}
\end{center}

\section{Compare to SOTA in RobustBench}
In this section, we list the quantitative result of the baselines in RobustBench~\cite{croce2021robustbench} . Table~\ref{tab:cifar10_robustbench_Linf}, \ref{tab:cifar100_robustbench_Linf} and \ref{tab:imagenet_robustbench_Linf} correspond to Fig.6(a), (b) and (c) of the main paper, respectively. Table~\ref{tab:cifar10_robustbench_L2} shows the baselines under Autoattack with $\epsilon_{2}=0.5$. The index displayed in each table corresponds to the index shown in Fig.6 in the main paper. 
The baselines of each table are grouped into No defense (first block), Adversarially trained defense in RobustBench (second block), Transformation based defense (third block) and  DISCO (last block). The results of adversarially trained baselines are copied from RobustBench, while the results of transformation-based defenses are obtained with our implementation. For STL~\cite{stl}, models with different sparse constraints $\lambda$ are used from the publicly available STL github\footnote{\label{stl}\url{https://github.com/GitBoSun/AdvDefense\_CSC}}. DISCO is also combined with various classifiers for evaluation. More discussion can be found in Sec. 4.1 of the paper.

\begin{table}[htb]
\adjustbox{max width=\linewidth}{%
    \centering
    \begin{tabular}{|cc|cccc||cc|cccc|}
      \hline 
      ID & Method &  Standard Acc. & Robust Acc. & Avg. Acc. &  Model & 
      ID &  Method  &  Standard Acc. & Robust Acc. & Avg. Acc. & Model\\
      \hline
      \hline
      0 & No Defense  &  94.78 & 0 & 47.39 & WRN28-10 &&&&&& \\
      \hline
      \hline
      1 & Rebuffi et al.~\cite{Rebuffi2021FixingDA} & 92.23 & 66.58 & 79.41 & WRN70-16 & 
      2 & Gowal et al.~\cite{NEURIPS2021_Gowal} &  88.74 & 66.11 & 77.43 & WRN70-16 \\
      %\hline
      3 & Gowal et al.~\cite{gowal2020uncovering} & 91.1 & 65.88 & 78.49 & WRN70-16 & 
      4 & Rebuffi et al.~\cite{Rebuffi2021FixingDA} &  88.5 & 64.64 & 76.57 & WRN106-16 \\
      %\hline
      5 & Rebuffi et al.~\cite{Rebuffi2021FixingDA} & 88.54 & 64.25 & 76.4 & WRN70-16 & 
      6 & Kang et al.~\cite{kang2021stable}  &  93.73 & 71.28 & 82.51 & WRN70-16 \\
      %\hline
      7 & Gowal et al.~\cite{NEURIPS2021_Gowal} & 87.5 & 63.44 & 75.47 & WRN28-10 & 
      8 & Pang et al.~\cite{Pang22} &  89.01 & 63.35 & 76.18 & WRN70-16 \\
      %\hline
      9 & Rade et al.~\cite{rade2021helperbased} & 91.47 & 62.83 & 77.15 & WRN34-10 & 
      10 & Sehwag et al.~\cite{sehwag2022robust} &  87.3 & 62.79 & 75.05 & ResNest152 \\
      %\hline
      11 & Gowal et al.~\cite{gowal2020uncovering} & 89.48 & 62.8 & 76.14 & WRN28-10 & 
      12 & Huang et al.~\cite{huang2021exploring} &  91.23 & 62.54 & 76.89 & WRN34-R \\
      %\hline
      13 & Huang et al.~\cite{huang2021exploring} & 90.56 & 61.56 & 76.06 & WRN34-R & 
      14 & Dai et al.~\cite{Sihui_corr21} &  87.02 & 61.55 & 74.29 & WRN28-10 \\
      %\hline
      15 & Pang et al.~\cite{Pang22} & 88.61 & 61.04 & 74.83 & WRN28-10 & 
      16 & Rade et al.~\cite{rade2021helperbased} &  88.16 & 60.97 & 74.57 & WRN28-10 \\
      %\hline
      17 & Rebuffi et al.~\cite{Rebuffi2021FixingDA} & 87.33 & 60.75 & 74.04 & WRN28-10 & 
      18 & Wu et al.~\cite{wu2021do} &  87.67 & 60.65 & 74.16 & WRN34-15 \\
      %\hline
      19 & Sridhar et al.~\cite{Sridhar2021RobustLV} & 86.53 & 60.41 & 73.47 & WRN34-15 & 
      20 & Sehwag et al.~\cite{sehwag2022Proxy} &  86.68 & 60.27 & 73.48 & WRN34-10 \\
      %\hline
      21 & Wu et al.~\cite{wu2020adversarial} & 88.25 & 60.04 & 74.15 & WRN28-10 & 
      22 & Sehwag et al.~\cite{sehwag2022Proxy} &  89.46 & 59.66 & 74.56 & WRN28-10 \\
      %\hline
      23 & Zhang et al.~\cite{zhang2021geometryaware} & 89.36 & 59.64 & 74.5 & WRN28-10 & 
      24 & Yair et al.~\cite{NEURIPS2019_Carmon} &  89.69 & 59.53 & 74.61 & WRN28-10 \\
      %\hline
      25 & Gowal et al.~\cite{NEURIPS2021_Gowal} & 87.35 & 58.63 & 72.99 & PreActRes18 & 
      26 & Addepalli et al.~\cite{addepalli2021towards} &  85.32 & 58.04 & 71.68 & WRN34-10 \\
      %\hline
      27 & Chen et al.~\cite{Chen21} & 86.03 & 57.71 & 71.87 & WRN34-20 & 
      28 & Rade et al.~\cite{rade2021helperbased} &  89.02 & 57.67 & 73.35 & PreActRes18 \\
      %\hline
      29 & Gowal et al.~\cite{gowal2020uncovering}  & 85.29 & 57.2 & 71.25 & WRN70-16 & 
      30 & Sehwag et al.~\cite{sehwag2020hydra} &  88.98 & 57.14 & 73.06 & WRN28-10 \\
      %\hline
      31 & Rade et al.~\cite{rade2021helperbased} & 86.86 & 57.09 & 71.98 & PreActRes18 & 
      32 & Chen et al.~\cite{Chen21} &  85.21 & 56.94 & 71.08 & WRN34-10 \\
      %\hline
      33 & Gowal et al.~\cite{gowal2020uncovering} & 85.64	& 56.86 & 71.25 & WRN34-20 & 
      34 & Rebuffi et al.~\cite{Rebuffi2021FixingDA} &  83.53 & 56.66 & 70.1 & PreActRes18 \\
      %\hline
      35 & Wang et al.~\cite{Wang2020Improving} & 87.5 & 56.29 & 71.9 & WRN28-10 & 
      36 & Wu et al.~\cite{wu2020adversarial} &  85.36 & 56.17 & 70.77 & WRN34-10 \\
      %\hline
      37 & Alayrac et al.~\cite{NEURIPS2019_Alayrac} & 86.46 & 56.03 & 71.25 & WRN28-10 & 
      38 & Sehwag et al.~\cite{sehwag2022Proxy} &  84.59 & 55.54 & 70.07 & Res18 \\
      %\hline
      39 & Dan et al.~\cite{hendrycks2019pretraining} & 87.11 & 54.92 & 71.02 & WRN28-10 & 
      40 & Pang et al.~\cite{pang2021bag} &  86.43 & 54.39 & 70.41 & WRN34-20 \\
      %\hline
      41 & Pang et al.~\cite{NEURIPS2020_Pang} & 85.14 & 53.74 & 69.44 & WRN34-20 & 
      42 & Cui et al.~\cite{Cui2021LearnableBG} &  88.7	& 53.57 & 71.14 & WRN34-20 \\
      %\hline
      43 & Zhang et al.~\cite{zhang2020fat} & 84.52	& 53.51 & 69.02 & WRN34-10 & 
      44 & Rice et al.~\cite{Rice2020OverfittingIA} &  85.34 & 53.42 & 69.38 & WRN34-20 \\
      %\hline
      45 & Huang et al.~\cite{huang2020self} & 83.48	& 53.34 & 68.41 & WRN34-10 & 
      46 & Zhang et al.~\cite{zhang2019theoretically} &  84.92 & 53.08 & 69 & WRN34-10 \\
      %\hline
      47 & Cui et al.~\cite{cui2020learnable} & 88.22	& 52.86 & 70.54 & WRN34-10 & 
      48 & Qin et al.~\cite{Qin2019AdversarialRT} &  86.28 & 52.84 & 69.56 & WRN40-8 \\
      %\hline
      49 & Chen et al.~\cite{Chen_2020_CVPR} & 86.04	& 51.56 & 68.8 & Res50 & 
      50 & Chen et al.~\cite{chen2021efficient} &  85.32 & 51.12 & 68.22 & WRN34-10 \\
      %\hline
      51 & Addepalli et al.~\cite{addepalli2021oaat} & 80.24	& 51.06 & 65.65 & Res18 & 
      52 & Chawin et al.~\cite{Chawin20} &  86.84 & 50.72 & 68.78 & WRN34-10 \\
      %\hline
      53 & Engstrom et al.~\cite{robustness} & 87.03	& 49.25 & 68.14 & Res50 & 
      54 & Sinha et al.~\cite{Sinha2019HarnessingTV} &  87.8 & 49.12 & 68.46 & WRN34-10 \\
      %\hline
      55 & Mao et al.~\cite{NEURIPS2019_c24cd76e} & 86.21 & 47.41 & 66.81 & WRN34-10 & 
      56 & Zhang et al.~\cite{Zhang19} &  87.2 & 44.83 & 66.02 & WRN34-10 \\
      %\hline
      57 & Madry et al.~\cite{madry2018towards} & 87.14	& 44.04 & 65.59 & WRN34-10 & 
      58 & Maksym et al.~\cite{andriushchenko2020understanding} &  79.84 & 43.93 & 61.89 & PreActRes18 \\
      %\hline
      59 & Pang et al.~\cite{Pang2020Rethinking} & 80.89	& 43.48 & 62.19 & Res32 & 
      60 & Wong et al.~\cite{Wong2020Fast} &  83.34 & 43.21 & 63.28 & PreActRes18 \\
      %\hline
      61 & Shafahi et al.~\cite{Shafahi2019AdversarialTF} & 86.11	& 41.47 & 63.79 & WRN34-10 & 
      62 & Ding et al.~\cite{Ding2020MMA} &  84.36 & 41.44 & 62.9 & WRN28-4 \\
      %\hline
      63 & Souvik et al.~\cite{Souvik21} & 87.32	& 40.41 & 63.87 & Res18 & 
      64 & Matan et al.~\cite{atzmon2019controlling} &  81.3	& 40.22 & 60.76 & Res18 \\
      %\hline
      65 & Moosavi-Dezfooli et al.~\cite{MoosaviDezfooli2019RobustnessVC} & 83.11	& 38.5 & 60.81 & Res18 & 
      66 & Zhang et al.~\cite{feature_scatter} &  89.98 & 36.64 & 63.31 & WRN28-10 \\
      %\hline
      67 & Zhang et al.~\cite{zhang2020adversarial} & 90.25	& 36.45 & 63.35 & WRN28-10 & 
      68 & Jang et al.~\cite{Jang19} &  78.91 & 34.95 & 56.93 & Res20 \\
      %\hline
      69 & Kim et al.~\cite{kim2020sensible} & 91.51	& 34.22 & 62.87 & WRN34-10 & 
      70 & Zhang et al.~\cite{zhang2020towards} &  44.73 & 32.64 & 38.69 & 5 layer CNN \\
      %\hline
      71 & Wang et al.~\cite{Wang2019BilateralAT} & 92.8 & 29.35 & 61.08 & WRN28-10 & 
      72 & Xiao et al.~\cite{Xiao2020Enhancing} &  79.28 & 18.5 & 48.89 & DenseNet121 \\
      %\hline
      73 & Jin et al.~\cite{jin2021manifold} & 90.84	& 1.35 & 46.1 & Res18 & 
      74 & Aamir et al.~\cite{Mustafa_2019_ICCV} &  89.16 & 0.28 & 44.72 & Res110 \\
      %\hline
      75 & Chan et al.~\cite{Chan2020Jacobian} & 93.79 & 0.26 & 47.03 & WRN34-10 & & & & & & \\
      \hline
      \hline
      76 & Bit Reduction~\cite{Xu2018FeatureSD} & 92.66 & 1.04 & 46.85 & WRN28-10 & 
      77 & Jpeg~\cite{Dziugaite2016ASO}  &  83.9 & 50.73 & 67.32 & WRN28-10 \\
      %\hline
      78 & Input Rand.~\cite{xie2017mitigating} & 94.3 & 8.59 & 51.45 & WRN28-10 & 
      79 & LIIF~\cite{chen2021learning}  &  94.85 & 0.22 & 47.54 & WRN28-10 \\
      %\hline
      80 & AutoEncoder & 76.54 & 67.41 & 71.98 & WRN28-10 & 
      %\hline
      81 & STL~\cite{stl} (k=64 s=8 $\lambda$=0.1) & 90.65 & 57.28 & 73.97 & WRN28-10 \\
      82 & STL~\cite{stl} (k=64 s=8 $\lambda$=0.15)  &  86.77 & 66.94 & 76.86 & WRN28-10 &
      %\hline
      83 & STL~\cite{stl} (k=64 s=8 $\lambda$=0.2)  &  82.22 & 67.92 & 75.07 & WRN28-10 \\
      84 & Median Filter  &  79.67 & 42.49 & 61.08 & WRN28-10 &&&&&& \\
      \hline
      \hline
      85 & DISCO & 89.26 & 85.56 $\pm$ 0.02 & 87.41 & WRN28-10 &&&&&& \\
      \hline
      
    \end{tabular}
    }
    \caption{Cifar10 baselines and DISCO under Autoattack ($\epsilon_{\infty}=8/255$). This table corresponds to Fig. 6(a) in the main paper.}
    \label{tab:cifar10_robustbench_Linf}
\end{table}

\begin{table}[htb]
\adjustbox{max width=\linewidth}{%
    \centering
    \begin{tabular}{|cc|cccc||cc|cccc|}
      \hline 
      ID & Method &  Standard Acc. & Robust Acc. & Avg. Acc. &  Model & 
      ID &  Method  &  Standard Acc. & Robust Acc. & Avg. Acc. & Model\\
      \hline
      \hline
      0 & No Defense & 94.78 & 0 & 47.39 & WRN28-10 &&&&&& \\
      \hline
      \hline
      1 & Rebuffi et al.~\cite{Rebuffi2021FixingDA} &  95.74 & 82.32 & 89.03 & WRN70-16 & 
      2 & Gowal et al.~\cite{gowal2020uncovering} &  94.74 & 80.53 & 87.64 & WRN70-16  \\
      %\hline
      3 & Rebuffi et al.~\cite{Rebuffi2021FixingDA} &  92.41 & 80.42 & 86.42 & WRN70-16 & 
      4 & Rebuffi et al.~\cite{Rebuffi2021FixingDA} &  91.79 & 78.8 & 85.30 & WRN28-10    \\
      %\hline
      5 & Augustin et al.~\cite{Augustin_eccv20} &  93.96 & 78.79 & 86.38 & WRN34-10 &  
      6 & Sehwag et al.~\cite{sehwag2022robust} &  90.93 & 77.24 & 84.09 & WRN34-10   \\
      %\hline
      7 & Augustin et al.~\cite{Augustin_eccv20} &  92.23 & 76.25 & 84.24 & WRN34-10 &  
      8 & Rade et al.~\cite{rade2021helperbased} &  90.57 & 76.15 & 83.36 & PreActRes18    \\
      %\hline
      9 & Rebuffi et al.~\cite{Rebuffi2021FixingDA} &  90.33 & 75.86 & 83.10 & PreActRes18 &
      10 & Gowal et al.~\cite{gowal2020uncovering} &  90.9 & 74.5 & 82.70 & WRN70-16  \\
      %\hline
      11 & Sehwag et al.~\cite{sehwag2022robust} &  89.76 & 74.41 & 82.09 & Res18 & 
      12 & Wu et al.~\cite{wu2020adversarial} &  88.51 & 73.66 & 81.09 & WRN34-10   \\
      %\hline
      13 & Augustin et al.~\cite{Augustin_eccv20} &  91.08 & 72.91 & 82.00 & Res50 &
      14 & Engstrom et al.~\cite{robustness} &  90.83 & 69.24 & 80.04 & Res50   \\
      %\hline
      15 & Rice et al.~\cite{Rice2020OverfittingIA} &  88.67 & 67.68 & 78.18 & PreActRes18 &  
      16 & Rony et al.~\cite{Rony2019DecouplingDA} &  89.05 & 66.44 & 77.75 & WRN28-10  \\
      %\hline
      17 & Ding et al.~\cite{Ding2020MMA} &  88.02 & 66.09 & 77.06 & WRN28-4 &
       &  &  &  &  &  \\
      %\hline
      \hline
      \hline
      18 & Bit Reduction~\cite{Xu2018FeatureSD} & 92.66 & 3.8 & 48.23 & WRN28-10 & 
      19 & Jpeg~\cite{Dziugaite2016ASO}  & 83.9  & 69.85 & 76.88 & WRN28-10 \\
      %\hline
      20 & Input Rand.~\cite{xie2017mitigating} & 94.3 &  25.71 & 60.01 & WRN28-10 & 
      21 & LIIF~\cite{chen2021learning}  & 94.85  & 0.22 & 47.54 & WRN28-10 \\
      %\hline
      22 & AutoEncoder & 76.54 & 71.71 & 74.13 & WRN28-10 & 
      %\hline
      23 & STL~\cite{stl} (k=64 s=8 $\lambda$=0.1) & 90.65 & 75.55 & 83.1 & WRN28-10 \\
      24 & STL~\cite{stl} (k=64 s=8 $\lambda$=0.15)  & 86.77  & 76.45 & 81.61 & WRN28-10 &
      %\hline
      25 & STL~\cite{stl} (k=64 s=8 $\lambda$=0.2)  & 82.22  & 74.33 & 78.28 & WRN28-10 \\
      26 & Median Filter  & 79.67 & 63.94 & 71.81 & WRN28-10 &&&&&& \\
      \hline
      \hline
      27 & DISCO & 89.26 & 88.47 $\pm$ 0.16 & 88.87 & WRN28-10 &&&&&& \\
      \hline
\end{tabular}
    }
    \caption{Cifar10 baselines and DISCO under Autoattack ($\epsilon_{2}=0.5$). }
    \label{tab:cifar10_robustbench_L2}
\end{table}

\begin{table}[htb]
\adjustbox{max width=\linewidth}{%
    \centering
    \begin{tabular}{|cc|cccc||cc|cccc|}
      \hline 
      ID & Method &  Standard Acc. & Robust Acc. & Avg. Acc. &  Model & 
      ID &  Method  &  Standard Acc. & Robust Acc. & Avg. Acc. & Model\\
      \hline
      \hline
      0 & No Defense & 80.37 & 0 & 41.78 & WRN28-10 &&&&&& \\
      \hline
      \hline
      1 & Gowal et al.~\cite{gowal2020uncovering} & 69.15 & 36.88 & 53.02 & WRN70-16 &
      2 & Rebuffi et al.~\cite{Rebuffi2021FixingDA} & 63.56 & 34.64 & 49.1 & WRN70-16 \\
      %\hline
      3 & Pang et al.~\cite{Pang22} & 65.56 & 33.05 & 49.31 & WRN70-16 &
      4 & Rebuffi et al.~\cite{Rebuffi2021FixingDA} & 62.41 & 32.06 & 47.24 & WRN28-10 \\
      %\hline
      5 & Sehwag et al.~\cite{sehwag2022robust} & 65.93 & 31.15 & 48.54 & WRN34-10 &
      6 & Pang et al.~\cite{Pang22} & 63.66 & 31.08 & 47.37 & WRN28-10\\
      %\hline
      7 & Chen et al.~\cite{Chen21} & 64.07 & 30.59 & 47.33 & WRN34-10  &
      8 & Addepalli et al.~\cite{addepalli2021oaat} & 65.73 & 30.35 & 48.04 & WRN34-10 \\
      %\hline
      9 & Cui et al.~\cite{Cui2021LearnableBG} & 62.55 & 30.2 & 46.38 & WRN34-20 &
      10 & Gowal et al.~\cite{gowal2020uncovering} & 60.86 & 30.03 & 45.45 & WRN70-16 \\
      %\hline
      11 & Cui et al.~\cite{Cui2021LearnableBG} & 60.64 & 29.33 & 44.99 & WRN34-10 &
      12 & Rade et al.~\cite{rade2021helperbased} & 61.5 & 28.88 & 45.19 & PreActRes18 \\
      %\hline
      13 & Wu et al.~\cite{wu2020adversarial} & 60.38 & 28.86 & 44.62 & WRN34-10 &
      14 & Rebuffi et al.~\cite{Rebuffi2021FixingDA} & 56.87 & 28.5 & 42.69 & PreActRes18\\
      %\hline
      15 & Dan et al.~\cite{hendrycks2019pretraining} & 59.23	& 28.42 & 43.83 & WRN28-10 &
      16 & Cui et al.~\cite{Cui2021LearnableBG} & 70.25 & 27.16 & 48.71 & WRN34-10 \\
      %\hline
      17 & Addepalli et al.~\cite{addepalli2021oaat} & 62.02 & 27.14 & 44.58 & PreActRes18 &
      18 & Chen et al.~\cite{chen2021efficient} & 62.15 & 26.94 & 44.55 & WRN34-10\\
      %\hline
      19 & Chawin et al.~\cite{Chawin20} & 62.82 & 24.57 & 43.7 & WRN34-10 &
      20 & Rice et al.~\cite{Rice2020OverfittingIA} & 53.83 & 18.95 & 36.39 & PreActRes18\\
      \hline
      \hline
      21 & Bit Reduction~\cite{Xu2018FeatureSD} & 76.86 & 3.78 & 40.32 & WRN28-10 & 
      22 & Jpeg~\cite{Dziugaite2016ASO}  &  61.89 & 39.59 & 50.74 & WRN28-10 \\
      %\hline
      23 & Input Rand.~\cite{xie2017mitigating} & 73.57 & 3.31 & 38.44 & WRN28-10 & 
      24 & LIIF~\cite{chen2021learning}  &  80.3 & 3.36 & 41.83 & WRN28-10 \\
      %\hline
      25 & AutoEncoder & 58.79 & 48.36 & 53.575 & WRN28-10 & 
      26 & STL~\cite{stl} (k=64 s=8 $\lambda$=0.1) & 74.28 & 30.05 & 52.17 & WRN28-10 \\
      %\hline
      27 & STL~\cite{stl} (k=64 s=8 $\lambda$=0.15)  &  70.3 & 41.82 & 56.06 & WRN28-10 & 
      28 & STL~\cite{stl} (k=64 s=8 $\lambda$=0.2)  &  67.41 & 46.07 & 56.74 & WRN28-10 \\
      29 & Median Filter  &  65.78 & 34.52 & 50.15 & WRN28-10 &&&&&& \\
      \hline
      \hline
      30 & DISCO & 72.07 & 67.93$\pm$0.17 & 70 & WRN28-10 &
      31 & DISCO & 71.62 & 69.01 $\pm$0.19 & 70.32 & WRN34-10 \\
      \hline
\end{tabular}
    }
    \caption{Cifar100 baselines and DISCO under Autoattack ($\epsilon_{\infty}=8/255$). This table corresponds to Fig. 6(b) in the main paper.}
    \label{tab:cifar100_robustbench_Linf}
\end{table}

\begin{table}[htb]
\adjustbox{max width=\linewidth}{%
    \centering
    \begin{tabular}{|cc|cccc||cc|cccc|}
      \hline 
      ID & Method &  Standard Acc. & Robust Acc. & Avg. Acc. &  Model & 
      ID &  Method  &  Standard Acc. & Robust Acc. & Avg. Acc. & Model\\
      \hline
      \hline
      0 & No Defense & 76.52 & 0 & 38.26 & Res50 &&&&&& \\
      \hline
      \hline
      1 & Hadi et al.~\cite{NEURIPS2020_24357dd0} & 68.46 & 38.14 &	53.3 & WRN50-2 &
      2 & Hadi et al.~\cite{NEURIPS2020_24357dd0} & 64.02 & 34.96 & 49.49 & Res50 \\
      %\hline
      3 & Engstrom et al.~\cite{robustness} & 62.56 & 29.22 & 45.89 & Res50 &
      4 & Wong et al.~\cite{Wong2020Fast} & 55.62 & 26.24 & 40.93 & Res50 \\
      %\hline
      5 & Hadi et al.~\cite{NEURIPS2020_24357dd0} & 52.92 & 25.32 & 39.12 & Res18 &&&&&& 
      \\
      \hline
      \hline
      6 & Bit Reduction~\cite{Xu2018FeatureSD} & 67.64 & 4.04 & 35.84 & Res18 & 
      7 & Bit Reduction~\cite{Xu2018FeatureSD} & 73.82 & 1.86 & 37.84 & Res50 \\
      %\hline
      8 & Bit Reduction~\cite{Xu2018FeatureSD} & 75.06 & 4.96 & 40.01 & WRN50-2 & 
      9 & Jpeg~\cite{Dziugaite2016ASO} & 67.18 & 13.08 & 40.13 & Res18 \\
      %\hline
      10 & Jpeg~\cite{Dziugaite2016ASO} & 73.64 & 33.42 & 53.53 & Res50 &
      11 & Jpeg~\cite{Dziugaite2016ASO} & 75.42 & 24.9 & 50.16 & WRN50-2 \\
      %\hline
      12 & Input Rand.~\cite{xie2017mitigating} & 64 & 17.78 & 40.89 & Res18 & 
      13 & Input Rand..~\cite{xie2017mitigating} & 74.02 & 18.84 & 46.43 & Res50 \\
      %\hline
      14 & Input Rand.~\cite{xie2017mitigating} & 71.7 & 23.58 & 47.64 & WRN50-2 &
      15 & STL~\cite{stl} (k=64 s=8 $\lambda$=0.1)  &  67.5 & 18.5 & 43 & Res18 \\
      %\hline
      16 & STL~\cite{stl} (k=64 s=8 $\lambda$=0.2)  &   65.64 & 32.9 & 49.27  & Res18 & 
      17 & STL~\cite{stl} (k=64 s=8 $\lambda$=0.1)  &  72.56 & 32.7 & 52.63 & Res50 \\
      %\hline
      18 & STL~\cite{stl} (k=64 s=8 $\lambda$=0.2)  & 68.3 & 50.16 & 59.23 & Res50 &
      19 & Median Filter & 66.1 & 10.34 & 38.22 & Res18 \\
      20 & Median Filter & 71.68 & 17.36 & 44.52 & Res50 &&&&&& \\
      \hline
      \hline
      21 & DISCO & 67.98 & 60.88$\pm$0.17 & 64.43 & Res18 &
      22 & DISCO & 72.64 & 68.2$\pm$0.29 & 70.42 & Res50 \\
      23 & DISCO & 75.1 & 69.5$\pm$0.23 & 72.3 & WRN50-2 &&&&&& \\
      \hline
\end{tabular}
    }
    \caption{ImageNet baselines and DISCO under Autoattack ($\epsilon_{\infty}=4/255$). This table corresponds to Fig. 6(c) in the main paper.}
    \label{tab:imagenet_robustbench_Linf}
\end{table}

\clearpage
\section{Defense Transfer}

In this section, we discuss the qualitative results of DISCO transferability across attacks. Table~\ref{tab:cifar10_defense_transfer_linf}, \ref{tab:cifar100_defense_transfer_linf_appendix} and \ref{tab:imagenet_defense_transfer_linf} represents the results for Cifar10, Cifar100 and ImageNet, respectively. The corresponding plots are illustrated in  Fig.~\ref{fig:cifar10_defense_transfer_linf},  \ref{fig:cifar100_defense_transfer_linf} and \ref{fig:imagenet_defense_transfer_linf}. More discussion can be found in Sec. 4.1 of the paper.

\begin{table}[htb]
\begin{minipage}{.6\linewidth}
    \captionof{table}{Defense Transfer across $L_{\infty}$ attacks ($\epsilon_{\infty}=8/255$) on Cifar10.}
    \label{tab:cifar10_defense_transfer_linf}
    \adjustbox{max width=1\linewidth}{%
    \begin{tabular}{|c|cc|c|}
    \hline
      Method & Rebuffi et al.~\cite{Rebuffi2021FixingDA} & Gowal et al.~\cite{gowal2020uncovering}  & DISCO \\
      Classifier & WRN70-16 & WRN28-10 & WRN28-10\\
      \hline
      FGSM~\cite{fgsm} &  \textbf{75.66} & 70.91 & 64.08\\
      PGD~\cite{pgd} & 69.93 & 66.02  & \textbf{82.99}\\
      BIM~\cite{bim} & 69.84 & 65.95  & \textbf{80.46}\\
      RFGSM~\cite{rfgsm} & 69.8 & 65.95  & \textbf{81.2}\\
      EotPgd~\cite{eotpgd} & 70.68 & 66.58  & \textbf{76.84}\\
      TPgd~\cite{zhang2019theoretically} & \textbf{82.32} & 80.48 &  81.61\\
      FFgsm~\cite{ffgsm} & \textbf{78.04} & 73.37  & 70.1\\
      MiFgsm~\cite{mifgsm} & \textbf{73.22} & 68.82 &  45.49 \\
      APgd~\cite{apgd} & 69.46 & 65.69 &  \textbf{85.79} \\
      Jitter~\cite{jitter} & 70.15 & 64.84 &  \textbf{80.49} \\
      \hline
      Avg. & 72.72 & 68.69  &  \textbf{75.88}  \\
     \hline
    \end{tabular}
    }
\end{minipage}%
\hspace{.5pt}
\begin{minipage}{.39\linewidth}
    \begin{tabular}{c}
        \includegraphics[width=\linewidth]{fig/defense_transfer_linf/cifar10_defense_transfer.pdf} 
    \end{tabular}
    \captionof{figure}{Defense Transfer across $L_{\infty}$ attacks on Cifar10.}
    \label{fig:cifar10_defense_transfer_linf}
\end{minipage}
    
\end{table}

\begin{table}[htb]
\begin{minipage}{.6\linewidth}
    \captionof{table}{Defense Transfer across $L_{\infty}$ attacks ($\epsilon_{\infty}=8/255$) on Cifar100.}
    \label{tab:cifar100_defense_transfer_linf_appendix}
    \adjustbox{max width=1\linewidth}{%
    \begin{tabular}{|c|cc|c|}
    \hline
      Method & Gowal et al.~\cite{gowal2020uncovering}  &  Rebuffi et al.~\cite{Rebuffi2021FixingDA}  & DISCO\\
      Classifier & WRN70-16 & WRN28-10 & WRN28-10\\
      \hline
      FGSM~\cite{fgsm} &  44.53 & 38.57 & \textbf{50.4}\\
      PGD~\cite{pgd} & 40.46 & 36.09  & \textbf{74.51}\\
      BIM~\cite{bim} & 40.38 & 36.03  & \textbf{72.25}\\
      RFGSM~\cite{rfgsm} & 40.42 & 35.99  & \textbf{72.1}\\
      EotPgd~\cite{eotpgd} & 41.07 & 36.45  & \textbf{74.8}\\
      TPgd~\cite{zhang2019theoretically} & 57.52 & 52.01 &  \textbf{74.06}\\
      FFgsm~\cite{ffgsm} & 47.61 & 41.47  & \textbf{64.29}\\
      MiFgsm~\cite{mifgsm} & 42.37 & 37.31 &  \textbf{44.14} \\
      APgd~\cite{apgd} & 39.99 & 35.64 &  \textbf{77.33} \\
      Jitter~\cite{jitter} & 38.38 & 33.04 &  \textbf{73.75} \\
      \hline
      Avg. & 43.27 & 38.26  &  \textbf{67.76}  \\
     \hline
    \end{tabular}
    }
\end{minipage}%
\hspace{.5pt}
\begin{minipage}{.39\linewidth}
    \begin{tabular}{c}
        \includegraphics[width=\linewidth]{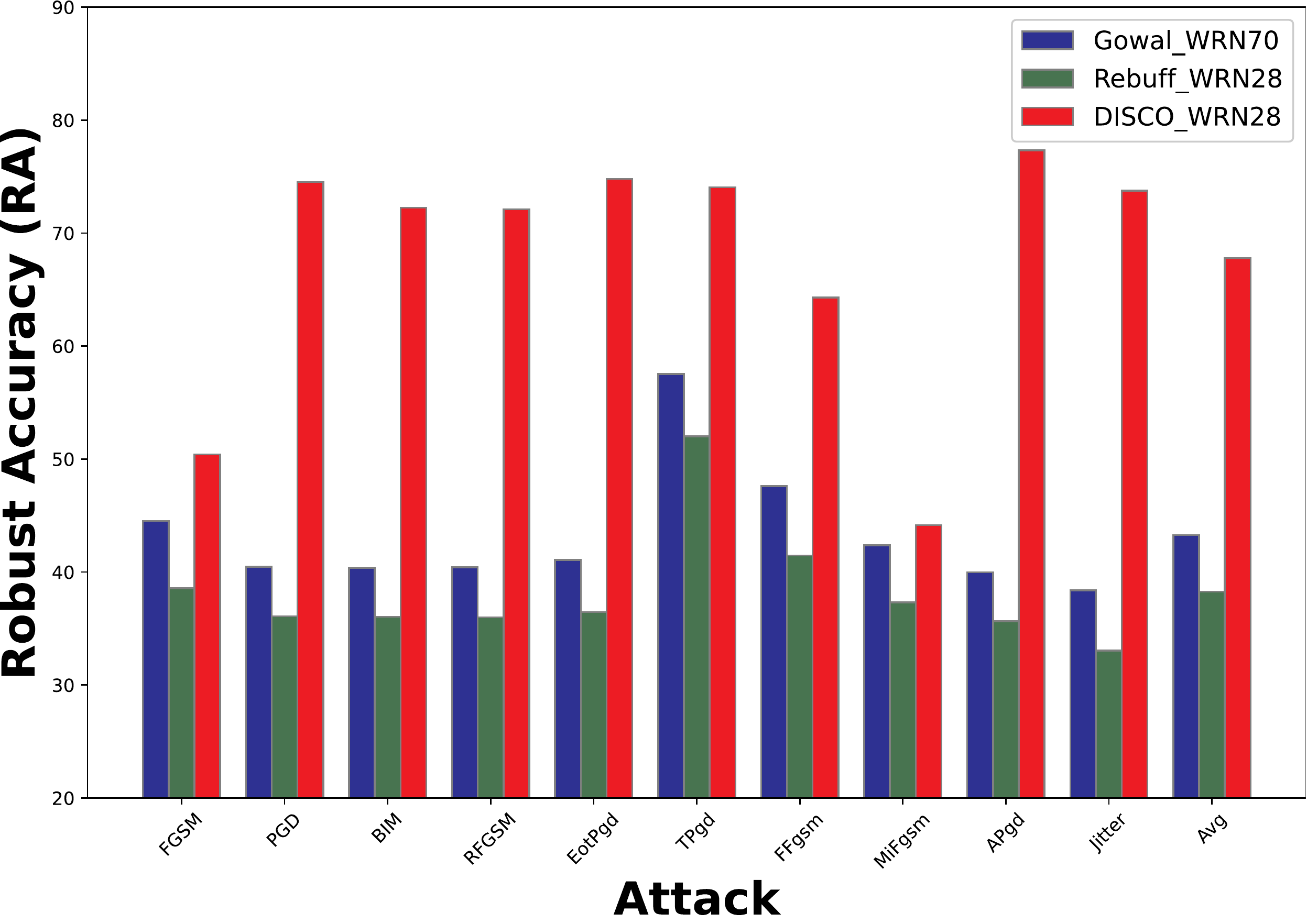} 
    \end{tabular}
    \captionof{figure}{Defense Transfer across $L_{\infty}$ attacks on Cifar100.}
    \label{fig:cifar100_defense_transfer_linf}
\end{minipage}
    
\end{table}

\begin{table}[htb]
\begin{minipage}{.6\linewidth}
    \captionof{table}{Defense Transfer across $L_{\infty}$ attacks ($\epsilon_{\infty}=4/255$) on ImageNet.}
    \label{tab:imagenet_defense_transfer_linf}
    \adjustbox{max width=1\linewidth}{%
    \begin{tabular}{|c|cc|c|}
    \hline
      Method & Hadi et al.~\cite{NEURIPS2020_24357dd0} & Engstrom et al.~\cite{robustness} & DISCO\\
      Classifier & Res50 & Res50 & Res50\\
      \hline
      Clean  & 64.1 & 62.54 & \textbf{72.64} \\
      FGSM~\cite{fgsm}   & 43.48 & 39.96 & \textbf{55.72} \\
      PGD~\cite{pgd}    & 39.28 & 33.32 & \textbf{66.32} \\
      BIM~\cite{bim}    & 39.26 & 33.2 & \textbf{66.4}\\
      RFGSM~\cite{rfgsm}  & 39.28 & 33.16 & \textbf{66.4}\\
      EotPgd~\cite{eotpgd} & 41.2 & 37.24 & \textbf{69.32}\\
      TPgd~\cite{zhang2019theoretically}   & 53.82 & 49.64 & \textbf{69.94}\\
      FFgsm~\cite{ffgsm}  & 43.58 & 40.1 & \textbf{57}\\
      MiFgsm~\cite{mifgsm} & 40.56 & 35.6 & \textbf{52.38}\\
      APgd~\cite{apgd}   & 38.42 & 32.22 & \textbf{68.3}\\
      Jitter~\cite{jitter} & 36.26 & 31.36 & \textbf{67.04} \\
      \hline 
      Avg. &  41.51  & 36.58 & \textbf{63.88}  \\
     \hline
    \end{tabular}
    }
\end{minipage}%
\hspace{.5pt}
\begin{minipage}{.39\linewidth}
    \begin{tabular}{c}
        \includegraphics[width=\linewidth]{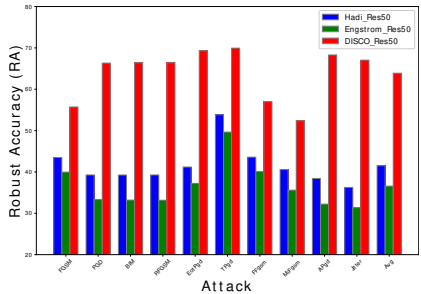} 
    \end{tabular}
    \captionof{figure}{Defense Transfer across $L_{\infty}$ attacks on ImageNet.}
    \label{fig:imagenet_defense_transfer_linf}
\end{minipage}
\end{table}

\section{Improving Cifar10 and Cifar100 SOTA on RobustBench}

\begin{table}[htb]
 \caption{Improving SOTA defenses on RobustBench~\cite{croce2021robustbench} for Cifar10 ($L_{2}$ and $L_{\infty}$) and Cifar100 ($L_{\infty}$) dataset.}
\label{tab:improving_sota_cifar}
\centering
\adjustbox{max width=1\linewidth}{%
    \begin{tabular}{|ccc|ccc|}
    \hline
     Method & Dataset & Norm & SA & RA & Avg.\\
     \hline
     Rebuffi et al.~\cite{Rebuffi2021FixingDA} & Cifar10 & $L_{\infty}$  & \textbf{92.23} & 66.58 & 79.41 \\
     w/ DISCO & Cifar10 & $L_{\infty}$  & 91.95 & \textbf{70.71} & \textbf{81.33}\\
     \hline
     Rebuffi et al.~\cite{Rebuffi2021FixingDA} & Cifar10 & $L_{2}$ &  \textbf{95.74} & 82.32 & 89.03
     \\
     w/ DISCO & Cifar10 & $L_{2}$  & 95.24 & \textbf{84.15} & \textbf{89.7}\\
     \hline
     Gowal et al.~\cite{gowal2020uncovering} & Cifar100 & $L_{\infty}$ & \textbf{69.15} & 36.88 & 53.02 \\
     w/ DISCO & Cifar100 & $L_{\infty}$  &
     68.56 & \textbf{39.77} & \textbf{54.17}\\
     \hline
    \end{tabular}
    }

\end{table}%

Sec. 4.1 in the main paper shows that DISCO can improve the prior SOTA defenses on the ImageNet dataset. In Table~\ref{tab:improving_sota_cifar}, we further investigate the gain of applying DISCO on SOTA Cifar10 and Cifar100 defenses. The first and second block of Table~\ref{tab:improving_sota_cifar} show the gains of applying DISCO on \cite{Rebuffi2021FixingDA}, which is the prior SOTA defense against $L_2$ and $L_{\infty}$ Autoattack on Cifar10. DISCO also improves the prior SOTA defense~\cite{gowal2020uncovering} on Cifar100 by 2.89\%. These results indicate that, beyond being a robust defense by itself, DISCO can also be applied to existing defenses to improve their robustness.

\section{Kernel Size s}
\begin{table}[htb]
 \caption{Ablation on various kernel size $s$.}
\label{tab:kernel_size}
\centering
\adjustbox{max width=1\linewidth}{%
    \begin{tabular}{|c|ccc|}
    \hline
     s & SA & RA & Avg.\\
     \hline
     1  & 71.22 & \textbf{69.52} & 70.37\\
     3  & 72.64 & 68.2 & \textbf{70.42}\\
     5  & \textbf{74.22} & 60.1 & 67.16\\
     \hline
    \end{tabular}
    }
\end{table}%

In this section, we ablate the kernel size used to train DISCO on ImageNet. The kernel size $s$ controls the feature neighborhood forwarded to the local implicit module. Table~\ref{tab:kernel_size} shows that $s=3$ achieves the best performance, which degrades for $s=5$ by a significant margin (3.26\%). This shows that while tasks like classification require large and global receptive fields, the projection of adversarial images into the natural image manifold can be done on small neighborhoods. Given that the complexity of modeling the manifold increases with the neighborhood size, it is not surprising that larger $s$ lead to weaker performance. This is consistent with the well known complexity of synthesizing images with global models, such as GANs. What is somewhat surprising is that even $s=1$ is sufficient to enable a robust defense. By default, we use $s=3$ in all our experiments.

\section{Computation Time for STL and DISCO}
\begin{table}[htb]
\caption{Computation time between of STL~\cite{stl} and DISCO for different image sizes. Note that STL requires   a 36.34$\times$ larger inference time when image size increases from 32 to 224.}
\label{tab:time_stl_disco}
\centering
\adjustbox{max width=1\linewidth}{%
    \begin{tabular}{|cc||c|ccccc|}
    \hline
     \multirow{2}{*}{Dataset} & Image & \multirow{2}{*}{STL~\cite{stl}} &  \multicolumn{5}{c|}{DISCO} \\
     & Size & & (K=1) & (K=2) & (K=3) & (K=4) & (K=5)\\
     \hline
     Cifar10 & 32 & 0.65 & 0.011 & 0.021 & 0.031 & 0.037 & 0.048\\
     ImageNet & 224 & 23.71 & 0.027 & 0.081 & 0.134 & 0.191 & 0.251\\
     \hline
     \multicolumn{2}{|c||}{Time Increase} & $\times$36.34 & $\times$2.41 & $\times$3.86 & $\times$4.35 & $\times$5.14 & $\times$5.19\\
     \hline
    \end{tabular}
    }
 
\end{table}

Table~\ref{tab:time_stl_disco} compares the inference time of STL~\cite{stl}, DISCO and cascade DISCO (from $K=$ 2 to 5) on Cifar10 and ImageNet. For a single image Cifar10 of size 32x32, STL requires an Cifar10 5.9$\times$ (0.65 vs 0.011)  larger than that of DISCO ($K$=1). When cascade DISCO is used, inference time increases approximately linearly with $K$. 

For a single ImageNet image of size 224, STL requires 23.71 seconds while DISCO (K=1) only requires 0.027. The inference time difference increases to $878.15\times$ (23.71 vs 0.027) on ImageNet , which is significantly larger than that of Cifar10 (5.9$\times$). This shows that DISCO is a better defense in the sense that it can handle widely varying input image sizes with minor variations of computing cost.

\section{Training Details}
On Cifar10 and Cifar100, we train the DISCO for 40 epochs. On ImageNet, DISCO is only trained for 3 epochs because ImageNet images are larger and produce more random crops. The learning rate is set to 0.0001 and the Adam optimizer is used in all experiments. All experiments are conducted using Pytorch~\cite{pytorch}. All time measurements, for both baselines and DISCO, are made on a single Nvidia Titan Xp GPU with Intel Xeon CPU E5-2630, with batch size 1 and averaged over 100 images.

\section{Adopted Code and Benchmark}
In this section, we list the url links that are used for training and evaluating DISCO. To create the adversarial-clean training pairs, we adopt the code from 
TorchAttack\footnote{\url{https://adversarial-attacks-pytorch.readthedocs.io/en/latest/}}
and Ares\footnote{\label{ares}\url{https://github.com/thu-ml/ares}}, which support the multiple attack methods. These attack methods are then used to attack pretrained classifiers on Cifar10, Cifar100 and ImageNet. We use the ResNet18 classifiers from Ares for Cifar10, the WideResNet  Cifar100 classifiers from this repository \footnote{\url{https://github.com/xternalz/WideResNet-pytorch}} and the ResNet18  ImageNet classifiers of Pytorch~\cite{pytorch}.

To evaluate DISCO, we adopt Autoattack from RobustBench~\cite{croce2021robustbench}\footnote{\url{https://github.com/RobustBench/robustbench}} and compare to the pretrained defenses on the RobustBench leaderboard. In addition to Autoattack, we use the AdverTorch\footnote{\url{https://github.com/BorealisAI/advertorch}} library to implement the BPDA attack~\cite{bpda} and the TorchAttack\footnote{\url{https://adversarial-attacks-pytorch.readthedocs.io/en/latest/}} library for other attacks, like FGSM~\cite{fgsm} and BIM~\cite{bim}.

For the adversarially trained defense baselines, we adopt the pretrained weights from
RobustBench~\cite{croce2021robustbench}\footnote{\url{https://github.com/RobustBench/robustbench}}, while the codes for transformation based baselines are adopted from 
Ares, Cifar autoencoder 
\footnote{\url{https://github.com/chenjie/PyTorch-CIFAR-10-autoencoder}} and STL~\cite{stl}. To implement DISCO, we use code from LIIF\footnote{\url{https://github.com/yinboc/liif}}~\cite{chen2021learning}.

\section{Visualizations}
\begin{figure}
    \centering
\begin{tabular}{cccc}
    {\begin{turn}{90}Clean \end{turn}} 
    & \includegraphics[width=0.3\linewidth]{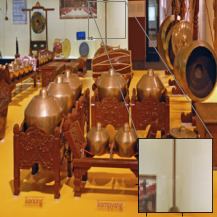} &  \includegraphics[width=0.3\linewidth]{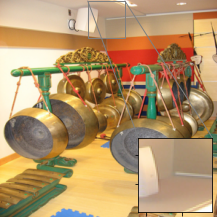} &  \includegraphics[width=0.3\linewidth]{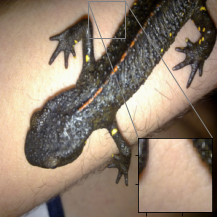} \\
    {\begin{turn}{90}Adversarial \end{turn}} &\includegraphics[width=0.3\linewidth]{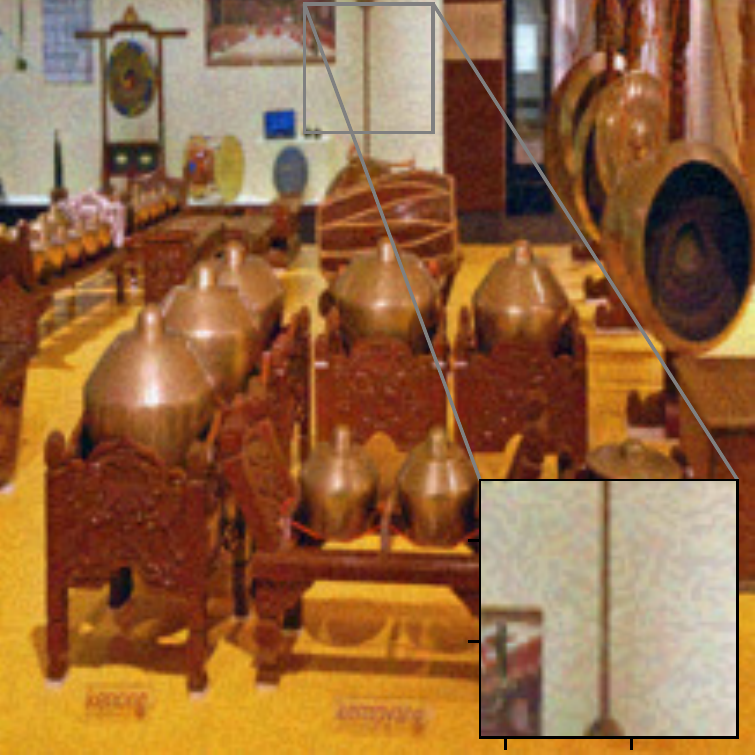} &
    \includegraphics[width=0.3\linewidth]{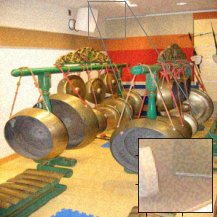} &
    \includegraphics[width=0.3\linewidth]{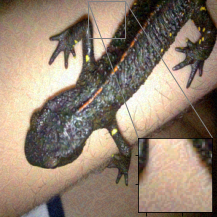} \\
    {\begin{turn}{90}DISCO (K=1) \end{turn}} &
    \includegraphics[width=0.3\linewidth]{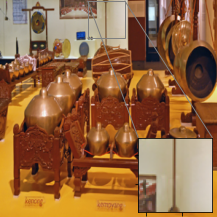} & 
    \includegraphics[width=0.3\linewidth]{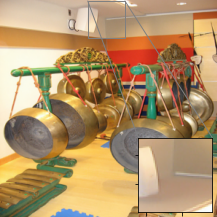} & 
    \includegraphics[width=0.3\linewidth]{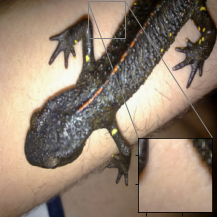} 
    \\
    {\begin{turn}{90}DISCO (K=2) \end{turn}} &
    \includegraphics[width=0.3\linewidth]{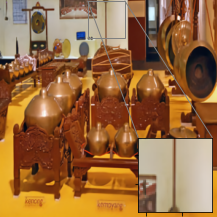} & 
    \includegraphics[width=0.3\linewidth]{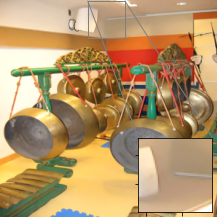} & 
    \includegraphics[width=0.3\linewidth]{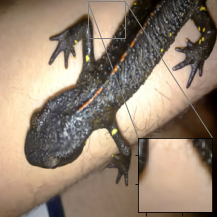}  \\
    
   {\begin{turn}{90}DISCO (K=3) \end{turn}} &
   \includegraphics[width=0.3\linewidth]{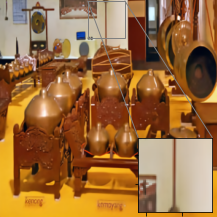} & 
    \includegraphics[width=0.3\linewidth]{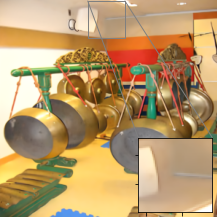} & 
    \includegraphics[width=0.3\linewidth]{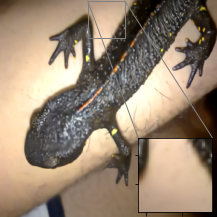} 
   \\
\end{tabular}
\caption{Comparison of Clean image, Adversarial image and DISCO output from $K=$ 1 to 3 under FGSM attack.}
\label{fig:fgsm_disco_output}
\end{figure}

\begin{figure}
    \centering
\begin{tabular}{cccc}
    {\begin{turn}{90}Clean \end{turn}} 
    & \includegraphics[width=0.3\linewidth]{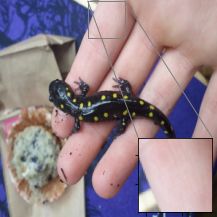} &  \includegraphics[width=0.3\linewidth]{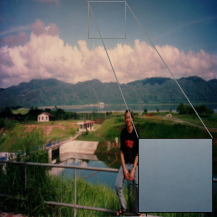} &  \includegraphics[width=0.3\linewidth]{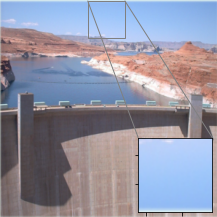} \\
    {\begin{turn}{90}Adversarial \end{turn}} &\includegraphics[width=0.3\linewidth]{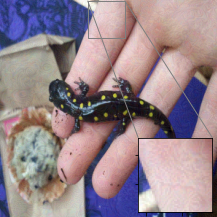} &
    \includegraphics[width=0.3\linewidth]{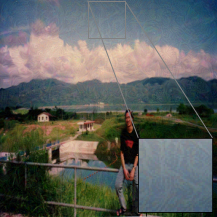} &
    \includegraphics[width=0.3\linewidth]{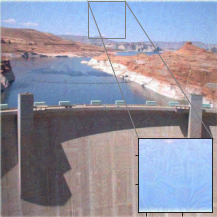} \\
    {\begin{turn}{90}DISCO (K=1) \end{turn}} &
    \includegraphics[width=0.3\linewidth]{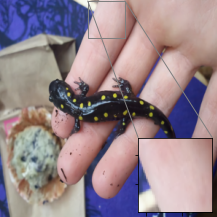} & 
    \includegraphics[width=0.3\linewidth]{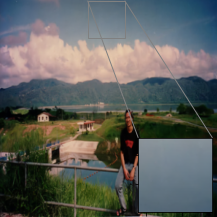} & 
    \includegraphics[width=0.3\linewidth]{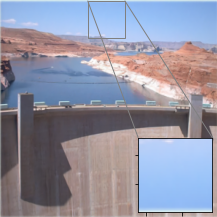} 
    \\
    {\begin{turn}{90}DISCO (K=2) \end{turn}} &
    \includegraphics[width=0.3\linewidth]{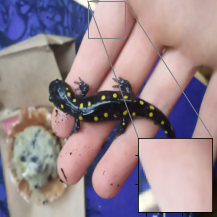} & 
    \includegraphics[width=0.3\linewidth]{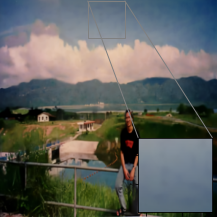} & 
    \includegraphics[width=0.3\linewidth]{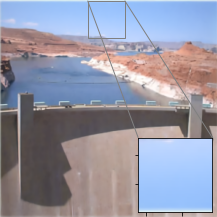}  \\
    
   {\begin{turn}{90}DISCO (K=3) \end{turn}} &
   \includegraphics[width=0.3\linewidth]{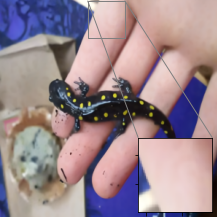} & 
    \includegraphics[width=0.3\linewidth]{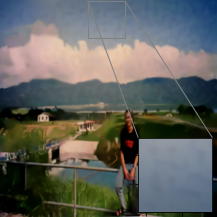} & 
    \includegraphics[width=0.3\linewidth]{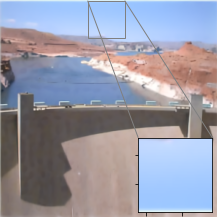} 
   \\
\end{tabular}
\caption{Comparison of Clean image, Adversarial image and DISCO output from $K=$ 1 to 3 under BIM attack.}
\label{fig:bim_disco_output}
\end{figure}

\begin{figure}
    \centering
\begin{tabular}{cccc}
    {\begin{turn}{90}Clean \end{turn}} 
    & \includegraphics[width=0.3\linewidth]{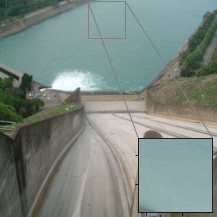} &  \includegraphics[width=0.3\linewidth]{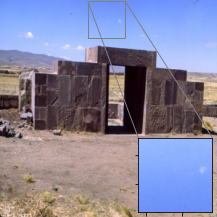} &  \includegraphics[width=0.3\linewidth]{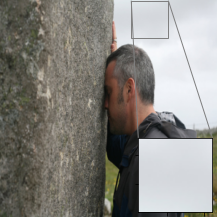} \\
    {\begin{turn}{90}Adversarial \end{turn}} &\includegraphics[width=0.3\linewidth]{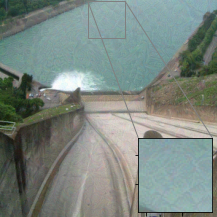} &
    \includegraphics[width=0.3\linewidth]{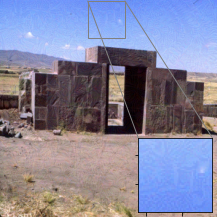} &
    \includegraphics[width=0.3\linewidth]{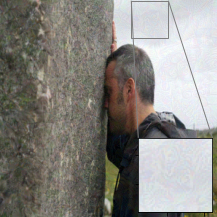} \\
    {\begin{turn}{90}DISCO (K=1) \end{turn}} &
    \includegraphics[width=0.3\linewidth]{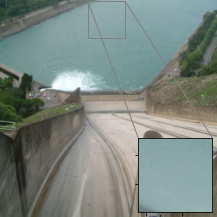} & 
    \includegraphics[width=0.3\linewidth]{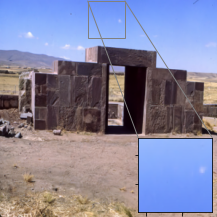} & 
    \includegraphics[width=0.3\linewidth]{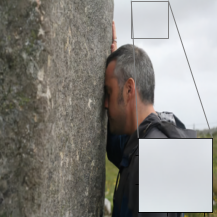} 
    \\
    {\begin{turn}{90}DISCO (K=2) \end{turn}} &
    \includegraphics[width=0.3\linewidth]{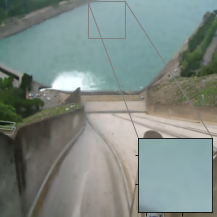} & 
    \includegraphics[width=0.3\linewidth]{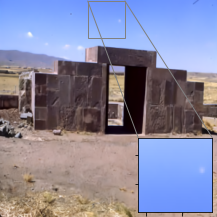} & 
    \includegraphics[width=0.3\linewidth]{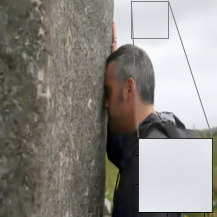}  \\
    
   {\begin{turn}{90}DISCO (K=3) \end{turn}} &
   \includegraphics[width=0.3\linewidth]{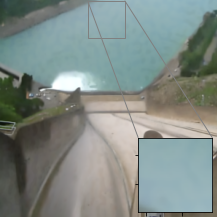} & 
    \includegraphics[width=0.3\linewidth]{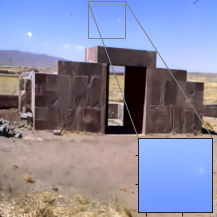} & 
    \includegraphics[width=0.3\linewidth]{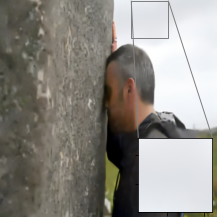} 
   \\
\end{tabular}
\caption{Comparison of Clean image, Adversarial image and DISCO output from $K=$ 1 to 3 under PGD attack.}
\label{fig:pgd_disco_output}
\end{figure}

DISCO defense outputs against FGSM~\cite{fgsm} and BIM~\cite{bim} and PGD~\cite{pgd} attacks are visualized in Fig.~\ref{fig:fgsm_disco_output}, \ref{fig:bim_disco_output} and \ref{fig:pgd_disco_output}, respectively. Take Fig.~\ref{fig:fgsm_disco_output} for example. The first and second rows show the clean and adversarial images, while rows 3-5 show the output of DISCO and cascade DISCO ($K=2$ and $K=3$). Clearly, both DISCO and its cascade version can effectively remove the adversarial perturbation. Note that these images are produced from the same DISCO model without retraining for any attack.
%In addition, Fig.~\ref{fig:multisize_disco_output} shows the DISCO output for various images sizes, from 128x128 to 512x512. Note that these images are produced from the same DISCO model without retraining for any output size or attack.

\clearpage
\bibliography{ref}{}

\begin{thebibliography}{100}

\bibitem{addepalli2021towards}
Sravanti Addepalli, Samyak Jain, Gaurang Sriramanan, Shivangi Khare, and
  Venkatesh~Babu Radhakrishnan.
\newblock Towards achieving adversarial robustness beyond perceptual limits.
\newblock In {\em ICML 2021 Workshop on Adversarial Machine Learning}, 2021.

\bibitem{addepalli2021oaat}
Sravanti Addepalli, Samyak Jain, Gaurang Sriramanan, Shivangi Khare, and
  Venkatesh~Babu Radhakrishnan.
\newblock Towards achieving adversarial robustness beyond perceptual limits.
\newblock In {\em ICML 2021 Workshop on Adversarial Machine Learning}, 2021.

\bibitem{9614158}
Naveed Akhtar, Ajmal Mian, Navid Kardan, and Mubarak Shah.
\newblock Advances in adversarial attacks and defenses in computer vision: A
  survey.
\newblock {\em IEEE Access}, 9:155161--155196, 2021.

\bibitem{NEURIPS2019_Alayrac}
Jean-Baptiste Alayrac, Jonathan Uesato, Po-Sen Huang, Alhussein Fawzi, Robert
  Stanforth, and Pushmeet Kohli.
\newblock Are labels required for improving adversarial robustness?
\newblock In H.~Wallach, H.~Larochelle, A.~Beygelzimer, F.~d\textquotesingle
  Alch\'{e}-Buc, E.~Fox, and R.~Garnett, editors, {\em Advances in Neural
  Information Processing Systems}, volume~32. Curran Associates, Inc., 2019.

\bibitem{Alfarra_Perez_Thabet_Bibi_Torr_Ghanem_2022}
Motasem Alfarra, Juan~C. Perez, Ali Thabet, Adel Bibi, Philip~H.S. Torr, and
  Bernard Ghanem.
\newblock Combating adversaries with anti-adversaries.
\newblock {\em Proceedings of the AAAI Conference on Artificial Intelligence},
  36(6):5992--6000, Jun. 2022.

\bibitem{squareattack}
Maksym Andriushchenko, Francesco Croce, Nicolas Flammarion, and Matthias Hein.
\newblock Square attack: A query-efficient black-box adversarial attack via
  random search.
\newblock In Andrea Vedaldi, Horst Bischof, Thomas Brox, and Jan-Michael Frahm,
  editors, {\em Computer Vision -- ECCV 2020}, pages 484--501, Cham, 2020.
  Springer International Publishing.

\bibitem{andriushchenko2020understanding}
Maksym Andriushchenko and Nicolas Flammarion.
\newblock Understanding and improving fast adversarial training.
\newblock In {\em NeurIPS}, 2020.

\bibitem{Ali20}
Ali ArjomandBigdeli, Maryam Amirmazlaghani, and Mohammad Khalooei.
\newblock Defense against adversarial attacks using dragan.
\newblock In {\em 2020 6th Iranian Conference on Signal Processing and
  Intelligent Systems (ICSPIS)}, pages 1--5, 2020.

\bibitem{bpda}
Anish Athalye, Nicholas Carlini, and David~A. Wagner.
\newblock Obfuscated gradients give a false sense of security: Circumventing
  defenses to adversarial examples.
\newblock In {\em ICML}, pages 274--283, 2018.

\bibitem{atzmon2019controlling}
Matan Atzmon, Niv Haim, Lior Yariv, Ofer Israelov, Haggai Maron, and Yaron
  Lipman.
\newblock Controlling neural level sets.
\newblock In {\em Advances in Neural Information Processing Systems}, pages
  2032--2041, 2019.

\bibitem{Augustin_eccv20}
Maximilian Augustin, Alexander Meinke, and Matthias Hein.
\newblock Adversarial robustness on in- and out-distribution improves
  explainability.
\newblock In Andrea Vedaldi, Horst Bischof, Thomas Brox, and Jan-Michael Frahm,
  editors, {\em Computer Vision -- ECCV 2020}, pages 228--245, Cham, 2020.
  Springer International Publishing.

\bibitem{Bai2019HilbertBasedGD}
Yang Bai, Yan Feng, Yisen Wang, Tao Dai, Shutao Xia, and Yong Jiang.
\newblock Hilbert-based generative defense for adversarial examples.
\newblock {\em 2019 IEEE/CVF International Conference on Computer Vision
  (ICCV)}, pages 4783--4792, 2019.

\bibitem{barlow}
H.~B. Barlow.
\newblock The exploitation of regularities in the environment by the brain.
\newblock {\em The Behavioral and brain sciences}, 24 4:602--7; discussion
  652--71, 2001.

\bibitem{BELL19973327}
Anthony~J. Bell and Terrence~J. Sejnowski.
\newblock The “independent components” of natural scenes are edge filters.
\newblock {\em Vision Research}, 37(23):3327--3338, 1997.

\bibitem{biggan}
Andrew Brock, Jeff Donahue, and Karen Simonyan.
\newblock Large scale {GAN} training for high fidelity natural image synthesis.
\newblock In {\em International Conference on Learning Representations}, 2019.

\bibitem{Brock2019LargeSG}
Andrew Brock, Jeff Donahue, and Karen Simonyan.
\newblock Large scale gan training for high fidelity natural image synthesis.
\newblock {\em ArXiv}, abs/1809.11096, 2019.

\bibitem{Brown2017AdversarialP}
Tom~B. Brown, Dandelion Man{\'e}, Aurko Roy, Mart{\'i}n Abadi, and Justin
  Gilmer.
\newblock Adversarial patch.
\newblock {\em ArXiv}, abs/1712.09665, 2017.

\bibitem{cw}
Nicholas Carlini and David~A. Wagner.
\newblock Towards evaluating the robustness of neural networks.
\newblock {\em 2017 IEEE Symposium on Security and Privacy (SP)}, pages 39--57,
  2017.

\bibitem{NEURIPS2019_Carmon}
Yair Carmon, Aditi Raghunathan, Ludwig Schmidt, John~C Duchi, and Percy~S
  Liang.
\newblock Unlabeled data improves adversarial robustness.
\newblock In H.~Wallach, H.~Larochelle, A.~Beygelzimer, F.~d\textquotesingle
  Alch\'{e}-Buc, E.~Fox, and R.~Garnett, editors, {\em Advances in Neural
  Information Processing Systems}, volume~32. Curran Associates, Inc., 2019.

\bibitem{Chakraborty2018AdversarialAA}
Anirban Chakraborty, Manaar Alam, Vishal Dey, Anupam Chattopadhyay, and Debdeep
  Mukhopadhyay.
\newblock Adversarial attacks and defences: A survey.
\newblock {\em ArXiv}, abs/1810.00069, 2018.

\bibitem{Chan2020Jacobian}
Alvin Chan, Yi~Tay, Yew~Soon Ong, and Jie Fu.
\newblock Jacobian adversarially regularized networks for robustness.
\newblock In {\em International Conference on Learning Representations}, 2020.

\bibitem{Chen21}
Erh-Chung Chen and Che-Rung Lee.
\newblock Ltd: Low temperature distillation for robust adversarial training,
  2021.

\bibitem{chen2021efficient}
Jinghui Chen, Yu~Cheng, Zhe Gan, Quanquan Gu, and Jingjing Liu.
\newblock Efficient robust training via backward smoothing, 2021.

\bibitem{Chen_2020_CVPR}
Tianlong Chen, Sijia Liu, Shiyu Chang, Yu~Cheng, Lisa Amini, and Zhangyang
  Wang.
\newblock Adversarial robustness: From self-supervised pre-training to
  fine-tuning.
\newblock In {\em The IEEE/CVF Conference on Computer Vision and Pattern
  Recognition (CVPR)}, June 2020.

\bibitem{chen2021learning}
Yinbo Chen, Sifei Liu, and Xiaolong Wang.
\newblock Learning continuous image representation with local implicit image
  function.
\newblock In {\em Proceedings of the IEEE/CVF Conference on Computer Vision and
  Pattern Recognition}, pages 8628--8638, 2021.

\bibitem{Chen2019LearningIF}
Zhiqin Chen and Hao Zhang.
\newblock Learning implicit fields for generative shape modeling.
\newblock {\em 2019 IEEE/CVF Conference on Computer Vision and Pattern
  Recognition (CVPR)}, pages 5932--5941, 2019.

\bibitem{fab}
F.~Croce and M.~Hein.
\newblock Minimally distorted adversarial examples with a fast adaptive
  boundary attack.
\newblock In {\em ICML}, 2020.

\bibitem{croce2021robustbench}
Francesco Croce, Maksym Andriushchenko, Vikash Sehwag, Edoardo Debenedetti,
  Nicolas Flammarion, Mung Chiang, Prateek Mittal, and Matthias Hein.
\newblock {RobustBench: a standardized adversarial robustness benchmark}.
\newblock In {\em Thirty-fifth Conference on Neural Information Processing
  Systems Datasets and Benchmarks Track}, 2021.

\bibitem{croce22a}
Francesco Croce, Sven Gowal, Thomas Brunner, Evan Shelhamer, Matthias Hein, and
  Taylan Cemgil.
\newblock Evaluating the adversarial robustness of adaptive test-time defenses.
\newblock In Kamalika Chaudhuri, Stefanie Jegelka, Le~Song, Csaba Szepesvari,
  Gang Niu, and Sivan Sabato, editors, {\em Proceedings of the 39th
  International Conference on Machine Learning}, volume 162 of {\em Proceedings
  of Machine Learning Research}, pages 4421--4435. PMLR, 17--23 Jul 2022.

\bibitem{autoattack}
Francesco Croce and Matthias Hein.
\newblock Reliable evaluation of adversarial robustness with an ensemble of
  diverse parameter-free attacks.
\newblock In {\em ICML}, 2020.

\bibitem{apgd}
Francesco Croce and Matthias Hein.
\newblock Reliable evaluation of adversarial robustness with an ensemble of
  diverse parameter-free attacks.
\newblock In {\em ICML}, 2020.

\bibitem{cui2020learnable}
Jiequan Cui, Shu Liu, Liwei Wang, and Jiaya Jia.
\newblock Learnable boundary guided adversarial training.
\newblock {\em arXiv preprint arXiv:2011.11164}, 2020.

\bibitem{Cui2021LearnableBG}
Jiequan Cui, Shu Liu, Liwei Wang, and Jiaya Jia.
\newblock Learnable boundary guided adversarial training.
\newblock {\em 2021 IEEE/CVF International Conference on Computer Vision
  (ICCV)}, pages 15701--15710, 2021.

\bibitem{Sihui_corr21}
Sihui Dai, Saeed Mahloujifar, and Prateek Mittal.
\newblock Parameterizing activation functions for adversarial robustness.
\newblock {\em CoRR}, abs/2110.05626, 2021.

\bibitem{Das2017KeepingTB}
Nilaksh Das, Madhuri Shanbhogue, Shang-Tse Chen, Fred Hohman, Li~Chen,
  Michael~E. Kounavis, and Duen~Horng Chau.
\newblock Keeping the bad guys out: Protecting and vaccinating deep learning
  with jpeg compression.
\newblock {\em ArXiv}, abs/1705.02900, 2017.

\bibitem{Demontis2019WhyDA}
Ambra Demontis, Marco Melis, Maura Pintor, Matthew Jagielski, Battista Biggio,
  Alina Oprea, Cristina Nita-Rotaru, and Fabio Roli.
\newblock Why do adversarial attacks transfer? explaining transferability of
  evasion and poisoning attacks.
\newblock In {\em USENIX Security Symposium}, 2019.

\bibitem{deng2009imagenet}
Jia Deng, Wei Dong, Richard Socher, Li-Jia Li, Kai Li, and Li~Fei-Fei.
\newblock Imagenet: A large-scale hierarchical image database.
\newblock In {\em 2009 IEEE conference on computer vision and pattern
  recognition}, pages 248--255. Ieee, 2009.

\bibitem{Ding2020MMA}
Gavin~Weiguang Ding, Yash Sharma, Kry Yik~Chau Lui, and Ruitong Huang.
\newblock Mma training: Direct input space margin maximization through
  adversarial training.
\newblock In {\em International Conference on Learning Representations}, 2020.

\bibitem{mifgsm}
Yinpeng Dong, Fangzhou Liao, Tianyu Pang, Hang Su, Jun Zhu, Xiaolin Hu, and
  Jianguo Li.
\newblock Boosting adversarial attacks with momentum.
\newblock {\em 2018 IEEE/CVF Conference on Computer Vision and Pattern
  Recognition}, pages 9185--9193, 2018.

\bibitem{tifgsm}
Yinpeng Dong, Tianyu Pang, Hang Su, and Jun Zhu.
\newblock Evading defenses to transferable adversarial examples by
  translation-invariant attacks.
\newblock In {\em Proceedings of the IEEE Computer Society Conference on
  Computer Vision and Pattern Recognition}, 2019.

\bibitem{dupont2021coin}
Emilien Dupont, Adam Golinski, Milad Alizadeh, Yee~Whye Teh, and Arnaud Doucet.
\newblock {COIN}: {CO}mpression with implicit neural representations.
\newblock In {\em Neural Compression: From Information Theory to Applications
  -- Workshop @ ICLR 2021}, 2021.

\bibitem{Dziugaite2016ASO}
Gintare~Karolina Dziugaite, Zoubin Ghahramani, and Daniel~M. Roy.
\newblock A study of the effect of jpg compression on adversarial images.
\newblock {\em ArXiv}, abs/1608.00853, 2016.

\bibitem{robustness}
Logan Engstrom, Andrew Ilyas, Hadi Salman, Shibani Santurkar, and Dimitris
  Tsipras.
\newblock Robustness (python library), 2019.

\bibitem{Field1987RelationsBT}
David~J. Field.
\newblock Relations between the statistics of natural images and the response
  properties of cortical cells.
\newblock {\em Journal of the Optical Society of America. A, Optics and image
  science}, 4 12:2379--94, 1987.

\bibitem{Field1994WhatIT}
David~J. Field.
\newblock What is the goal of sensory coding?
\newblock {\em Neural Computation}, 6:559--601, 1994.

\bibitem{Genova2020LocalDI}
Kyle Genova, Forrester Cole, Avneesh Sud, Aaron Sarna, and Thomas~A.
  Funkhouser.
\newblock Local deep implicit functions for 3d shape.
\newblock {\em 2020 IEEE/CVF Conference on Computer Vision and Pattern
  Recognition (CVPR)}, pages 4856--4865, 2020.

\bibitem{gilmer2018adversarial}
Justin Gilmer, Luke Metz, Fartash Faghri, Sam Schoenholz, Maithra Raghu, Martin
  Wattenberg, and Ian Goodfellow.
\newblock Adversarial spheres, 2018.

\bibitem{fgsm}
Ian Goodfellow, Jonathon Shlens, and Christian Szegedy.
\newblock Explaining and harnessing adversarial examples.
\newblock In {\em International Conference on Learning Representations}, 2015.

\bibitem{gowal2020uncovering}
Sven Gowal, Chongli Qin, Jonathan Uesato, Timothy Mann, and Pushmeet Kohli.
\newblock Uncovering the limits of adversarial training against norm-bounded
  adversarial examples.
\newblock {\em arXiv preprint arXiv:2010.03593}, 2020.

\bibitem{NEURIPS2021_Gowal}
Sven Gowal, Sylvestre-Alvise Rebuffi, Olivia Wiles, Florian Stimberg,
  Dan~Andrei Calian, and Timothy~A Mann.
\newblock Improving robustness using generated data.
\newblock In M.~Ranzato, A.~Beygelzimer, Y.~Dauphin, P.S. Liang, and J.~Wortman
  Vaughan, editors, {\em Advances in Neural Information Processing Systems},
  volume~34, pages 4218--4233. Curran Associates, Inc., 2021.

\bibitem{field}
Daniel~J. Graham and David~J. Field.
\newblock Statistical regularities of art images and natural scenes: spectra,
  sparseness and nonlinearities.
\newblock {\em Spatial vision}, 21 1-2:149--64, 2007.

\bibitem{guo2018countering}
Chuan Guo, Mayank Rana, Moustapha Cisse, and Laurens van~der Maaten.
\newblock Countering adversarial images using input transformations.
\newblock In {\em International Conference on Learning Representations}, 2018.

\bibitem{guo2021adnerf}
Yudong Guo, Keyu Chen, Sen Liang, Yongjin Liu, Hujun Bao, and Juyong Zhang.
\newblock Ad-nerf: Audio driven neural radiance fields for talking head
  synthesis.
\newblock In {\em IEEE/CVF International Conference on Computer Vision (ICCV)},
  2021.

\bibitem{He2016DeepRL}
Kaiming He, X.~Zhang, Shaoqing Ren, and Jian Sun.
\newblock Deep residual learning for image recognition.
\newblock {\em 2016 IEEE Conference on Computer Vision and Pattern Recognition
  (CVPR)}, pages 770--778, 2016.

\bibitem{hendrycks2019pretraining}
Dan Hendrycks, Kimin Lee, and Mantas Mazeika.
\newblock Using pre-training can improve model robustness and uncertainty.
\newblock {\em Proceedings of the International Conference on Machine
  Learning}, 2019.

\bibitem{Ho2019CatastrophicCP}
Chih-Hui Ho, Brandon Leung, Erik Sandstr{\"o}m, Yen Chang, and Nuno
  Vasconcelos.
\newblock Catastrophic child's play: Easy to perform, hard to defend
  adversarial attacks.
\newblock {\em 2019 IEEE/CVF Conference on Computer Vision and Pattern
  Recognition (CVPR)}, pages 9221--9229, 2019.

\bibitem{senet}
Jie Hu, Li~Shen, and Gang Sun.
\newblock Squeeze-and-excitation networks.
\newblock In {\em 2018 IEEE/CVF Conference on Computer Vision and Pattern
  Recognition}, pages 7132--7141, 2018.

\bibitem{hu2021naturalistic}
Yu-Chih-Tuan Hu, Bo-Han Kung, Daniel~Stanley Tan, Jun-Cheng Chen, Kai-Lung Hua,
  and Wen-Huang Cheng.
\newblock Naturalistic physical adversarial patch for object detectors.
\newblock In {\em Proceedings of the IEEE/CVF International Conference on
  Computer Vision (ICCV)}, 2021.

\bibitem{huang2021exploring}
Hanxun Huang, Yisen Wang, Sarah~Monazam Erfani, Quanquan Gu, James Bailey, and
  Xingjun Ma.
\newblock Exploring architectural ingredients of adversarially robust deep
  neural networks.
\newblock In {\em NeurIPS}, 2021.

\bibitem{786990}
Jinggang Huang and D.~Mumford.
\newblock Statistics of natural images and models.
\newblock In {\em Proceedings. 1999 IEEE Computer Society Conference on
  Computer Vision and Pattern Recognition (Cat. No PR00149)}, volume~1, pages
  541--547 Vol. 1, 1999.

\bibitem{huang2020self}
Lang Huang, Chao Zhang, and Hongyang Zhang.
\newblock Self-adaptive training: beyond empirical risk minimization.
\newblock In {\em Advances in Neural Information Processing Systems},
  volume~33, 2020.

\bibitem{huang2022transferable}
Yi~Huang and Adams Wai-Kin Kong.
\newblock Transferable adversarial attack based on integrated gradients.
\newblock In {\em International Conference on Learning Representations}, 2022.

\bibitem{Huang2020BlackBox}
Zhichao Huang and Tong Zhang.
\newblock Black-box adversarial attack with transferable model-based embedding.
\newblock In {\em International Conference on Learning Representations}, 2020.

\bibitem{Jang19}
Yunseok Jang, Tianchen Zhao, Seunghoon Hong, and Honglak Lee.
\newblock Adversarial defense via learning to generate diverse attacks.
\newblock In {\em 2019 IEEE/CVF International Conference on Computer Vision
  (ICCV)}, pages 2740--2749, 2019.

\bibitem{Jha18}
Susmit Jha, Uyeong Jang, Somesh Jha, and Brian Jalaian.
\newblock Detecting adversarial examples using data manifolds.
\newblock In {\em MILCOM 2018 - 2018 IEEE Military Communications Conference
  (MILCOM)}, pages 547--552, 2018.

\bibitem{Jia2019ComDefendAE}
Xiaojun Jia, Xingxing Wei, Xiaochun Cao, and Hassan Foroosh.
\newblock Comdefend: An efficient image compression model to defend adversarial
  examples.
\newblock {\em 2019 IEEE/CVF Conference on Computer Vision and Pattern
  Recognition (CVPR)}, pages 6077--6085, 2019.

\bibitem{Jiang2020LocalIG}
Chiyu~Max Jiang, Avneesh Sud, Ameesh Makadia, Jingwei Huang, Matthias
  Nie{\ss}ner, and Thomas~A. Funkhouser.
\newblock Local implicit grid representations for 3d scenes.
\newblock {\em 2020 IEEE/CVF Conference on Computer Vision and Pattern
  Recognition (CVPR)}, pages 6000--6009, 2020.

\bibitem{jin2021manifold}
Charles Jin and Martin Rinard.
\newblock Manifold regularization for locally stable deep neural networks,
  2021.

\bibitem{kang2021stable}
QIYU KANG, Yang Song, Qinxu Ding, and Wee~Peng Tay.
\newblock Stable neural {ODE} with lyapunov-stable equilibrium points for
  defending against adversarial attacks.
\newblock In A.~Beygelzimer, Y.~Dauphin, P.~Liang, and J.~Wortman Vaughan,
  editors, {\em Advances in Neural Information Processing Systems}, 2021.

\bibitem{Karras2021AliasFreeGA}
Tero Karras, Miika Aittala, Samuli Laine, Erik Harkonen, Janne Hellsten, Jaakko
  Lehtinen, and Timo Aila.
\newblock Alias-free generative adversarial networks.
\newblock In {\em NeurIPS}, 2021.

\bibitem{Karras2019ASG}
Tero Karras, Samuli Laine, and Timo Aila.
\newblock A style-based generator architecture for generative adversarial
  networks.
\newblock {\em 2019 IEEE/CVF Conference on Computer Vision and Pattern
  Recognition (CVPR)}, pages 4396--4405, 2019.

\bibitem{Karras2020AnalyzingAI}
Tero Karras, Samuli Laine, Miika Aittala, Janne Hellsten, Jaakko Lehtinen, and
  Timo Aila.
\newblock Analyzing and improving the image quality of stylegan.
\newblock {\em 2020 IEEE/CVF Conference on Computer Vision and Pattern
  Recognition (CVPR)}, pages 8107--8116, 2020.

\bibitem{Khoury2018OnTG}
Marc Khoury and Dylan Hadfield-Menell.
\newblock On the geometry of adversarial examples.
\newblock {\em ArXiv}, abs/1811.00525, 2018.

\bibitem{kim2020sensible}
Jungeum Kim and Xiao Wang.
\newblock Sensible adversarial learning, 2020.

\bibitem{semantic_srn}
A.~Kohli, V.~Sitzmann, and G.~Wetzstein.
\newblock {Semantic Implicit Neural Scene Representations with Semi-supervised
  Training}.
\newblock In {\em International Conference on 3D Vision (3DV)}, 2020.

\bibitem{Kolesnikov17}
Alexander Kolesnikov and Christoph~H. Lampert.
\newblock Pixelcnn models with auxiliary variables for natural image modeling.
\newblock In {\em Proceedings of the 34th International Conference on Machine
  Learning - Volume 70}, ICML'17, page 1905–1914. JMLR.org, 2017.

\bibitem{openimages}
Ivan Krasin, Tom Duerig, Neil Alldrin, Vittorio Ferrari, Sami Abu-El-Haija,
  Alina Kuznetsova, Hassan Rom, Jasper Uijlings, Stefan Popov, Andreas Veit,
  Serge Belongie, Victor Gomes, Abhinav Gupta, Chen Sun, Gal Chechik, David
  Cai, Zheyun Feng, Dhyanesh Narayanan, and Kevin Murphy.
\newblock Openimages: A public dataset for large-scale multi-label and
  multi-class image classification.
\newblock {\em Dataset available from https://github.com/openimages}, 2017.

\bibitem{cifar10}
Alex Krizhevsky, Vinod Nair, and Geoffrey Hinton.
\newblock Cifar-10 (canadian institute for advanced research).

\bibitem{cifar100}
Alex Krizhevsky, Vinod Nair, and Geoffrey Hinton.
\newblock Cifar-100 (canadian institute for advanced research).

\bibitem{Souvik21}
Souvik Kundu, Mahdi Nazemi, Peter~A. Beerel, and Massoud Pedram.
\newblock Dnr: A tunable robust pruning framework through dynamic network
  rewiring of dnns.
\newblock In {\em Proceedings of the 26th Asia and South Pacific Design
  Automation Conference}, ASPDAC '21, page 344–350, New York, NY, USA, 2021.
  Association for Computing Machinery.

\bibitem{bim}
Alexey Kurakin, Ian Goodfellow, and Samy Bengio.
\newblock Adversarial examples in the physical world.
\newblock {\em ICLR Workshop}, 2017.

\bibitem{Laidlaw2019FunctionalAA}
Cassidy Laidlaw and Soheil Feizi.
\newblock Functional adversarial attacks.
\newblock In {\em NeurIPS}, 2019.

\bibitem{Lee2017GenerativeAT}
Hyeungill Lee, Sungyeob Han, and Jungwoo Lee.
\newblock Generative adversarial trainer: Defense to adversarial perturbations
  with gan.
\newblock {\em ArXiv}, abs/1705.03387, 2017.

\bibitem{leung2022black}
Brandon Leung, Chih-Hui Ho, and Nuno Vasconcelos.
\newblock Black-box test-time shape refinement for single view 3d
  reconstruction.
\newblock In {\em Proceedings of the IEEE/CVF Conference on Computer Vision and
  Pattern Recognition}, pages 4080--4090, 2022.

\bibitem{Li2021ExploringAF}
Dongze Li, Wei Wang, Hongxing Fan, and Jing Dong.
\newblock Exploring adversarial fake images on face manifold.
\newblock {\em 2021 IEEE/CVF Conference on Computer Vision and Pattern
  Recognition (CVPR)}, pages 5785--5794, 2021.

\bibitem{edsr}
Bee Lim, Sanghyun Son, Heewon Kim, Seungjun Nah, and Kyoung~Mu Lee.
\newblock Enhanced deep residual networks for single image super-resolution.
\newblock In {\em The IEEE Conference on Computer Vision and Pattern
  Recognition (CVPR) Workshops}, July 2017.

\bibitem{NEURIPS2020_23937b42}
Wei-An Lin, Chun~Pong Lau, Alexander Levine, Rama Chellappa, and Soheil Feizi.
\newblock Dual manifold adversarial robustness: Defense against lp and non-lp
  adversarial attacks.
\newblock In H.~Larochelle, M.~Ranzato, R.~Hadsell, M.F. Balcan, and H.~Lin,
  editors, {\em Advances in Neural Information Processing Systems}, volume~33,
  pages 3487--3498. Curran Associates, Inc., 2020.

\bibitem{eotpgd}
Xuanqing Liu, Yao Li, Chongruo Wu, and Cho-Jui Hsieh.
\newblock Adv-{BNN}: Improved adversarial defense through robust bayesian
  neural network.
\newblock In {\em International Conference on Learning Representations}, 2019.

\bibitem{Liu2019FeatureDD}
Zihao Liu, Qi~Liu, Tao Liu, Yanzhi Wang, and Wujie Wen.
\newblock Feature distillation: Dnn-oriented jpeg compression against
  adversarial examples.
\newblock {\em 2019 IEEE/CVF Conference on Computer Vision and Pattern
  Recognition (CVPR)}, pages 860--868, 2019.

\bibitem{pgd}
Aleksander Madry, Aleksandar Makelov, Ludwig Schmidt, Dimitris Tsipras, and
  Adrian Vladu.
\newblock Towards deep learning models resistant to adversarial attacks.
\newblock In {\em International Conference on Learning Representations}, 2018.

\bibitem{madry2018towards}
Aleksander Madry, Aleksandar Makelov, Ludwig Schmidt, Dimitris Tsipras, and
  Adrian Vladu.
\newblock Towards deep learning models resistant to adversarial attacks.
\newblock In {\em International Conference on Learning Representations}, 2018.

\bibitem{Mahendran2015UnderstandingDI}
Aravindh Mahendran and Andrea Vedaldi.
\newblock Understanding deep image representations by inverting them.
\newblock {\em 2015 IEEE Conference on Computer Vision and Pattern Recognition
  (CVPR)}, pages 5188--5196, 2015.

\bibitem{9710949}
Chengzhi Mao, Mia Chiquier, Hao Wang, Junfeng Yang, and Carl Vondrick.
\newblock Adversarial attacks are reversible with natural supervision.
\newblock In {\em 2021 IEEE/CVF International Conference on Computer Vision
  (ICCV)}, pages 641--651, 2021.

\bibitem{NEURIPS2019_c24cd76e}
Chengzhi Mao, Ziyuan Zhong, Junfeng Yang, Carl Vondrick, and Baishakhi Ray.
\newblock Metric learning for adversarial robustness.
\newblock In H.~Wallach, H.~Larochelle, A.~Beygelzimer, F.~d\textquotesingle
  Alch\'{e}-Buc, E.~Fox, and R.~Garnett, editors, {\em Advances in Neural
  Information Processing Systems}, volume~32. Curran Associates, Inc., 2019.

\bibitem{magnet}
Dongyu Meng and Hao Chen.
\newblock Magnet: A two-pronged defense against adversarial examples.
\newblock In {\em Proceedings of the 2017 ACM SIGSAC Conference on Computer and
  Communications Security}, CCS '17, page 135–147, New York, NY, USA, 2017.
  Association for Computing Machinery.

\bibitem{Mescheder2019OccupancyNL}
Lars~M. Mescheder, Michael Oechsle, Michael Niemeyer, Sebastian Nowozin, and
  Andreas Geiger.
\newblock Occupancy networks: Learning 3d reconstruction in function space.
\newblock {\em 2019 IEEE/CVF Conference on Computer Vision and Pattern
  Recognition (CVPR)}, pages 4455--4465, 2019.

\bibitem{Mildenhall2020NeRFRS}
Ben Mildenhall, Pratul~P. Srinivasan, Matthew Tancik, Jonathan~T. Barron, Ravi
  Ramamoorthi, and Ren Ng.
\newblock Nerf: Representing scenes as neural radiance fields for view
  synthesis.
\newblock In {\em ECCV}, 2020.

\bibitem{deepfool}
Seyed-Mohsen Moosavi-Dezfooli, Alhussein Fawzi, and Pascal Frossard.
\newblock Deepfool: A simple and accurate method to fool deep neural networks.
\newblock {\em 2016 IEEE Conference on Computer Vision and Pattern Recognition
  (CVPR)}, pages 2574--2582, 2016.

\bibitem{MoosaviDezfooli2019RobustnessVC}
Seyed-Mohsen Moosavi-Dezfooli, Alhussein Fawzi, Jonathan Uesato, and Pascal
  Frossard.
\newblock Robustness via curvature regularization, and vice versa.
\newblock {\em 2019 IEEE/CVF Conference on Computer Vision and Pattern
  Recognition (CVPR)}, pages 9070--9078, 2019.

\bibitem{Mustafa_2019_ICCV}
Aamir Mustafa, Salman Khan, Munawar Hayat, Roland Goecke, Jianbing Shen, and
  Ling Shao.
\newblock Adversarial defense by restricting the hidden space of deep neural
  networks.
\newblock In {\em The IEEE International Conference on Computer Vision (ICCV)},
  October 2019.

\bibitem{Nguyen2017PlugP}
Anh~M Nguyen, Jeff Clune, Yoshua Bengio, Alexey Dosovitskiy, and Jason
  Yosinski.
\newblock Plug \& play generative networks: Conditional iterative generation of
  images in latent space.
\newblock {\em 2017 IEEE Conference on Computer Vision and Pattern Recognition
  (CVPR)}, pages 3510--3520, 2017.

\bibitem{nie2022DiffPure}
Weili Nie, Brandon Guo, Yujia Huang, Chaowei Xiao, Arash Vahdat, and Anima
  Anandkumar.
\newblock Diffusion models for adversarial purification.
\newblock In {\em International Conference on Machine Learning (ICML)}, 2022.

\bibitem{Niemeyer2020DifferentiableVR}
Michael Niemeyer, Lars~M. Mescheder, Michael Oechsle, and Andreas Geiger.
\newblock Differentiable volumetric rendering: Learning implicit 3d
  representations without 3d supervision.
\newblock {\em 2020 IEEE/CVF Conference on Computer Vision and Pattern
  Recognition (CVPR)}, pages 3501--3512, 2020.

\bibitem{Olshausen1996EmergenceOS}
Bruno~A. Olshausen and David~J. Field.
\newblock Emergence of simple-cell receptive field properties by learning a
  sparse code for natural images.
\newblock {\em Nature}, 381:607--609, 1996.

\bibitem{Osadchy17}
Margarita Osadchy, Julio Hernandez-Castro, Stuart Gibson, Orr Dunkelman, and
  Daniel Pérez-Cabo.
\newblock No bot expects the deepcaptcha! introducing immutable adversarial
  examples, with applications to captcha generation.
\newblock {\em IEEE Transactions on Information Forensics and Security},
  12(11):2640--2653, 2017.

\bibitem{OZDAG2018152}
Mesut Ozdag.
\newblock Adversarial attacks and defenses against deep neural networks: A
  survey.
\newblock {\em Procedia Computer Science}, 140:152--161, 2018.
\newblock Cyber Physical Systems and Deep Learning Chicago, Illinois November
  5-7, 2018.

\bibitem{Pang22}
Tianyu Pang, Min Lin, Xiao Yang, Jun Zhu, and Shuicheng Yan.
\newblock Robustness and accuracy could be reconcilable by (proper) definition,
  2022.

\bibitem{Pang2020Rethinking}
Tianyu Pang, Kun Xu, Yinpeng Dong, Chao Du, Ning Chen, and Jun Zhu.
\newblock Rethinking softmax cross-entropy loss for adversarial robustness.
\newblock In {\em International Conference on Learning Representations}, 2020.

\bibitem{pang2021bag}
Tianyu Pang, Xiao Yang, Yinpeng Dong, Hang Su, and Jun Zhu.
\newblock Bag of tricks for adversarial training.
\newblock In {\em International Conference on Learning Representations}, 2021.

\bibitem{NEURIPS2020_Pang}
Tianyu Pang, Xiao Yang, Yinpeng Dong, Kun Xu, Jun Zhu, and Hang Su.
\newblock Boosting adversarial training with hypersphere embedding.
\newblock In H.~Larochelle, M.~Ranzato, R.~Hadsell, M.F. Balcan, and H.~Lin,
  editors, {\em Advances in Neural Information Processing Systems}, volume~33,
  pages 7779--7792. Curran Associates, Inc., 2020.

\bibitem{Papernot17}
Nicolas Papernot, Patrick McDaniel, Ian Goodfellow, Somesh Jha, Z.~Berkay
  Celik, and Ananthram Swami.
\newblock Practical black-box attacks against machine learning.
\newblock In {\em Proceedings of the 2017 ACM on Asia Conference on Computer
  and Communications Security}, ASIA CCS '17, page 506–519, New York, NY,
  USA, 2017. Association for Computing Machinery.

\bibitem{Park2019DeepSDFLC}
Jeong~Joon Park, Peter~R. Florence, Julian Straub, Richard~A. Newcombe, and
  S.~Lovegrove.
\newblock Deepsdf: Learning continuous signed distance functions for shape
  representation.
\newblock {\em 2019 IEEE/CVF Conference on Computer Vision and Pattern
  Recognition (CVPR)}, pages 165--174, 2019.

\bibitem{pytorch}
Adam Paszke, Sam Gross, Francisco Massa, Adam Lerer, James Bradbury, Gregory
  Chanan, Trevor Killeen, Zeming Lin, Natalia Gimelshein, Luca Antiga, Alban
  Desmaison, Andreas Kopf, Edward Yang, Zachary DeVito, Martin Raison, Alykhan
  Tejani, Sasank Chilamkurthy, Benoit Steiner, Lu~Fang, Junjie Bai, and Soumith
  Chintala.
\newblock Pytorch: An imperative style, high-performance deep learning library.
\newblock In H.~Wallach, H.~Larochelle, A.~Beygelzimer, F.~d\textquotesingle
  Alch\'{e}-Buc, E.~Fox, and R.~Garnett, editors, {\em Advances in Neural
  Information Processing Systems 32}, pages 8024--8035. Curran Associates,
  Inc., 2019.

\bibitem{Prakash2018DeflectingAA}
Aaditya Prakash, Nick Moran, Solomon Garber, Antonella DiLillo, and James~A.
  Storer.
\newblock Deflecting adversarial attacks with pixel deflection.
\newblock {\em 2018 IEEE/CVF Conference on Computer Vision and Pattern
  Recognition}, pages 8571--8580, 2018.

\bibitem{Qin2019AdversarialRT}
Chongli Qin, James Martens, Sven Gowal, Dilip Krishnan, Krishnamurthy
  Dvijotham, Alhussein Fawzi, Soham De, Robert Stanforth, and Pushmeet Kohli.
\newblock Adversarial robustness through local linearization.
\newblock In {\em NeurIPS}, 2019.

\bibitem{rade2021helperbased}
Rahul Rade and Seyed-Mohsen Moosavi-Dezfooli.
\newblock Helper-based adversarial training: Reducing excessive margin to
  achieve a better accuracy vs. robustness trade-off.
\newblock In {\em ICML 2021 Workshop on Adversarial Machine Learning}, 2021.

\bibitem{Rebuffi2021FixingDA}
Sylvestre-Alvise Rebuffi, Sven Gowal, Dan~Andrei Calian, Florian Stimberg,
  Olivia Wiles, and Timothy~A. Mann.
\newblock Fixing data augmentation to improve adversarial robustness.
\newblock {\em ArXiv}, abs/2103.01946, 2021.

\bibitem{REN2020346}
Kui Ren, Tianhang Zheng, Zhan Qin, and Xue Liu.
\newblock Adversarial attacks and defenses in deep learning.
\newblock {\em Engineering}, 6(3):346--360, 2020.

\bibitem{Rice2020OverfittingIA}
Leslie Rice, Eric Wong, and J.~Zico Kolter.
\newblock Overfitting in adversarially robust deep learning.
\newblock In {\em ICML}, 2020.

\bibitem{Luke21}
Luke~E. Richards, Andr{\'{e}}~T. Nguyen, Ryan Capps, Steven Forsyth, Cynthia
  Matuszek, and Edward Raff.
\newblock Adversarial transfer attacks with unknown data and class overlap.
\newblock {\em CoRR}, abs/2109.11125, 2021.

\bibitem{Rony2019DecouplingDA}
J{\'e}r{\^o}me Rony, Luiz~G. Hafemann, Luiz Oliveira, Ismail~Ben Ayed, Robert
  Sabourin, and Eric Granger.
\newblock Decoupling direction and norm for efficient gradient-based l2
  adversarial attacks and defenses.
\newblock {\em 2019 IEEE/CVF Conference on Computer Vision and Pattern
  Recognition (CVPR)}, pages 4317--4325, 2019.

\bibitem{Ruderman1994TheSO}
Daniel~L. Ruderman.
\newblock The statistics of natural images.
\newblock {\em Network: Computation In Neural Systems}, 5:517--548, 1994.

\bibitem{rusu22a}
Andrei~A Rusu, Dan~Andrei Calian, Sven Gowal, and Raia Hadsell.
\newblock Hindering adversarial attacks with implicit neural representations.
\newblock In Kamalika Chaudhuri, Stefanie Jegelka, Le~Song, Csaba Szepesvari,
  Gang Niu, and Sivan Sabato, editors, {\em Proceedings of the 39th
  International Conference on Machine Learning}, volume 162 of {\em Proceedings
  of Machine Learning Research}, pages 18910--18934. PMLR, 17--23 Jul 2022.

\bibitem{Saito2019PIFuPI}
Shunsuke Saito, Zeng Huang, Ryota Natsume, Shigeo Morishima, Angjoo Kanazawa,
  and Hao Li.
\newblock Pifu: Pixel-aligned implicit function for high-resolution clothed
  human digitization.
\newblock {\em 2019 IEEE/CVF International Conference on Computer Vision
  (ICCV)}, pages 2304--2314, 2019.

\bibitem{Salimans2017PixelCNNIT}
Tim Salimans, Andrej Karpathy, Xi~Chen, and Diederik~P. Kingma.
\newblock Pixelcnn++: Improving the pixelcnn with discretized logistic mixture
  likelihood and other modifications.
\newblock {\em ArXiv}, abs/1701.05517, 2017.

\bibitem{NEURIPS2020_24357dd0}
Hadi Salman, Andrew Ilyas, Logan Engstrom, Ashish Kapoor, and Aleksander Madry.
\newblock Do adversarially robust imagenet models transfer better?
\newblock In H.~Larochelle, M.~Ranzato, R.~Hadsell, M.F. Balcan, and H.~Lin,
  editors, {\em Advances in Neural Information Processing Systems}, volume~33,
  pages 3533--3545. Curran Associates, Inc., 2020.

\bibitem{defensegan}
Pouya Samangouei, Maya Kabkab, and Rama Chellappa.
\newblock Defense-{GAN}: Protecting classifiers against adversarial attacks
  using generative models.
\newblock In {\em International Conference on Learning Representations}, 2018.

\bibitem{jitter}
Leo Schwinn, Ren{\'{e}} Raab, An~Nguyen, Dario Zanca, and Bjoern~M. Eskofier.
\newblock Exploring misclassifications of robust neural networks to enhance
  adversarial attacks.
\newblock {\em CoRR}, abs/2105.10304, 2021.

\bibitem{sehwag2022robust}
Vikash Sehwag, Saeed Mahloujifar, Tinashe Handina, Sihui Dai, Chong Xiang, Mung
  Chiang, and Prateek Mittal.
\newblock Robust learning meets generative models: Can proxy distributions
  improve adversarial robustness?
\newblock In {\em International Conference on Learning Representations}, 2022.

\bibitem{sehwag2022Proxy}
Vikash Sehwag, Saeed Mahloujifar, Tinashe Handina, Sihui Dai, Chong Xiang, Mung
  Chiang, and Prateek Mittal.
\newblock Robust learning meets generative models: Can proxy distributions
  improve adversarial robustness?
\newblock In {\em International Conference on Learning Representations}, 2022.

\bibitem{sehwag2020hydra}
Vikash Sehwag, Shiqi Wang, Prateek Mittal, and Suman Jana.
\newblock Hydra: Pruning adversarially robust neural networks.
\newblock {\em Advances in Neural Information Processing Systems}, 33, 2020.

\bibitem{shafahi2018are}
Ali Shafahi, W.~Ronny Huang, Christoph Studer, Soheil Feizi, and Tom Goldstein.
\newblock Are adversarial examples inevitable?
\newblock In {\em International Conference on Learning Representations}, 2019.

\bibitem{Shafahi2019AdversarialTF}
Ali Shafahi, Mahyar Najibi, Amin Ghiasi, Zheng Xu, John~P. Dickerson, Christoph
  Studer, Larry~S. Davis, Gavin Taylor, and Tom Goldstein.
\newblock Adversarial training for free!
\newblock In {\em NeurIPS}, 2019.

\bibitem{Simoncelli2001NaturalIS}
Eero~P. Simoncelli and Bruno~A. Olshausen.
\newblock Natural image statistics and neural representation.
\newblock {\em Annual review of neuroscience}, 24:1193--216, 2001.

\bibitem{vgg}
Karen Simonyan and Andrew Zisserman.
\newblock Very deep convolutional networks for large-scale image recognition.
\newblock {\em CoRR}, abs/1409.1556, 2015.

\bibitem{Sinha2019HarnessingTV}
Abhishek Sinha, Mayank~Kumar Singh, Nupur Kumari, Balaji Krishnamurthy,
  Harshitha Machiraju, and Vineeth~N. Balasubramanian.
\newblock Harnessing the vulnerability of latent layers in adversarially
  trained models.
\newblock {\em ArXiv}, abs/1905.05186, 2019.

\bibitem{Chawin20}
Chawin Sitawarin, Supriyo Chakraborty, and David~A. Wagner.
\newblock Improving adversarial robustness through progressive hardening.
\newblock {\em CoRR}, abs/2003.09347, 2020.

\bibitem{Sitzmann2020ImplicitNR}
Vincent Sitzmann, Julien N.~P. Martel, Alexander~W. Bergman, David~B. Lindell,
  and Gordon Wetzstein.
\newblock Implicit neural representations with periodic activation functions.
\newblock {\em ArXiv}, abs/2006.09661, 2020.

\bibitem{pixeldefend}
Yang Song, Taesup Kim, Sebastian Nowozin, Stefano Ermon, and Nate Kushman.
\newblock Pixeldefend: Leveraging generative models to understand and defend
  against adversarial examples.
\newblock In {\em International Conference on Learning Representations}, 2018.

\bibitem{Sridhar2021RobustLV}
Kaustubh Sridhar, Oleg Sokolsky, Insup Lee, and James Weimer.
\newblock Robust learning via persistency of excitation.
\newblock {\em ArXiv}, abs/2106.02078, 2021.

\bibitem{Srivastava2004OnAI}
A.~Srivastava, A.~B. Lee, Eero~P. Simoncelli, and S.-C. Zhu.
\newblock On advances in statistical modeling of natural images.
\newblock {\em Journal of Mathematical Imaging and Vision}, 18:17--33, 2004.

\bibitem{onepixelattack}
Jiawei Su, Danilo~Vasconcellos Vargas, and Kouichi Sakurai.
\newblock One pixel attack for fooling deep neural networks.
\newblock {\em IEEE Transactions on Evolutionary Computation}, 23:828--841,
  2019.

\bibitem{stl}
Bo~Sun, Nian hsuan Tsai, Fangchen Liu, Ronald Yu, and Hao Su.
\newblock Adversarial defense by stratified convolutional sparse coding.
\newblock {\em 2019 IEEE/CVF Conference on Computer Vision and Pattern
  Recognition (CVPR)}, pages 11439--11448, 2019.

\bibitem{inceptionnet}
Christian Szegedy, Vincent Vanhoucke, Sergey Ioffe, Jon Shlens, and Zbigniew
  Wojna.
\newblock Rethinking the inception architecture for computer vision.
\newblock In {\em 2016 IEEE Conference on Computer Vision and Pattern
  Recognition (CVPR)}, pages 2818--2826, 2016.

\bibitem{efficientnet}
Mingxing Tan and Quoc~V. Le.
\newblock Efficientnet: Rethinking model scaling for convolutional neural
  networks.
\newblock In {\em ICML}, 2019.

\bibitem{Torralba2003StatisticsON}
Antonio Torralba and Aude Oliva.
\newblock Statistics of natural image categories.
\newblock {\em Network: Computation in Neural Systems}, 14:391 -- 412, 2003.

\bibitem{adaptiveattack}
Florian Tramer, Nicholas Carlini, Wieland Brendel, and Aleksander Madry.
\newblock On adaptive attacks to adversarial example defenses.
\newblock In H.~Larochelle, M.~Ranzato, R.~Hadsell, M.F. Balcan, and H.~Lin,
  editors, {\em Advances in Neural Information Processing Systems}, volume~33,
  pages 1633--1645. Curran Associates, Inc., 2020.

\bibitem{rfgsm}
Florian Tramèr, Alexey Kurakin, Nicolas Papernot, Ian Goodfellow, Dan Boneh,
  and Patrick McDaniel.
\newblock Ensemble adversarial training: Attacks and defenses.
\newblock In {\em International Conference on Learning Representations}, 2018.

\bibitem{Oord2016PixelRN}
A{\"a}ron van~den Oord, Nal Kalchbrenner, and Koray Kavukcuoglu.
\newblock Pixel recurrent neural networks.
\newblock {\em ArXiv}, abs/1601.06759, 2016.

\bibitem{vasconcelos08}
Manuela Vasconcelos and Nuno Vasconcelos.
\newblock Natural image statistics and low-complexity feature selection.
\newblock {\em IEEE Transactions on Pattern Analysis and Machine Intelligence},
  31(2):228--244, 2009.

\bibitem{Wang2019BilateralAT}
Jianyu Wang.
\newblock Bilateral adversarial training: Towards fast training of more robust
  models against adversarial attacks.
\newblock {\em 2019 IEEE/CVF International Conference on Computer Vision
  (ICCV)}, pages 6628--6637, 2019.

\bibitem{Wang2021IMAGINEIS}
Pei Wang, Yijun Li, Krishna~Kumar Singh, Jingwan Lu, and Nuno Vasconcelos.
\newblock Imagine: Image synthesis by image-guided model inversion.
\newblock {\em 2021 IEEE/CVF Conference on Computer Vision and Pattern
  Recognition (CVPR)}, pages 3680--3689, 2021.

\bibitem{Wang_2021_CVPR}
Xiaosen Wang and Kun He.
\newblock Enhancing the transferability of adversarial attacks through variance
  tuning.
\newblock In {\em Proceedings of the IEEE/CVF Conference on Computer Vision and
  Pattern Recognition (CVPR)}, pages 1924--1933, June 2021.

\bibitem{Wang2021AdmixET}
Xiaosen Wang, Xu~He, Jingdong Wang, and Kun He.
\newblock Admix: Enhancing the transferability of adversarial attacks.
\newblock {\em 2021 IEEE/CVF International Conference on Computer Vision
  (ICCV)}, pages 16138--16147, 2021.

\bibitem{Wang2020Improving}
Yisen Wang, Difan Zou, Jinfeng Yi, James Bailey, Xingjun Ma, and Quanquan Gu.
\newblock Improving adversarial robustness requires revisiting misclassified
  examples.
\newblock In {\em International Conference on Learning Representations}, 2020.

\bibitem{Wong2020Fast}
Eric Wong, Leslie Rice, and J.~Zico Kolter.
\newblock Fast is better than free: Revisiting adversarial training.
\newblock In {\em International Conference on Learning Representations}, 2020.

\bibitem{ffgsm}
Eric Wong, Leslie Rice, and J.~Zico Kolter.
\newblock Fast is better than free: Revisiting adversarial training.
\newblock In {\em International Conference on Learning Representations}, 2020.

\bibitem{wu2021do}
Boxi Wu, Jinghui Chen, Deng Cai, Xiaofei He, and Quanquan Gu.
\newblock Do wider neural networks really help adversarial robustness?
\newblock In A.~Beygelzimer, Y.~Dauphin, P.~Liang, and J.~Wortman Vaughan,
  editors, {\em Advances in Neural Information Processing Systems}, 2021.

\bibitem{Wu2020Skip}
Dongxian Wu, Yisen Wang, Shu-Tao Xia, James Bailey, and Xingjun Ma.
\newblock Skip connections matter: On the transferability of adversarial
  examples generated with resnets.
\newblock In {\em International Conference on Learning Representations}, 2020.

\bibitem{wu2020adversarial}
Dongxian Wu, Shu-Tao Xia, and Yisen Wang.
\newblock Adversarial weight perturbation helps robust generalization.
\newblock In {\em NeurIPS}, 2020.

\bibitem{Wu2020IFDefense3A}
Ziyi Wu, Yueqi Duan, He~Wang, Qingnan Fan, and Leonidas~J. Guibas.
\newblock If-defense: 3d adversarial point cloud defense via implicit function
  based restoration.
\newblock {\em ArXiv}, abs/2010.05272, 2020.

\bibitem{Xiao2020Enhancing}
Chang Xiao, Peilin Zhong, and Changxi Zheng.
\newblock Enhancing adversarial defense by k-winners-take-all.
\newblock In {\em International Conference on Learning Representations}, 2020.

\bibitem{xie2017mitigating}
Cihang Xie, Jianyu Wang, Zhishuai Zhang, Zhou Ren, and Alan Yuille.
\newblock Mitigating adversarial effects through randomization.
\newblock In {\em International Conference on Learning Representations}, 2018.

\bibitem{difgsm}
Cihang Xie, Zhishuai Zhang, Yuyin Zhou, Song Bai, Jianyu Wang, Zhou Ren, and
  Alan Yuille.
\newblock Improving transferability of adversarial examples with input
  diversity.
\newblock In {\em Computer Vision and Pattern Recognition}. IEEE, 2019.

\bibitem{NIPS2019_8340}
Qiangeng Xu, Weiyue Wang, Duygu Ceylan, Radomir Mech, and Ulrich Neumann.
\newblock Disn: Deep implicit surface network for high-quality single-view 3d
  reconstruction.
\newblock In H.~Wallach, H.~Larochelle, A.~Beygelzimer, F.~d\textquotesingle
  Alch\'{e}-Buc, E.~Fox, and R.~Garnett, editors, {\em Advances in Neural
  Information Processing Systems 32}, pages 492--502. Curran Associates, Inc.,
  2019.

\bibitem{Xu2018FeatureSD}
Weilin Xu, David Evans, and Yanjun Qi.
\newblock Feature squeezing: Detecting adversarial examples in deep neural
  networks.
\newblock {\em ArXiv}, abs/1704.01155, 2018.

\bibitem{Yin2020DreamingTD}
Hongxu Yin, Pavlo Molchanov, Zhizhong Li, Jos{\'e}~Manuel {\'A}lvarez, Arun
  Mallya, Derek Hoiem, Niraj~Kumar Jha, and Jan Kautz.
\newblock Dreaming to distill: Data-free knowledge transfer via deepinversion.
\newblock {\em 2020 IEEE/CVF Conference on Computer Vision and Pattern
  Recognition (CVPR)}, pages 8712--8721, 2020.

\bibitem{Yin2020GAT}
Xuwang Yin, Soheil Kolouri, and Gustavo~K Rohde.
\newblock Gat: Generative adversarial training for adversarial example
  detection and robust classification.
\newblock In {\em International Conference on Learning Representations}, 2020.

\bibitem{Yoon2021AdversarialPW}
Jongmin Yoon, Sung~Ju Hwang, and Juho Lee.
\newblock Adversarial purification with score-based generative models.
\newblock In {\em ICML}, 2021.

\bibitem{yu2022generating}
Sihyun Yu, Jihoon Tack, Sangwoo Mo, Hyunsu Kim, Junho Kim, Jung-Woo Ha, and
  Jinwoo Shin.
\newblock Generating videos with dynamics-aware implicit generative adversarial
  networks.
\newblock In {\em International Conference on Learning Representations}, 2022.

\bibitem{Yuan2020EnsembleGC}
Jianhe Yuan and Zhihai He.
\newblock Ensemble generative cleaning with feedback loops for defending
  adversarial attacks.
\newblock {\em 2020 IEEE/CVF Conference on Computer Vision and Pattern
  Recognition (CVPR)}, pages 578--587, 2020.

\bibitem{BMVC2016_87}
Sergey Zagoruyko and Nikos Komodakis.
\newblock Wide residual networks.
\newblock In Edwin R.~Hancock Richard C.~Wilson and William A.~P. Smith,
  editors, {\em Proceedings of the British Machine Vision Conference (BMVC)},
  pages 87.1--87.12. BMVA Press, September 2016.

\bibitem{Zhang19}
Dinghuai Zhang, Tianyuan Zhang, Yiping Lu, Zhanxing Zhu, and Bin Dong.
\newblock You only propagate once: Accelerating adversarial training via
  maximal principle.
\newblock In H.~Wallach, H.~Larochelle, A.~Beygelzimer, F.~d\textquotesingle
  Alch\'{e}-Buc, E.~Fox, and R.~Garnett, editors, {\em Advances in Neural
  Information Processing Systems}, volume~32. Curran Associates, Inc., 2019.

\bibitem{feature_scatter}
Haichao Zhang and Jianyu Wang.
\newblock Defense against adversarial attacks using feature scattering-based
  adversarial training.
\newblock In {\em Advances in Neural Information Processing Systems}, 2019.

\bibitem{zhang2020adversarial}
Haichao Zhang and Wei Xu.
\newblock Adversarial interpolation training: A simple approach for improving
  model robustness, 2020.

\bibitem{zhang2019theoretically}
Hongyang Zhang, Yaodong Yu, Jiantao Jiao, Eric~P. Xing, Laurent~El Ghaoui, and
  Michael~I. Jordan.
\newblock Theoretically principled trade-off between robustness and accuracy.
\newblock In {\em International Conference on Machine Learning}, 2019.

\bibitem{zhang2020towards}
Huan Zhang, Hongge Chen, Chaowei Xiao, Sven Gowal, Robert Stanforth, Bo~Li,
  Duane Boning, and Cho-Jui Hsieh.
\newblock Towards stable and efficient training of verifiably robust neural
  networks.
\newblock In {\em International Conference on Learning Representations}, 2020.

\bibitem{zhang2020fat}
Jingfeng Zhang, Xilie Xu, Bo~Han, Gang Niu, Lizhen Cui, Masashi Sugiyama, and
  Mohan Kankanhalli.
\newblock Attacks which do not kill training make adversarial learning
  stronger.
\newblock In {\em ICML}, 2020.

\bibitem{zhang2021geometryaware}
Jingfeng Zhang, Jianing Zhu, Gang Niu, Bo~Han, Masashi Sugiyama, and Mohan
  Kankanhalli.
\newblock Geometry-aware instance-reweighted adversarial training.
\newblock In {\em International Conference on Learning Representations}, 2021.

\bibitem{Zhou_eccv20}
Jianli Zhou, Chao Liang, and Jun Chen.
\newblock Manifold projection for adversarial defense on face recognition.
\newblock In {\em Computer Vision – ECCV 2020: 16th European Conference,
  Glasgow, UK, August 23–28, 2020, Proceedings, Part XXX}, page 288–305,
  Berlin, Heidelberg, 2020. Springer-Verlag.

\bibitem{zuiderveld2021towards}
Jan Zuiderveld, Marco Federici, and Erik~J Bekkers.
\newblock Towards lightweight controllable audio synthesis with conditional
  implicit neural representations.
\newblock In {\em NeurIPS 2021 Workshop on Deep Generative Models and
  Downstream Applications}, 2021.

\end{thebibliography}
\bibliographystyle{plain}

\end{document}